\documentclass[journal]{IEEEtran}

\usepackage{times}
\usepackage{epsfig}
\usepackage{graphicx}
\usepackage{amsmath}
\usepackage{amssymb}
\usepackage{subfigure}
\usepackage{xspace}
\usepackage{color}

\usepackage{algorithmic}
\usepackage{algorithm}

\newcommand{\norm}[1]{\left\Vert#1\right\Vert}
\newcommand{\abs}[1]{\left\vert#1\right\vert}

\newcommand\vect[1]{{\bf#1}}
\newcommand\matr[1]{{\bf#1}}

\newcommand\alphabf{{\boldsymbol{\alpha}}}
\newcommand{\argmin}{\operatornamewithlimits{argmin}}

\newcommand{\Real}{\mathbb R}
\newcommand{\Natural}{\mathbb N}
\newcommand\RR[1]{\mathbb{R}^{#1}}
\DeclareRobustCommand\onedot{\futurelet\@let@token\@onedot}
\def\@onedot{\ifx\@let@token.\else.\null\fi\xspace}

\newcommand{\rg}[1]{\textcolor{black}{#1}}

\begin{document}

\title{Sparsity Based Poisson Denoising with Dictionary Learning}

\author{Raja Giryes and Michael Elad\\
Department of Computer Science, The Technion - Israel Institute of Technology\\
Haifa, 32000, Israel\\
{\tt\small \{raja,elad\}@cs.technion.ac.il}
}

\maketitle

\begin{abstract}
The problem of Poisson denoising appears in various imaging applications,
such as low-light photography, medical imaging and microscopy.
In cases of high SNR, several transformations exist so as to convert
the Poisson noise into an additive i.i.d. Gaussian noise, for which many effective algorithms are available.
However, in a low SNR regime, these transformations are significantly less accurate,
and a strategy that relies directly on the true noise statistics is required.
A recent work by Salmon et al. \cite{Salmon12PoissonConf, Salmon12Poisson} took this route, proposing a patch-based exponential
image representation model based on GMM (Gaussian mixture model), leading to state-of-the-art results.
In this paper, we propose to harness sparse-representation modeling to the image patches, adopting the same exponential idea.
Our scheme uses a greedy pursuit with boot-strapping based stopping condition and dictionary learning within the denoising process.
The reconstruction performance of the proposed scheme is competitive with leading methods in high SNR,
and achieving state-of-the-art results in cases of low SNR.
\end{abstract}

\section{Introduction}

Poisson noise appears in many applications such as
night vision, computed tomography (CT), fluorescence microscopy, astrophysics
and spectral imaging. Given a Poisson noisy image $\vect{y} \in \left(\Natural \cup \left\{0 \right\}\right) ^m$ (represented as a column-stacked vector),
our task is to recover the original true image $\vect{x} \in \Real^m$,
where the entries in $\vect{y}$ (given $\vect{x}$) are Poisson distributed independent random variables with mean and variance $\vect{x}[i]$, i.e.,
\begin{eqnarray}
\label{eq:pois_dist}
P(\vect{y}[i]\big|\vect{x}[i]) = \left\{ \begin{array}{cc}
                               \frac{(\vect{x}[i])^{\vect{y}[i]}}{\vect{y}[i]!}\exp(-\vect{x}[i]) & \vect{x}[i] >0 \\
                               \delta_{0}(\vect{y}[i]) & \vect{x}[i] = 0,
                             \end{array} \right.
\end{eqnarray}
where $\delta_{0}$ is the Kronecker delta function and $\vect{x}[i]$ and $\vect{y}[i]$ are the $i$-th component in $\vect{x}$ and $\vect{y}$ respectively. Notice that Poisson noise is not additive
and its strength is dependent on the image intensity. Lower intensity in the image yields a stronger noise as
the SNR in each pixel is $\sqrt{\vect{x}[i]}$. Thus, it is natural to define the noise power in an image by the maximal value in $\vect{x}$ (its peak value)\footnote{The peak is a good measure under the assumption that the pixels' values are spread uniformly over the whole dynamic range. This assumption it true for most natural images.}.

Many schemes for recovering $\vect{x}$ from $\vect{y}$ exist \cite{Danielyan11Deblurring, Rodrigo11Efficient, Figueiredo10Restoration, Zhang12Novel}. 
A very popular strategy \cite{Boulanger10Patch, Makitalo11Optimal, Zhang08Wavelets} relies on transformations,
such as Anscombe \cite{anscombe48transformation} and Fisz \cite{Fisz55Limiting},
that convert the Poisson denoising problem into a Gaussian one, for which plenty of methods exist (e.g. \cite{Dabov07BM3D,Mairal09Non,Yu12Solving}).
The noise becomes approximately white  Gaussian with unit variance.

\begin{figure}[htb]
\centering
{\subfigure{\includegraphics[width=0.44\linewidth]{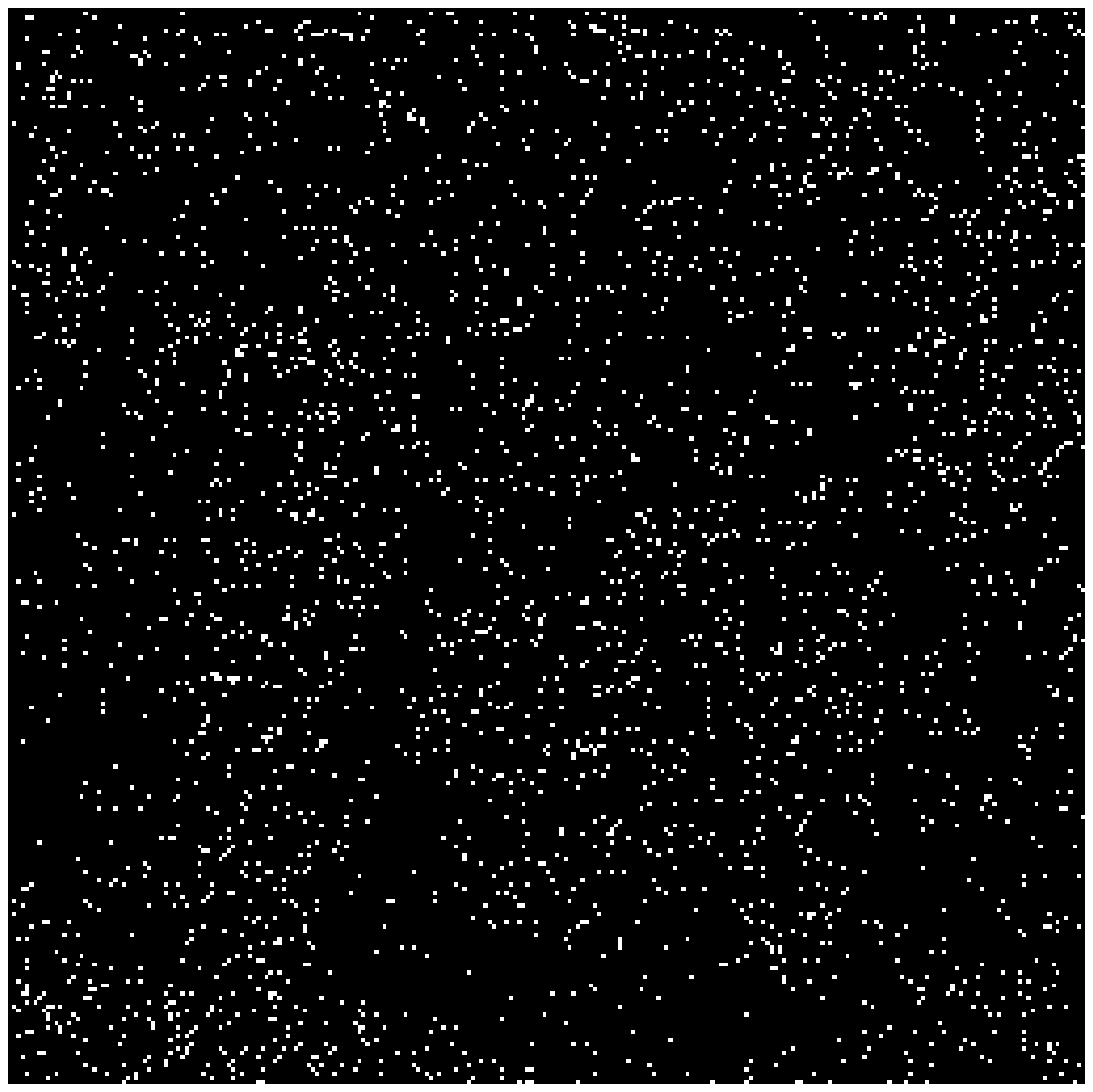}}%
\hfil
\subfigure{\includegraphics[width=0.44\linewidth]{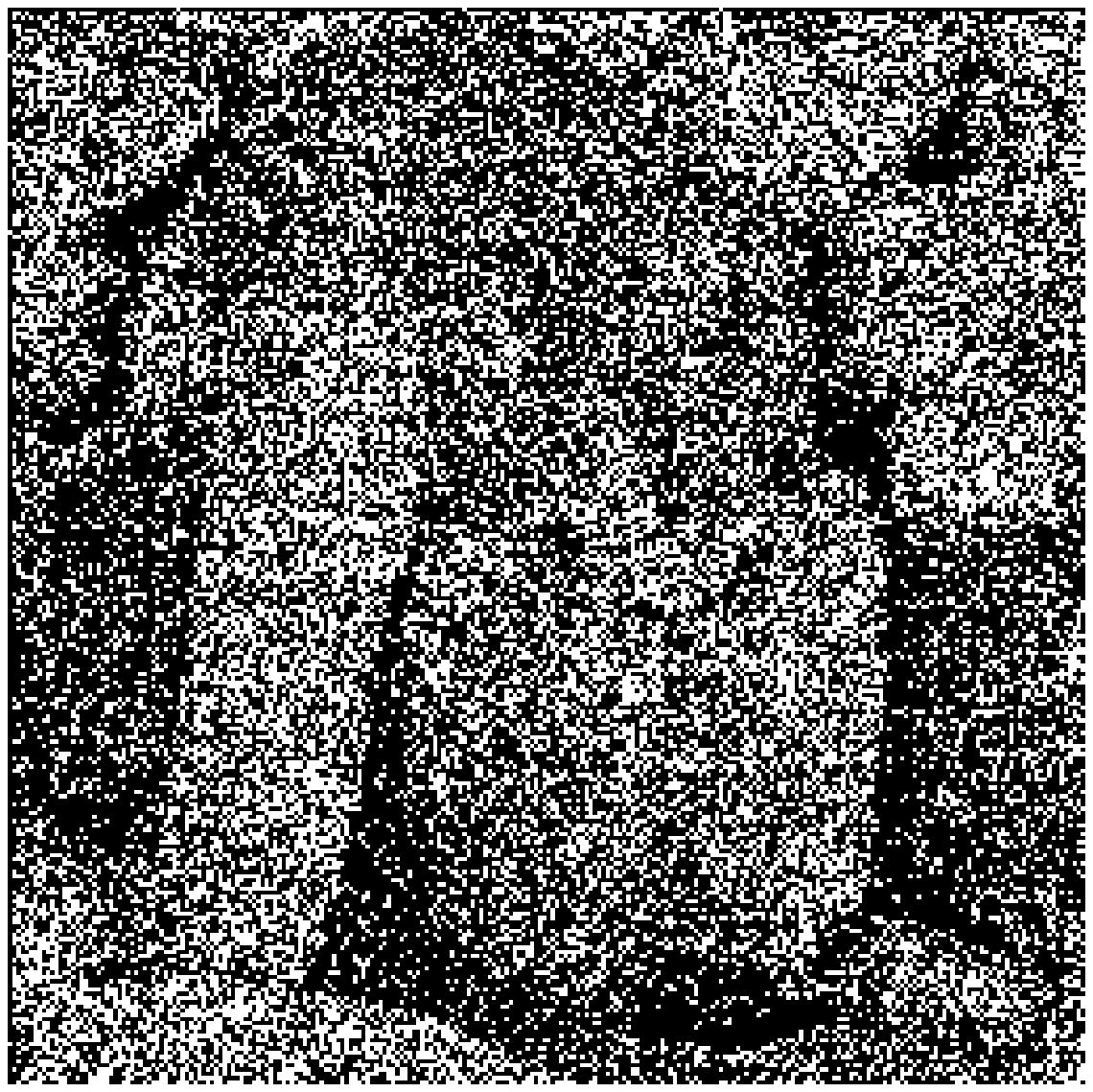}}
\hfil
\subfigure{\includegraphics[width=0.44\linewidth]{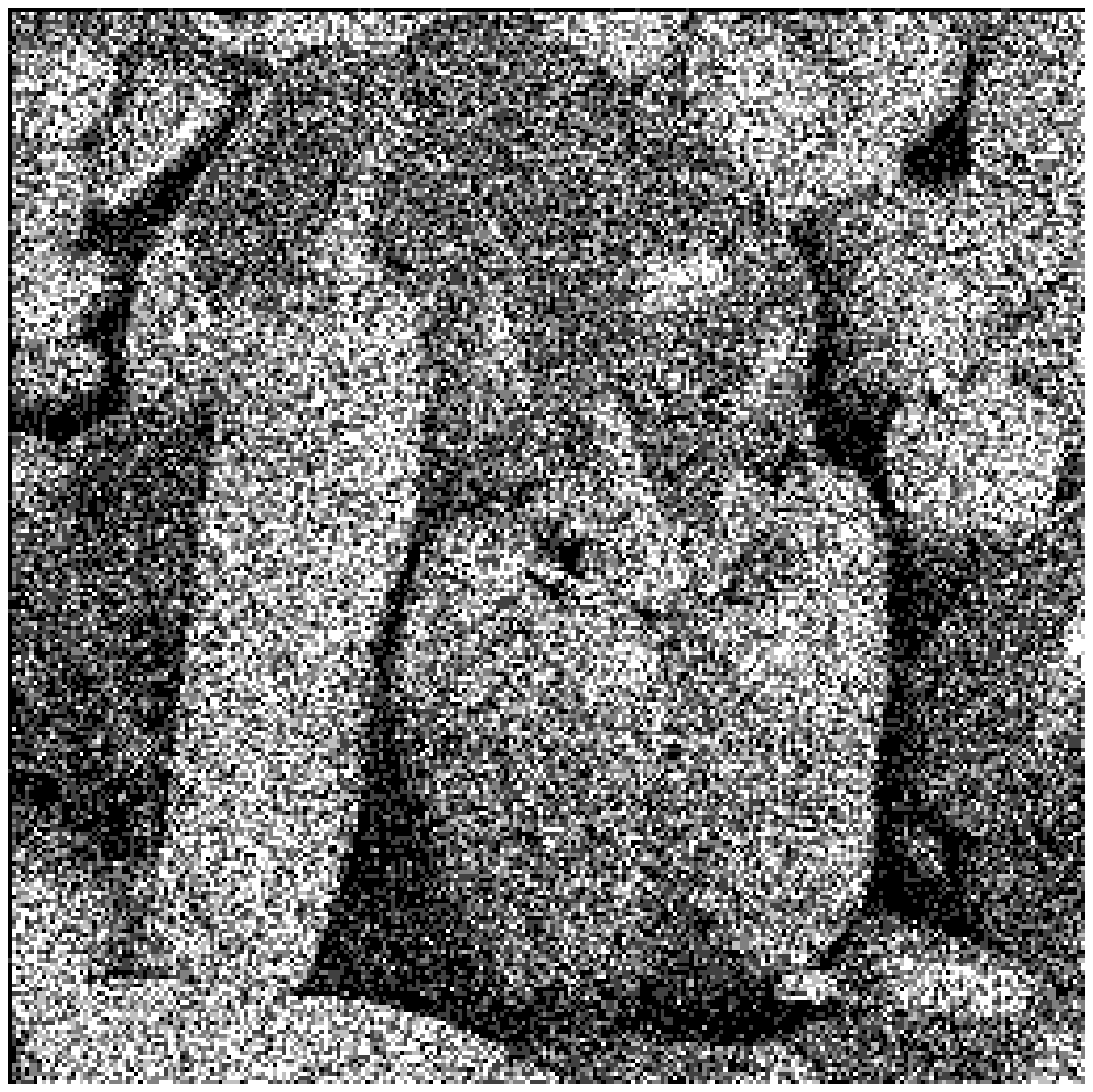}}
\hfil
\subfigure{\includegraphics[width=0.44\linewidth]{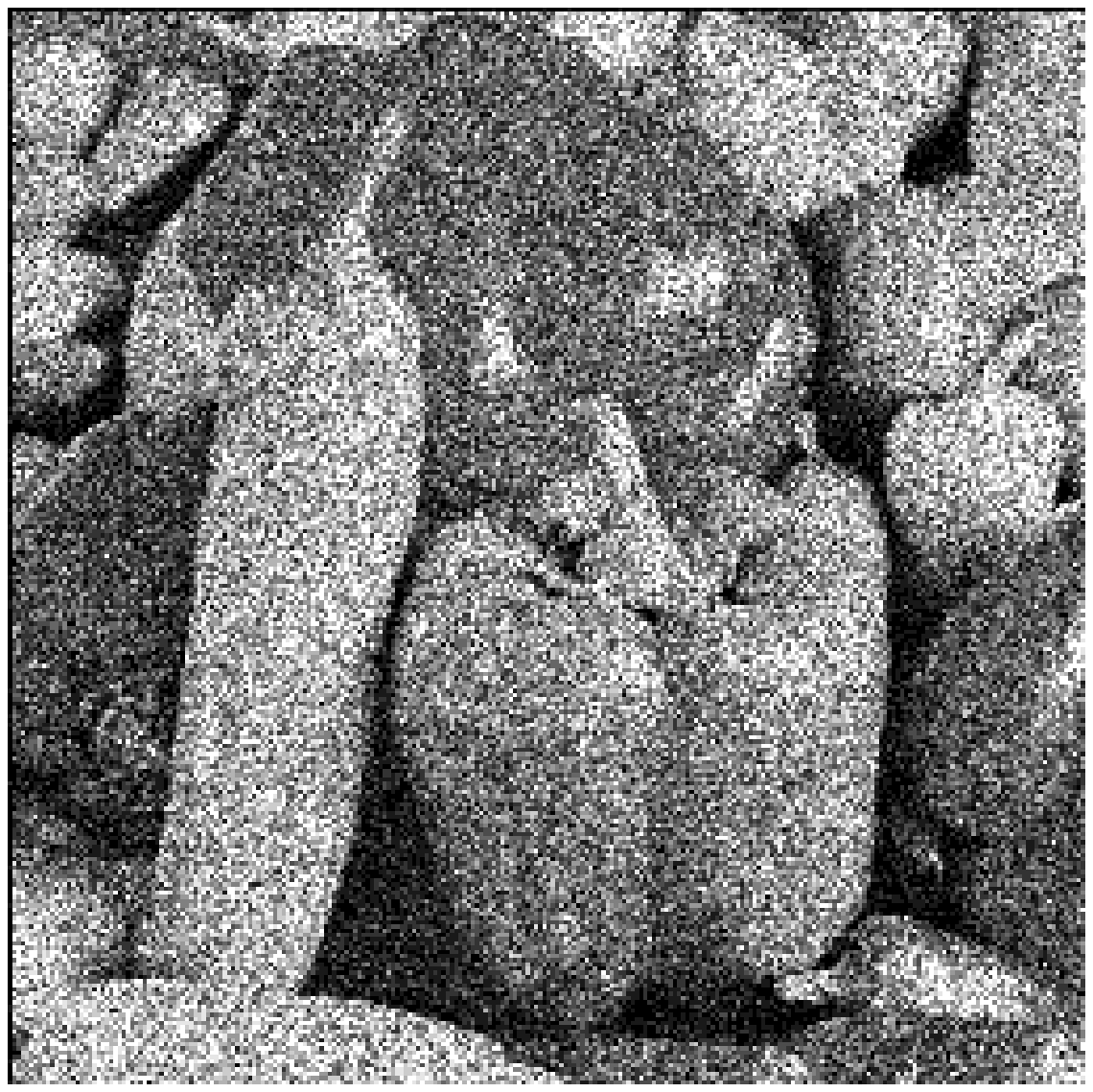}}
}%
\caption{Poisson noisy versions of the image {\em peppers} with different peak values.
From left to right: Peaks $0.1$, $1$, $4$ and $10$.}
\label{fig:peppers_noisy_images}
\end{figure}

The problem with these approximations is the fact that they hold true only
when the measured pixels have high intensity \cite{Makitalo11Optimal, Salmon12PoissonConf,Salmon12Poisson},
 i.e. when a high photon count is  measured in the detectors.
As a thumb rule, these transformations are accurate only when the peak value in $\vect{x}$ is larger than $3$ \cite{Salmon12Poisson}\footnote{One may argue that the threshold value is $2$ as a conclusion from the denoising results in 
\cite{Salmon12PoissonConf}, so this threshold value is by no means an exact and definite one.}.
In this case the noise looks very similar to a Gaussian one.
When the peak value is smaller, the structure of the noisy image is quite different,
with many zero pixels and others that have very small (integer) values.
As an example, when the peak equals $0.1$ we have almost a binary image, containing mainly either zeros or ones.
Fig~\ref{fig:peppers_noisy_images} shows noisy versions of {\em peppers}
with different peak values. It can be seen that indeed, as the peak value increases, the noise
"looks" more and more like Gaussian.
In this work we aim at denoising Poisson noisy images with peak$\le 4$ \rg{where both Anscombe and Fitz transformations are} less effective.


The Anscombe transform is a non-linear element-wise transformation  defined as
\begin{eqnarray}
f_{Anscombe} \left(\vect{y}[i] \right) = 2 \sqrt{\vect{y}[i] + \frac{3}{8}}.
\end{eqnarray}
Therefore, applying a denoising technique on the stablized data results with an estimate for $f_{Anscombe} \left(\vect{x} \right)$ rather than for $\vect{x}$. Thus, there is a need to apply an inverse transform in order to get an estimate for $\vect{x}$. 
Note that using the algebraic inverse results with a biased estimator
due to the non-linearity of the transform. 
In \cite{Makitalo11Optimal}, this problem has been addressed by providing an exact unbiased inverse for the Anscombe transform,
which eventually leads to a better recovery performance.

However, as said above, even with this exact inverse the recovery error dramatically increases for peak$<3$ \cite{Salmon12Poisson}.
In order to deal with this deficiency, a strategy that relies directly on the Poisson statistics is required.
This direction has been taken in \cite{Salmon12PoissonConf, Salmon12Poisson},
providing a Gaussian mixture model (GMM) \cite{Yu12Solving}
based approach that relies directly on the Poisson noise properties. By dividing the image into overlapping patches,
dividing them to few large clusters and then performing a projection
onto the largest components of a PCA-like basis for each group,
state-of-the-art results have been reported for small peak values.
This approach has two versions.
In the first, the non-local PCA (NLPCA), the projection is computed by  minimizing a Poissonian Bregman divergence using Newton steps,
while in the second, the non-local sparse PCA (NLSPCA), the SPIRAL method \cite{Harmany12SPIRAL} that adds an $\ell_1$ regularization term to the minimized objective is used resulting with a better recovery performance.

\subsection{Our Contribution}

In this work we take a similar path as \cite{Salmon12Poisson}
in the image patches modeling, using this for the recovery process.
However, we take a different route and propose a sparse representation
modeling and a dictionary learning based denoising strategy \cite{Mairal09Non} instead of the GMM.
We employ a greedy OMP-like method for sparse coding of the patches
and use a smart boot-strapped stopping criterion.
We demonstrate the superiority of the proposed scheme in various experiments.

We have presented a preliminary version of the proposed OMP-like technique 
that achieves a relatively poor recovery performance in a local conference \cite{Giryes12Sparsity}. In this paper, 
both the algorithmic and the experimental parts have been remarkably improved.

The main contributions of this paper are:
\begin{itemize}
\item We introduce a greedy technique for the Poisson sparse model.  Such pursuit methods for a Gaussian noise are commonly used, and extensive work has been devoted in the past two decades to their construction and analysis.  Thus, a proposal of a greedy strategy for the Poisson model is of importance and may open the door for many other variants and theoretical study, similar to what exists in the Gaussian regime. For example, our greedy method can be extended for the Poisson deconvolution problem as done for other greedy techniques in \cite{Dupe13greedy, Dinh14Composite} or serve as a basis for Poisson inpainting \cite{Giryes14Inpainting}.

We mention that in some low light Poisson denoising applications the dictionary is already known. One example is Fluorescence microscopy where the \rg{measured image might be sparse by itself}. Hence, in such cases the introduction of a new recovery method by itself is important.  
\item	We introduce a novel stopping criteria for iterative algorithm, and its incorporation into the pursuit leads to much improved results. To the best of our knowledge, this boot strapping based stopping criterion first appears in our paper. 
\item The interplay between GMM and dictionary learning based models has significance of its own, as seen in the treatment of the simpler Gaussian image denoising problem. The migration from GMM to dictionary learning poses series of difficulties that our paper describes and solves.  Note that in this paper we utilize the learning strategy in \cite{Smith13Improving}.
\item The integration of the above leads to state-of-the-art results, especially for the very low SNR case.

\end{itemize}

\subsection{Organization}

The organization of the paper is as follows.
Section~\ref{sec:background} describes the Poisson denoising problem
with more details, and presents the previous contributions.
Section~\ref{sec:alg} introduces the proposed denoising algorithm,
starting with the pursuit task, moving to the
clustering that we employ for achieving non-locality in the denoising process,
discussing the role of learning the dictionary,
and concluding with the overall scheme.
Section~\ref{sec:exp} presents various tests and comparisons that demonstrate the denoising performance and superiority of the proposed scheme.
Section~\ref{sec:conc} discusses future work directions and conclusions.

\section{Poisson Sparsity Model}
\label{sec:background}

Various image processing applications use the fact that image patches can be represented sparsely
with a given dictionary $\matr{D} \in \RR{d \times n}$  \cite{Elad10Sparse}.
Under this assumption each patch $\vect{p}_i \in \RR{d}$ \rg{(the patch is of size $\sqrt{d} \times \sqrt{d}$ held in a column stack representation)} from $\vect{x}$ can be represented as $\vect{p}_i = \matr{D}\alphabf_i$, where
$\alphabf_i \in \RR{n}$ is $k$-sparse, i.e., it has only $k$ non-zero entries, where $k\ll d$. \rg{If our image is of dimension $m_1 \times m_2$ and we treat all the overlapping patches in the image then $1\le i \le (m_1-\sqrt{d}+1)(m_2-\sqrt{d}+1)$.}

This model leads to state-of-the-art results in Gaussian denoising \cite{Dabov07BM3D, Mairal09Non, Elad10Sparse}.
In order to use this sparsity-inspired model for the Poisson noise case one has two options:
(i) convert the Poisson noise into a Gaussian one, as done in \cite{Makitalo11Optimal};
or (ii) adapt the Gaussian denoising tools to the Poisson statistics.
As explained above, the later is important for cases where the Anscombe is non-effective,
and this approach is indeed practiced in \cite{ Salmon12PoissonConf,Salmon12Poisson, Harmany12SPIRAL, Lingenfelter09Sparsity}.
Maximizing of the log-likelihood of the Poisson distribution in \eqref{eq:pois_dist} provides us with the following minimization problem
for recovering the $i$-th patch $\vect{p}_i$ from its corresponding Poisson noisy patch $\vect{q}_i$,
\begin{eqnarray}
\label{eq:pois_obj}
\min_{\vect{p}_i} \vect{1}^*\vect{p}_i - \vect{q}_i^*\log(\vect{p}_i) & s.t. &  \vect{p}_i \ge 0.
\end{eqnarray}
where $\vect{1}$ is a vector composed of ones and the $\log(\cdot)$ operation is applied element-wise.
Note that this minimization problem allows zero entries in $\vect{p}_i$ only if the corresponding entries in $\vect{q}_i$ are zeros as well.
The minimizer of \eqref{eq:pois_obj} is the noisy patch $\vect{q}_i$ itself, and thus using only \eqref{eq:pois_obj} is not enough
and a prior is needed.
By using the standard sparsity prior for patches as practiced in \cite{Mairal09Non} and elsewhere, $\vect{p}_i=\matr{D}\alphabf_i$, we end up with the following minimization problem:
\begin{eqnarray}
\label{eq:pois_obj_reg_standard}
&& \hspace{-1.2in} \min_{\alphabf_i} ~ \vect{1}^*\matr{D}\alphabf_i - \vect{q}_i^*\log(\matr{D}\alphabf_i) \\ \nonumber & s.t. & \norm{\alphabf_i}_0 \le k, \matr{D}\alphabf_i \ge 0,
\end{eqnarray}
where $\norm{\cdot}_0$ is the $\ell_0$ semi-norm which counts the number of non-zeros in a vector.
Besides the fact that \eqref{eq:pois_obj_reg_standard} is a combinatorial problem, it also imposes a non-negativity constraint
on the recovered patch, which further complicates the numerical task at hand.
In order to resolve the latter issue we follow \cite{Salmon12Poisson} and set $\vect{p}_i = \exp(\matr{D}\alphabf_i)$
where $\exp(\cdot)$ is applied element-wise and $\alphabf_i$ is still a $k$-sparse vector. This leads to the following minimization problem,
\begin{eqnarray}
\label{eq:pois_sparsity_obj}
\min_{\alphabf_i} \vect{1}^*\exp(\matr{D}\alphabf_i) - \vect{q}_i^*\matr{D}\alphabf_i &s.t.& \norm{\alphabf_i}_0 \le k.
\end{eqnarray}

Having the non-negativity constraint removed,
we still need to have an approximation algorithm for solving \eqref{eq:pois_sparsity_obj} as it is likely to be NP-hard.
One option is to use an $\ell_1$ relaxation, which leads to the SPIRAL method \cite{Harmany12SPIRAL}.
Another, simpler option is to reduce the dictionary $\matr{D}$ to have only $k$ columns and thus \eqref{eq:pois_sparsity_obj}
can be minimized with any standard optimization toolbox for convex optimization\footnote{For a given support (location of the non-zeros in $\alphabf$), the problem posed in \eqref{eq:pois_sparsity_obj} is convex.}.
This approach is taken in the NLPCA technique \cite{Salmon12PoissonConf} where the patches of $\vect{y}$ are clustered into
a small number of disjoint groups and each group has its own narrow dictionary.
Denoting by $j$, $\{\vect{q}_{j,1},\dots, \vect{q}_{j,N_j}\}$, $\{\alphabf_{j,1},\dots, \alphabf_{j,N_j}\}$ and $\matr{D}_j \in \RR{d \times k}$ the group number, its noisy patches, the representations of the patches and their dictionary, the
minimization problem NLPCA aims at solving for the $j$-th group is
\begin{eqnarray}
\label{eq:NLPCA_sparsity_obj}
\min_{\matr{D}_j,{\alphabf}_{j,1},\dots,{\alphabf}_{j,N_j}} \sum_{i=1}^{N_j} \vect{1}^*\exp(\matr{D}_j\alphabf_{j,i}) - \vect{q}_{j,i}^*\matr{D}_j\alphabf_{j,i}.
\end{eqnarray}
Notice that $\matr{D}_j$ is calculated also as part of the minimization process as each group has its own dictionary that should be optimized.
The typical sparsity $k$ and number of clusters used in NLPCA are $4$ and $14$ respectively.
As it is hard to minimize \eqref{eq:NLPCA_sparsity_obj} both with respect to the dictionary and to the representations,
an alternating minimization process is used by applying Newton steps updating the dictionary and the representations alternately.
Having the recovered patches, we return each to its corresponding location in the recovered image
and average pixels that are mapped to the same place (since overlapping patches are used).
The authors in \cite{Salmon12PoissonConf} suggest to repeat the whole process,
using the output of the algorithm as an input for the clustering process and then applying the algorithm again
with the new division.
An improved version of NLPCA, the NLSPCA, is proposed in \cite{Salmon12Poisson},
replacing the Newton step with the SPIRAL algorithm \cite{Harmany12SPIRAL} for calculating the patches' representations.

The NLPCA for Poisson denoising is based on ideas related to the GMM model developed for the Gaussian denoising case \cite{Yu12Solving}.
Of course, this method can be used for the Poisson noise by applying an Anscombe transform.
However, such an approach is shown in \cite{Salmon12Poisson} to be inferior to the Poisson model based strategy in the low-photon count case.

Like \cite{Salmon12PoissonConf, Salmon12Poisson}, in this work we do not use the Anscombe and rely on the Poisson based model.
However, as opposed to \cite{Salmon12PoissonConf, Salmon12Poisson}, we use
one global dictionary for all patches and propose a greedy algorithm for finding the representation of each patch.
In an approach similar to the one advocated in \cite{Mairal09Non},
we treat similar patches jointly by forcing them to share the same atoms in their representations.
This introduces a non-local force and sharing of information between different regions in the image.
A dictionary learning process is utilized in our scheme as well,
in order to improve the initial dictionary used for the data,
as the initial dictionary we embark from is global for all images.

Before moving to the next section we mention a technique proposed in \cite{Salmon12Poisson} to enhance the SNR in noisy images.
This method is effective especially in very low SNR scenarios such as peak smaller than $0.5$.
Instead of denoising the given image directly, one can downsample the image by applying a low-pass filter followed by down-sampling.
This provides us with a smaller image but with a higher SNR. For example, if our low-pass filter is a kernel of size $3 \times 3$ containing ones,
and we sample every third row and every third column, we end up with a nine times smaller image that has a nine times larger peak value.
Having the low-res noisy image, one may apply on it any Poisson denoising algorithm
and then perform an upscaling interpolation on the recovered small image
in order to return to the original dimensions.
This method is referred to as binning in \cite{Salmon12Poisson} and related to multi-scale programs \cite{Mester11Improving}.
Note that this technique is especially effective for the Anscombe based techniques as the peak value
of the processed image is larger than the initial value.

\section{Sparse Poisson Denoising Algorithm (SPDA)}
\label{sec:alg}

\begin{algorithm}[t]
\caption{Patch Grouping Algorithm} \label{alg:patch_group}
\begin{algorithmic}

\REQUIRE $\vect{y}$ is a given image, $\sqrt{d} \times \sqrt{d}$ is the patch size to use, $l$ is a target group size,
$h$ is a given convolution kernel (typically a wide Gaussian) and $\epsilon$ is a tolerance factor.

\ENSURE Division to disjoint groups of patches, \rg{where the $g$-th group is of size $N_g \ge l$ and is $Q_g = \{\vect{q}_{g,1},\dots, \vect{q}_{g,N_g}\}$}.

\STATE Begin Algorithm:

\STATE -Convolve the image with the kernel: $\tilde{\vect{y}} = \vect{y} \ast h$. We take $\tilde{\vect{y}}$ to be of the same size of $\vect{y}$.

\STATE -Extract all overlapping patches $Q = \{\vect{q}_1,\dots,\vect{q}_N \}$ of size $\sqrt{d} \times \sqrt{d}$ from $\vect{y}$
and their corresponding patches $\{\tilde{\vect{q}}_1,\dots,\tilde{\vect{q}}_N \}$ of size $\sqrt{d} \times \sqrt{d}$ from $\tilde{\vect{y}}$.

\STATE -Set first group pivot index: $s_0 = \argmin_{1 \le i \le N}\norm{\tilde{\vect{q}}_i}_2^2$.

\STATE -Initialize $g = 0$ and ${i}_g^{\text{prev}} =s_0$.

\WHILE{$Q \ne \emptyset$}

\STATE -Initialize group $g$: $Q_g = \emptyset$ and $l_g = 0$.

\STATE -Select first candidate: $i_g = \argmin_{i \rg{: \vect{q}_i \in Q}}\norm{\tilde{\vect{q}}_{s_g}- \tilde{\vect{q}}_i}_2^2$

\WHILE{$\left(l_g \le l \text{ or } \abs{\norm{\tilde{\vect{q}}_{s_g}- \tilde{\vect{q}}_{i_g}}_2^2 - \norm{\tilde{\vect{q}}_{s_g}- \tilde{\vect{q}}_{{i}_g^{\text{prev}}}}_2^2} \le \epsilon^2\right)$
and $Q \ne \emptyset$}

\STATE -Add patch to group $j$: $Q_g = Q_g \cup \{\vect{q}_{i_g} \}$, $l_g = l_g + 1$.

\STATE -Exclude patch from search: $Q = Q \setminus \{\vect{q}_{i_g} \}$.

\STATE -Save previous selection: ${i}_g^{\text{prev}} = i_g$.

\STATE -Select new candidate: $i_g = \argmin_{i}\norm{\tilde{\vect{q}}_{s_g}- \tilde{\vect{q}}_i}_2^2$.

\ENDWHILE

\STATE -Set pivot index for next group: $g = g+1$, $s_g = i_{g-1}$.

\ENDWHILE
\STATE -Merge the last group with the previous one to ensure its size to be bigger than $l$.

\end{algorithmic}
\end{algorithm}

\begin{algorithm}[t]
\caption{Poisson Greedy Algorithm} \label{alg:pga}
\begin{algorithmic}

\REQUIRE $k, \matr{D} \in \RR{d\times n}, \{\vect{q}_1,\dots,\vect{q}_{\tilde{l}} \}$ where $\vect{q}_i \in \RR{d}$ is a Poisson distributed vector
with mean and variance approximated (modeled) by $\exp(\matr{D}\alphabf_i)$, and $k$ is the maximal cardinality of $\alphabf_i$.
All representations $\alphabf_i$ are assumed to have the same support.
Optional parameter: Estimates of the true image patches $\{\vect{p}_1,\dots,\vect{p}_{\tilde{l}} \}$.

\ENSURE \rg{Estimates $\hat{\vect{p}}_i = \exp(\matr{D}\hat\alphabf_i)$  for  $\vect{q}_i$, $i=1 \dots \tilde{l}$.}

\STATE Begin Algorithm:

\STATE -Initialize the support $T^0 =\emptyset$ and set $t = 0$.

\WHILE{$t < k$}

\STATE -Update iteration counter: $t = t + 1$.

\STATE -Set initial objective value: $v_{o} = \inf$.

\FOR{$j=1:n$}

\STATE -Check atom $j$: $\tilde{T}^t = T^{t -1} \cup \{j\}$.

\STATE \flushleft{-Calculate current objective value: $v_{c} =$} $\min_{\tilde\alphabf_1,\dots,\tilde\alphabf_{\tilde{l}}} \sum_{i=1}^{\tilde{l}} \vect{1}^*\exp(\matr{D}_{\tilde{T}^{t}}\tilde\alphabf_i) - \vect{q}_{i}^*\matr{D}_{\tilde{T}^{t}}\tilde\alphabf_i$

\IF {$v_o > v_c$}

\STATE -Update selection: $j^t = j$ and $v_o = v_c$.

\ENDIF

\ENDFOR

\STATE -Update the support: $T^t = T^{t -1} \cup \{j^t\}$.

\STATE \flushleft{-\rg{Update representation estimate:}}
\STATE ~~~ -\rg{Set} $[\hat\alphabf^t_1,\dots, \hat\alphabf^t_{\tilde{l}}] =$ $[\vect{0}, \dots, \vect{0}].$ 
\STATE ~~~ -\rg{Update on-support values:} $[(\hat\alphabf^t_1)_{T^t},\dots, (\hat\alphabf^t_{\tilde{l}})_{T^t}] =$ 
\STATE ~~~~~$\argmin_{\tilde\alphabf_1,\dots,\tilde\alphabf_{\tilde{l}}} \sum_{i=1}^{\tilde{l}} \vect{1}^*\exp(\matr{D}_{T^{t}}\tilde\alphabf_i) - \vect{q}_{i}^*\matr{D}_{T^{t} }\tilde\alphabf_i$.

\IF {$\{\vect{p}_1,\dots,\vect{p}_{\tilde{l}} \}$ are given}

\STATE -Estimate error: $e_t = \sum_{i=1}^{\tilde{l}}\norm{\exp(\matr{D}\hat\alphabf^t_i) - \vect{p}_i}_2^2$.

\IF{$t>1$ and $e_t > e_{t-1}$ }

\STATE -Set $t = t -1$ and break (exit while and return the result of the previous iteration).

\ENDIF

\ENDIF

\ENDWHILE
\STATE -Form the final estimate $\hat{\vect{p}}_i = \exp(\matr{D}\hat\alphabf^t_i), 1 \le i \le \tilde{l}$.

\end{algorithmic}
\end{algorithm}

Our denoising strategy is based on a dictionary learning based approach.
We start by extracting a set of overlapping patches $\{ \vect{q}_i | 1\le i \le (m_1-\sqrt{d}+1)(m_2-\sqrt{d}+1)\}$
($m_1$ and $m_2$ are the vertical and horizontal dimensions of the noisy image $\vect{y}$ respectively)
of size $\sqrt{d} \times \sqrt{d}$ from the noisy image $\vect{y}$.
The goal is to find the dictionary $\matr{D}$ that
leads to the sparsest representation of this set of patches
under the exponential formulation.
In other words our target is to minimize
\begin{eqnarray}
\label{eq:SPDA_sparsity_obj}
&& \hspace{-1.2in}\min_{\matr{D},{\alphabf}_{1},\dots,{\alphabf}_{N}} \sum_{i=1}^{N} \vect{1}^*\exp(\matr{D}\alphabf_{i}) - \vect{q}_i^*\matr{D}\alphabf_{i}
\\ \nonumber & s.t. & \norm{\alphabf_i}_0 \le k, 1\le i \le N,
\end{eqnarray}
\rg{where $N = (m_1 - \sqrt{d} + 1)(m_2 - \sqrt{d} + 1)$}. 
As in \cite{Salmon12PoissonConf, Salmon12Poisson}, since minimizing both with respect to the dictionary
and the representations at the same time is a hard problem,
we minimize this function alternately.
The pursuit (updating the representations) is performed using a greedy technique which returns a $k$-sparse representation for each patch,
unlike the global Newton step or SPIRAL which is not guaranteed to have a sparse output. For learning the dictionary we use the technique in \cite{Smith13Improving} that updates the dictionary together with the representations while their supports are kept fixed \cite{Salmon12Poisson}.
In order to further boost the performance of our algorithm and exploit the fact that similar patches
in the image domain may \rg{have the same support in their sparse representation},
the patches are clustered into a large number
of small disjoint groups of similar patches.
We turn  to describe in details each step of the algorithm.

\subsection{Patch Clustering}

Ideally, as Poisson noisy images with very low SNR are almost binary images,
a good criterion for measuring the similarity between patches would be the earth mover's distance (EMD).
We approximate this measure by setting the distance between patches
to be their Euclidean distance in the image after it passes through a Gaussian filter. It is clear that the Euclidean distance is not the only option, nor the best. Nevertheless, the reason we have selected the $\ell_2$ distance is that it gives a bigger weight for entries where we find noisy pixels with large values. Those are usually rare and reflect locations of high intensity in the original image. We are expecting that patches with high intensity, which are similar in the original image, should have concentrations of high photons counts at the same locations.

The grouping algorithm is described in details in Algorithm~\ref{alg:patch_group}.
It creates disjoint groups of size (at least) $l$ in a sequential way, adding elements one by one.
Once a group gets to the destination size, the algorithm continues to add
elements whose distance from the first element in the group (the pivot) is up to
$\epsilon$ away from the distance of the last added element (See Algorithm~\ref{alg:patch_group}).

The reason we have selected to use this strategy for clustering is due to its ability to divide the patches to groups of similar size. The reason we target similar sizes is that we want to guarantee that we have ``enough'' atoms in each group for selecting the ``correct'' support for the group in the sparse coding step (described in the next subsection). On the other hand, we need to have as many groups as possible since otherwise we will not have enough information for updating the dictionary, as we select the same support for all the patches in the same group.
We could have selected a group for each atom separately, which would result with overlapping groups, but we have chosen not to do so due to computational reasons. This is also exactly the reason why we have chosen to use the current clustering method and not other off-the-shelf methods as its greedy nature seems to make it faster. We should note that we are well aware of the fact that our choice of grouping method is suboptimal, and yet, as the results in Section~\ref{sec:exp} indicate, it is sufficiently good-performing.

\subsection{Sparse Coding}

For calculating the representations we use a joint sparsity assumption for each group of patches
with a greedy algorithm that finds the representations of each group together.
This algorithm is iterative and in each iteration it adds the atom
that reduces the most the cost function \eqref{eq:pois_sparsity_obj} for all the representations that belong to the same group.
\rg{We further explain this matter when we discuss Equation \eqref{eq:pois_sparsity_obj_fixed_T_iter_j}.}

An important aspect in our pursuit algorithm is to decide how many atoms to associate with each
patch, i.e., what should be the stopping criterion of the algorithm.
We employ two options. The first is to run the algorithm with a constant number of iterations.
However, this choice leads to a suboptimal denoising effect
as different patches contain different content-complexity and thus require different sparsity levels.

{\em Bootstrapping Based Stopping Criterion:} Another option is to set a different cardinality for each group.
In order to do so we need a way to evaluate the error in each iteration with respect to the true image and then stop
the iterations once the error starts increasing.
One option for estimating the error is using the Poission unbiased risk  estimator (PURE), a variant of the Stein's unbiased risk estimator (SURE) for Poisson noise,  as done in \cite{Deledalle10Poisson} for the NL-Means algorithm.
However, in our context the computation of the PURE is too demanding, as it requires re-applying the denoising method over and over again for the same image by changing one pixel each time. In \cite{Deledalle10Poisson}, this is done efficiently due to the simple structure of the denoising method (NL-Means) used there. In our case this (as far as we can see) cannot be done. 

Thus, we use boot-strapping -- we rely on the fact that our scheme is iterative and after each step of representation decoding
and dictionary update we have a new estimate of the original image.
We use the patches of the  reconstructed image from the previous iteration
as a proxy for the patches of the true image and compute the error with respect to them.
One might think that if the patch from the previous iteration is used to determine the number of iterations for the same patch in the current iteration, it will make no change and we will ``get stuck'' with the same outcome. However, as an averaging process is applied in the middle together with a dictionary update step, we are not expected to get back the same patch but rather a different denoising result.

Note that since we update the dictionary between iterations
and average the patches in the recovered image, we dot not get stuck on the same patch again by using this stopping criteria.
In practice, this condition improves the recovery performance significantly
as each group of patches is represented with a cardinality that suites its content better.

The greedy Poisson algorithm is summarized in Algorithm~\ref{alg:pga}.
It starts with an empty support $T^0$ and adds one element to it gradually in each iteration. Given the support $T^{t-1}$ of the previous iteration, the next added atom to the support is chosen by iterating over all the atoms in the dictionary and calculating for each the following minimization problem:
\begin{eqnarray}
\label{eq:pois_sparsity_obj_fixed_T_iter_j}
\min_{\tilde\alphabf_1,\dots,\tilde\alphabf_l} \sum_{i=1}^{\tilde{l}} \vect{1}^*\exp(\matr{D}_{{T}^{t-1} \cup \left\{j \right\}}\tilde\alphabf_i) - \vect{q}_{i}^*\matr{D}_{{T}^{t-1} \cup \left\{j \right\}}\tilde\alphabf_i,
\end{eqnarray}
where $j$ is the index of a potential atom for addition, $\left\{\vect{q}_i \right\}_{i=1}^{\tilde{l}}$ are the patches we decode, $\left\{\alphabf_i\right\}_{i=1}^{\tilde{l}}$ are representations of size $t$ (cardinality in the current iteration) and $\matr{D}_{{T}^{t-1} \cup \left\{j \right\}}$ is  $\matr{D}$ restricted to the support ${{T}^{t-1} \cup \left\{j \right\}}$ \rg{(In a similar way $(\hat\alphabf^t_i)_{T^t}$ is the vector $\hat\alphabf^t_i$ restricted to the entries supported on $T^t$)}. Notice that all the patches in the cluster are enforced with the same sparsity pattern  as we perform the minimization for all the $\tilde{l}$ patches together and select the same atom for all.
Notice that the problem in \eqref{eq:pois_sparsity_obj_fixed_T_iter_j} is the same as the one in \eqref{eq:pois_sparsity_obj} but for a fixed given support.
When the support is fixed it is a convex problem,
which appears also in the NLPCA technique, and can be solved by the Newton method or by any convex optimization toolbox.

The bootstrapping based stopping criterion is being used only if the ``oracle'' patches $\left\{\vect{p}_i \right\}_{i=1}^{\tilde{l}}$ are provided. They serve as an estimate for the patches of the true image and therefore we add atoms to the representation till the distance from these patches starts increasing. In this case the output of the sparse coding is the decoded representation from the previous ($t-1$) iteration.

One may argue that we should have used a Poissonian Bregman divergence for calculating the error and not the $\ell_2$ error. 
The reason we use the $\ell_2$ criterion is that it is the standard measure for checking the quality in image reconstruction. Note that we use this distance measure with the denoised image (which is no longer Poissonian) and not with the noisy image.

At the first time we apply the sparse coding algorithm in our \rg{recovery process}, we cannot use the bootstrapping based stopping criterion as we do not have an estimate for the original image yet. Therefore, we use the same cardinality $k$ for all groups. In the subsequent iterations the patches of the recovered image from the previous stage are used as an input to the algorithm for the bootstrapping stopping criterion.

\begin{figure}[htb]
\centering
\includegraphics[width=0.5\linewidth]{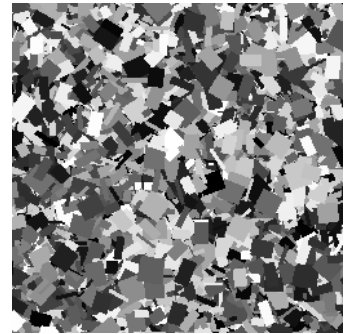}%
\caption{The piecewise constant image used for the offline training of the initial dictionary. For each range of peak values the image is scaled appropriately.}
\label{fig:triangles}
\end{figure}

\begin{algorithm}[t]
\caption{Sparse Poisson Denoising Algorithm \rg{(SPDA)}} \label{alg:SPDA}
\begin{algorithmic}

\REQUIRE $\vect{y}, \matr{D} \in \RR{d\times n}, k, l, h, R, L_1, L$ where $\vect{y}$ is a Poisson noise data 
with mean and variance $\vect{x}$, the patches in $\vect{x}$ are approximated (modeled) by $\exp(\matr{D}\alphabf_i)$, $k$ is the cardinality to be used for the representations at the first sparse coding step, $l$ is the group sizes, $h$ is the kernel filter used with the clustering algorithm, $R$ is the number of advanced dictionary learning rounds, $L_1$ is the number of inner learning iterations at the first round and $L$ is the number at the rest of the rounds.

\ENSURE $\hat{\vect{x}}$ an estimate for $\vect{x}$.

\STATE Begin Algorithm:

\STATE -Extract overlapping patches from $\vect{y}$.

\STATE -Apply the patch grouping algorithm on $\vect{y}$ with $h$.

\STATE -For each group apply the Poisson Greedy Algorithm with $k$.

\STATE -Put in $\hat{\vect{x}}$ a first estimate for $\vect{x}$ by a re-projection of the recovered patches by averaging.

\STATE -Set $t=0$

\WHILE{$t < R$}

\STATE -Update iteration counter: $t = t + 1$.

\IF{$t =1$}

\STATE Apply $L_1$ alternating \rg{N}ewton update steps for \eqref{eq:Poisson_sparsity_obj_fixed_supports}.

\ELSE 

\STATE Apply $L$ alternating \rg{N}ewton update steps for \eqref{eq:Poisson_sparsity_obj_fixed_supports}.

\ENDIF

\STATE -Put in $\hat{\vect{x}}$ an estimate for $\vect{x}$ by a re-projection of the recovered patches by averaging.

\STATE -Extract overlapping patches from $\vect{x}$.

\STATE -For each group apply the Poisson Greedy Algorithm with the bootstrapping stopping criterion that relies on the patches extracted from $\vect{x}$.

\ENDWHILE

\STATE -Apply the patch grouping algorithm on $\hat{\vect{x}}$ with no kernel ($h = [1]$).

\STATE -Repeat the whole process again using the new clustering and using the current dictionary as the initial dictionary.

\end{algorithmic}
\end{algorithm}

\subsection{Dictionary Learning}

Given the decoded representations we proceed to update the dictionary. 
We could have used the simple classical learning mechanism and just update the dictionary atoms, which can be done using a Newton step in a similar way to \cite{Salmon12Poisson} (with an addition of an Armijo rule). Instead, we utilize an advanced learning strategy from \cite{Smith13Improving}. Keeping the supports of the representations fixed, we update both the dictionary and the representations without changing their support.

Denoting by $T_i$ the support of the $i$-th patch, the problem we aim at solving in this case is
\begin{eqnarray}
\label{eq:Poisson_sparsity_obj_fixed_supports}
\min_{\matr{D},{\tilde\alphabf}_{1},\dots,{\tilde\alphabf}_{N}} \sum_{i=1}^{N} \vect{1}^*\exp(\matr{D}_{T_i}\tilde\alphabf_{i}) - \vect{q}_{i}^*\matr{D}_{T_i}\tilde\alphabf_{i}.
\end{eqnarray} 
First, note that it may happen that some atoms in the dictionary are not used by any of the representations. Therefore, we have no information for updating them. In this case, those are removed from the
dictionary resulting with a narrower dictionary.

Second, notice the similarity between \eqref{eq:Poisson_sparsity_obj_fixed_supports} and \eqref{eq:NLPCA_sparsity_obj}. As we can set $\matr{D}$ to be a concatenation of the different small dictionaries in  \eqref{eq:NLPCA_sparsity_obj}, we can view  \eqref{eq:Poisson_sparsity_obj_fixed_supports}  as a generalization of \eqref{eq:NLPCA_sparsity_obj} where groups can share atoms. In this sense we can look at the advanced learning strategy as a generalization of the GMM. The fact that we use small groups and apply a sparse coding after each advanced learning step gives more freedom and versatility to our scheme. 

With the above connection between the two, we can solve \eqref{eq:Poisson_sparsity_obj_fixed_supports}  using the technique for  solving \eqref{eq:NLPCA_sparsity_obj}: Applying alternating Newton steps (with an addition of an Armijo rule) both on the dictionary and the representations. Since these inner iterations reinforce the relation between the dictionary atoms and the related representation, 
the number of these iterations depends on our confidence in the selected supports. We select a different number of iterations for the first learning round and the rest because all the supports at the first round are an outcome of the sparse coding with a fixed cardinality and have the same size.

{\em Initial Dictionary Selection:}
The initial dictionary we use is a dictionary trained off-line on patches of a clean piecewise constant image shown in Fig.~\ref{fig:triangles}.
This training process for an initial dictionary is needed since, unlike the standard sparsity model
where many good dictionaries are present, for the exponential sparsity model no such dictionary is intuitively known.
Notice also that the representation in the new model is sensitive to the scale of the image, unlike
the standard one which is scale invariant. This is due to the fact that for a given constant $c$ we have necessarily that
\begin{eqnarray}
\matr{D}c\alphabf = c \matr{D}\alphabf
\end{eqnarray}
but in our case
\begin{eqnarray}
\exp\left(\matr{D}c\alphabf \right) \ne c \exp\left(\matr{D}\alphabf \right).
\end{eqnarray}
Thus, we train different initialization dictionaries for different peak values. The training is done by applying our denoising algorithm on a clean image (with no noise) scaled to the required peak value.
We should mention that this training is done once and off-line.

\begin{figure*}
\centering
\includegraphics[width=1\linewidth]{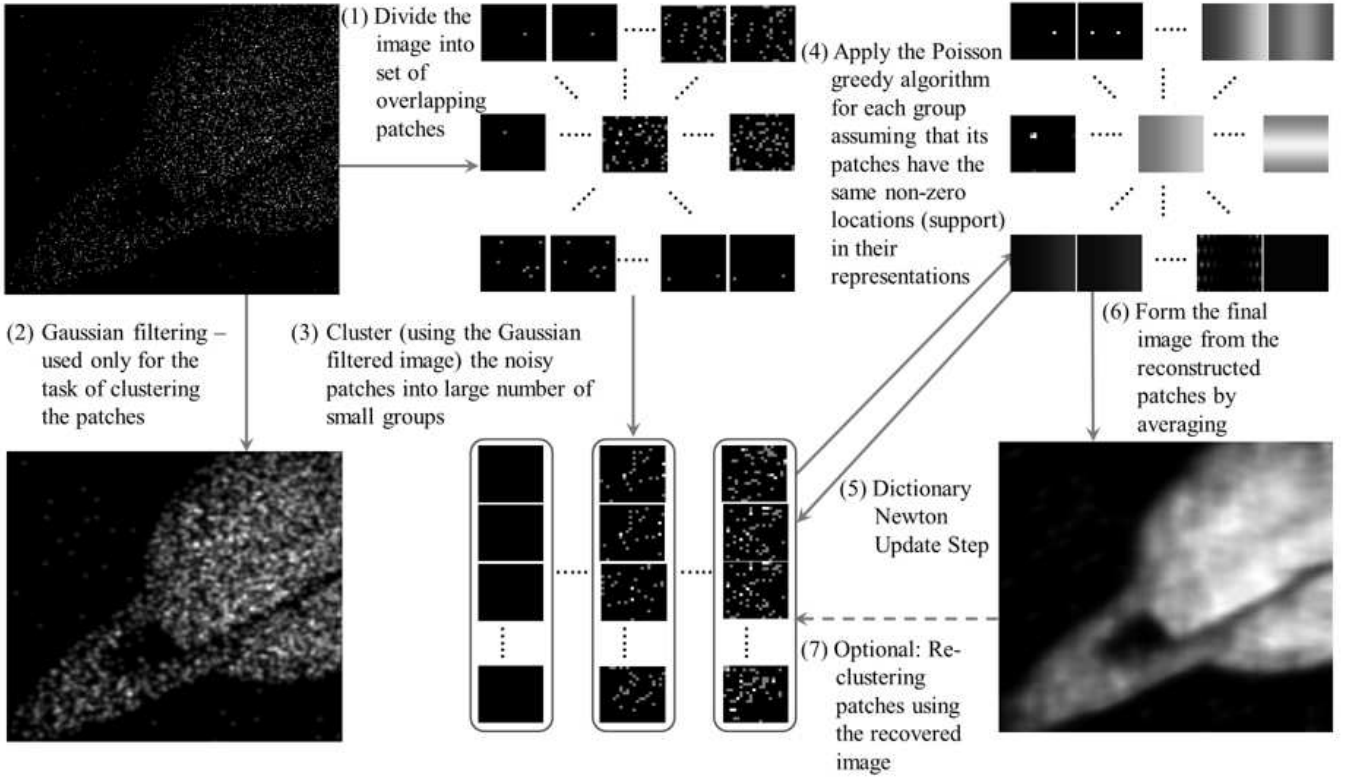}%
\caption{The proposed sparse Poisson denoising algorithm (SPDA).}
\label{fig:poisson_algorithm}
\end{figure*}

\subsection{SPDA Summary}

Iterating over the pursuit and dictionary learning stages we get an estimate for the image patches.
For recovering the whole image we reproject each patch to its corresponding location
with averaging. At this point it is possible to re-cluster the patches according
to the new recovered image and repeat all the process.
Our proposed sparse Poisson denoising algorithm (SPDA) is summarized in Fig.~\ref{fig:poisson_algorithm} and Algorithm~\ref{alg:SPDA}.

\subsection{Comparison to NLPCA and NLSPCA}

The main difference between our proposed algorithm and the NLPCA and NLSPCA
is reminiscent of the difference between the K-SVD image denoising algorithm \cite{Mairal09Non}
and the GMM alternative approach \cite{Yu12Solving}, both developed for the Gaussian denoising problem.
Furthermore, when it comes to the clustering of patches, NLPCA and NLSPCA use a small number of large clusters obtained
using a k-means like algorithm. In our scheme the notion of joint sparsity is used which implies a large number of disjoint small groups
that are divided using a simple algorithm. In NLPCA and NLSPCA, a different set of basis vectors is used for each cluster while we use
one (thicker) dictionary for all patches together, from which subspaces are created by different choices of small groups of atoms.
Unlike NLPCA and NLSPCA, our dictionary can change its size during the learning steps,
and the cardinalities allocated to different patches are dynamic.
As a last point we mention that NLPCA uses a Newton step and NLSPCA uses the SPIRAL algorithm
for the representation decoding while we propose a new greedy pursuit
for this task which guarantees a destination cardinality or reconstruction error.

We compare our algorithm's complexity to the one of NLPCA. As we are interested only in the order of the computational cost we focus on the bottlenecks of each algorithm. In NLPCA, the major computational part is solving \eqref{eq:NLPCA_sparsity_obj}. It is done by applying a constant number of Newton steps on the patches and the local dictionaries. The complexity of the Newton steps is of the order of the Hessian inversion. 
 Using the special properties of the Hessian in \eqref{eq:NLPCA_sparsity_obj} it is possible to perform its inversion with complexity $O(kd^{\frac{3}{2}})$ \cite{Salmon12PoissonConf}, where $k$ is the number of atoms in each local dictionary in NLPCA. Therefore, given that the number of patches is $N$, the total complexity of NLPCA is $O(Nk{d}^{\frac{3}{2}})$. For NLSPCA we get a similar complexity.
 
The bottleneck in SPDA is the sparse coding. At each iteration of adding an element to the representation, the sparse coding passes over all the atoms in the dictionary and solves \eqref{eq:pois_sparsity_obj_fixed_T_iter_j} checking the potential error of each of them. For calculating this error, we use a constant number of Newton steps. As before, the complexity of minimizing \eqref{eq:pois_sparsity_obj_fixed_T_iter_j} is $O(kd^{\frac{3}{2}})$. Since we start with a square dictionary and we target $k$ non-zero elements, the worst-case complexity is $O(k^2d^{\frac{5}{2}})$ per patch, as for each sparsity level we need to pass over all the atoms in the dictionary and for each solve \eqref{eq:pois_sparsity_obj_fixed_T_iter_j}.
Therefore, the overall complexity of our algorithm is  $O(Nk^2d^{\frac{5}{2}})$, where $N$ is the number of processed patches.

Though our method is more computationally demanding, we shall see in the next section that in most cases this cost leads to better denoising results. It should be mentioned also that SPDA is highly parallelizable which allows a great reduction in its running time.
Note also that since the size of our dictionary may shrink between  the sparse coding steps, the complexity of the sparse coding is likely to become smaller at the latter stages of SPDA.

\section{Experiments}
\label{sec:exp}

\begin{figure}[htb]
\centering
{\subfigure{\includegraphics[width=0.24\linewidth]{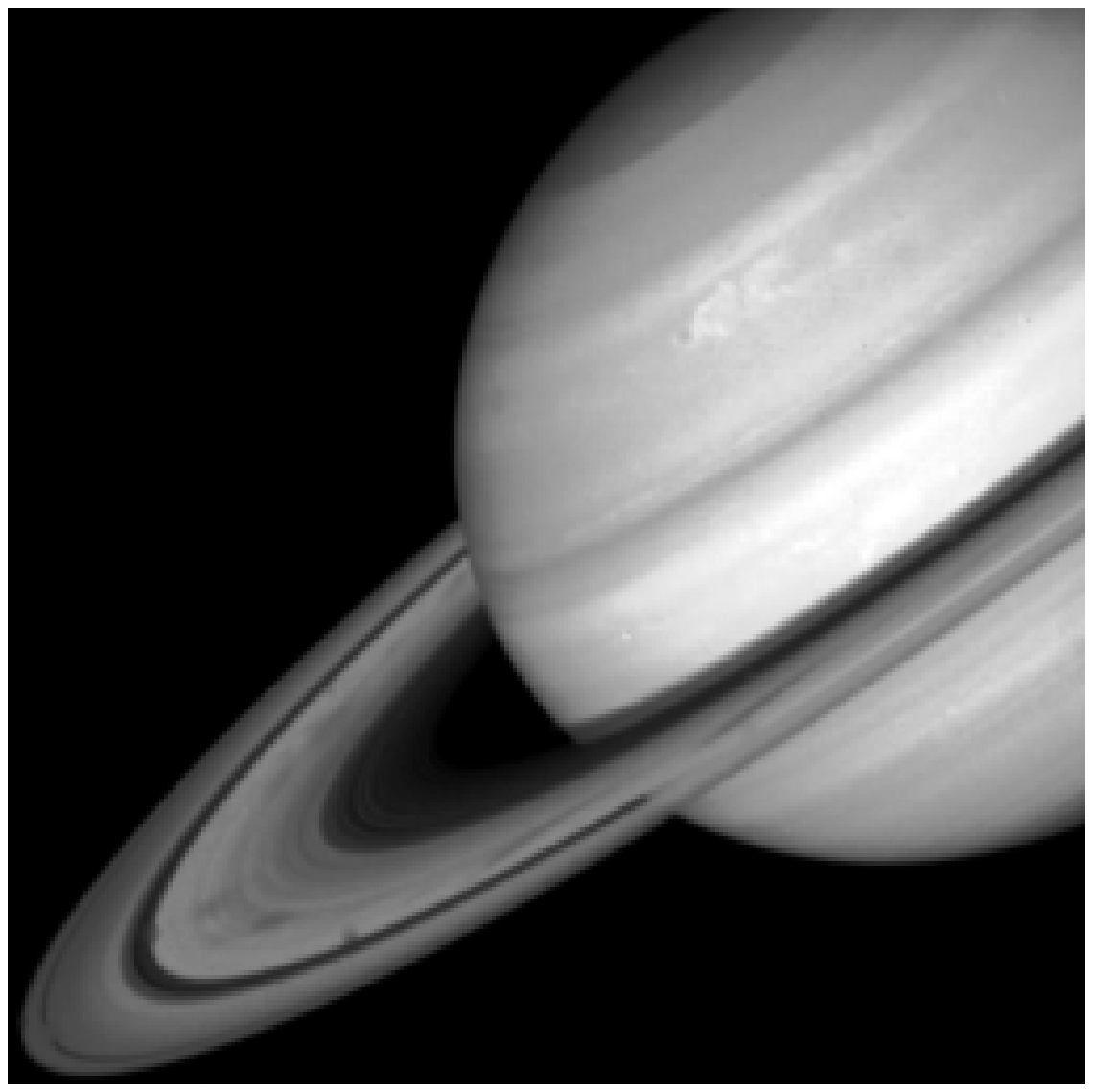}}%
\hfil
\subfigure{\includegraphics[width=0.24\linewidth]{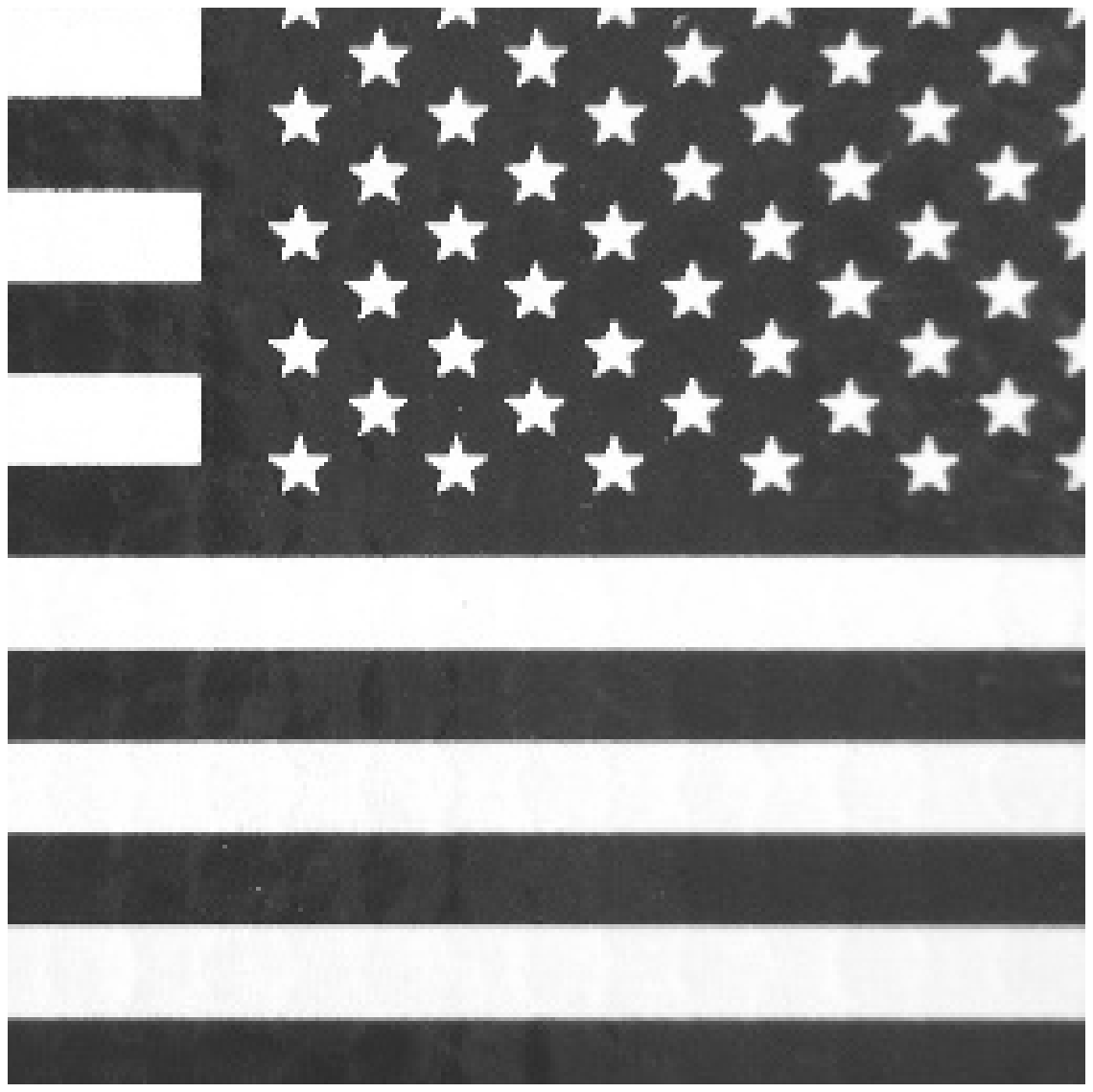}}
\hfil
\subfigure{\includegraphics[width=0.24\linewidth]{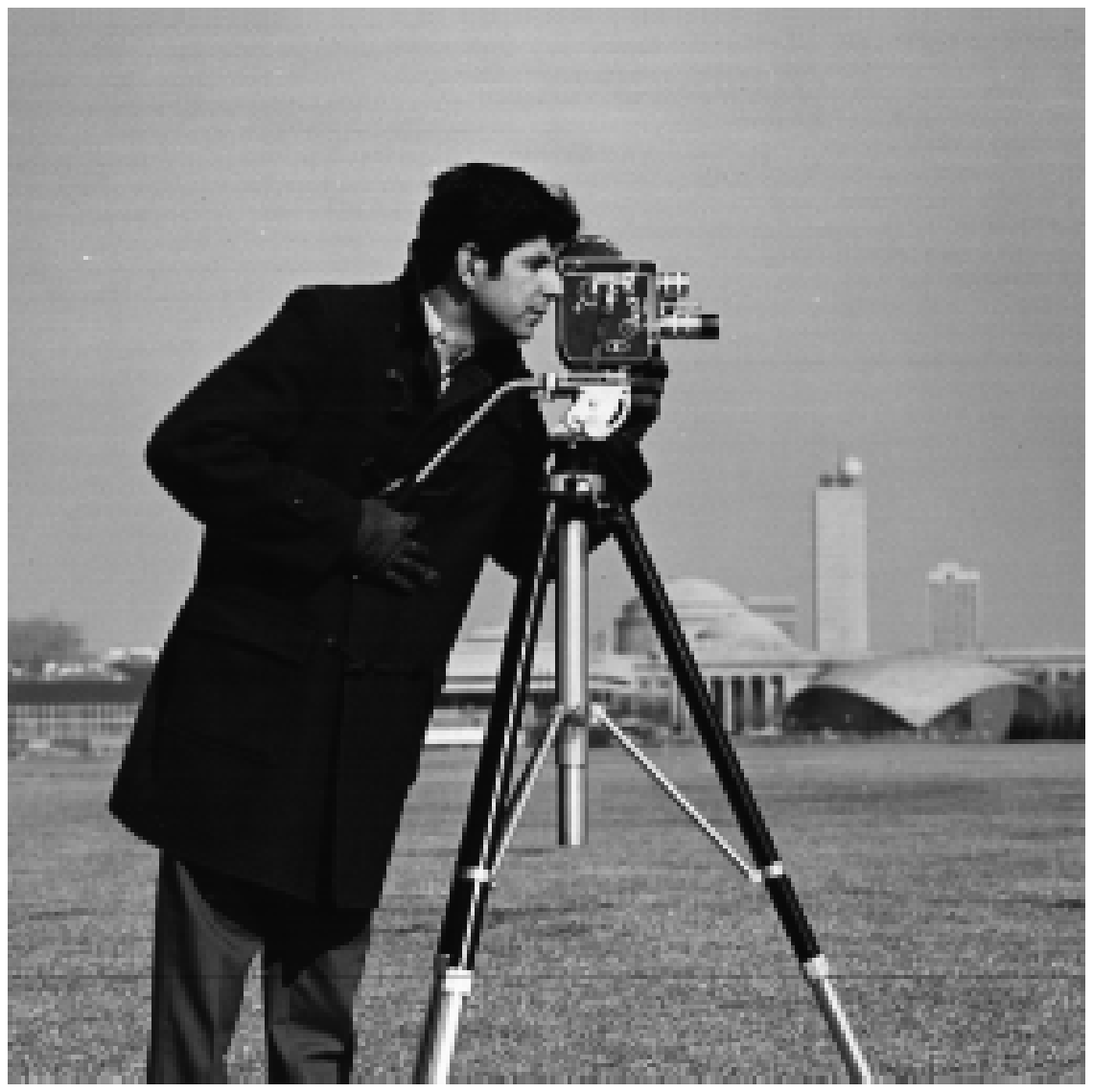}}
\hfil
\subfigure{\includegraphics[width=0.24\linewidth]{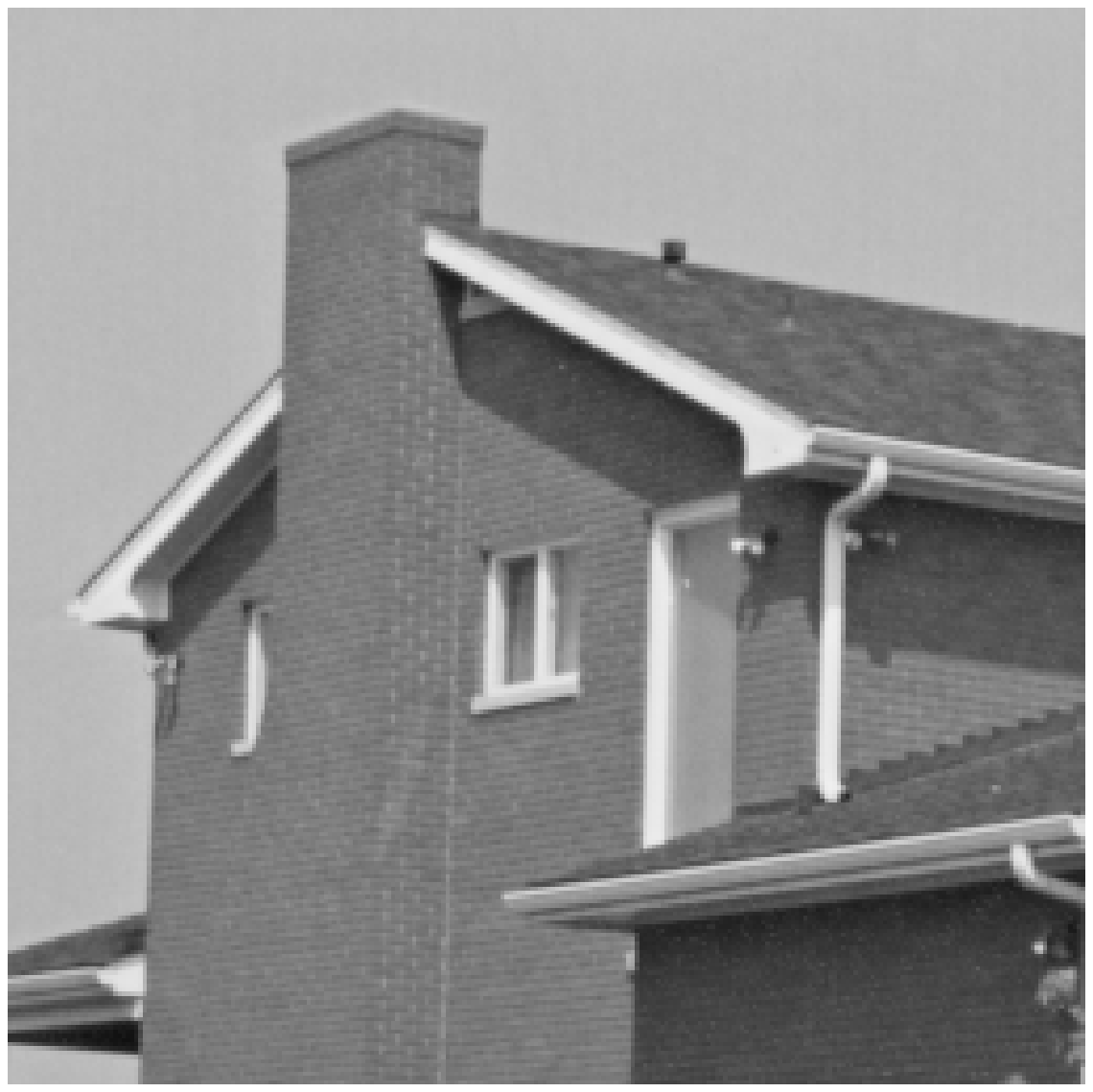}}
\hfil
\subfigure{\includegraphics[width=0.24\linewidth]{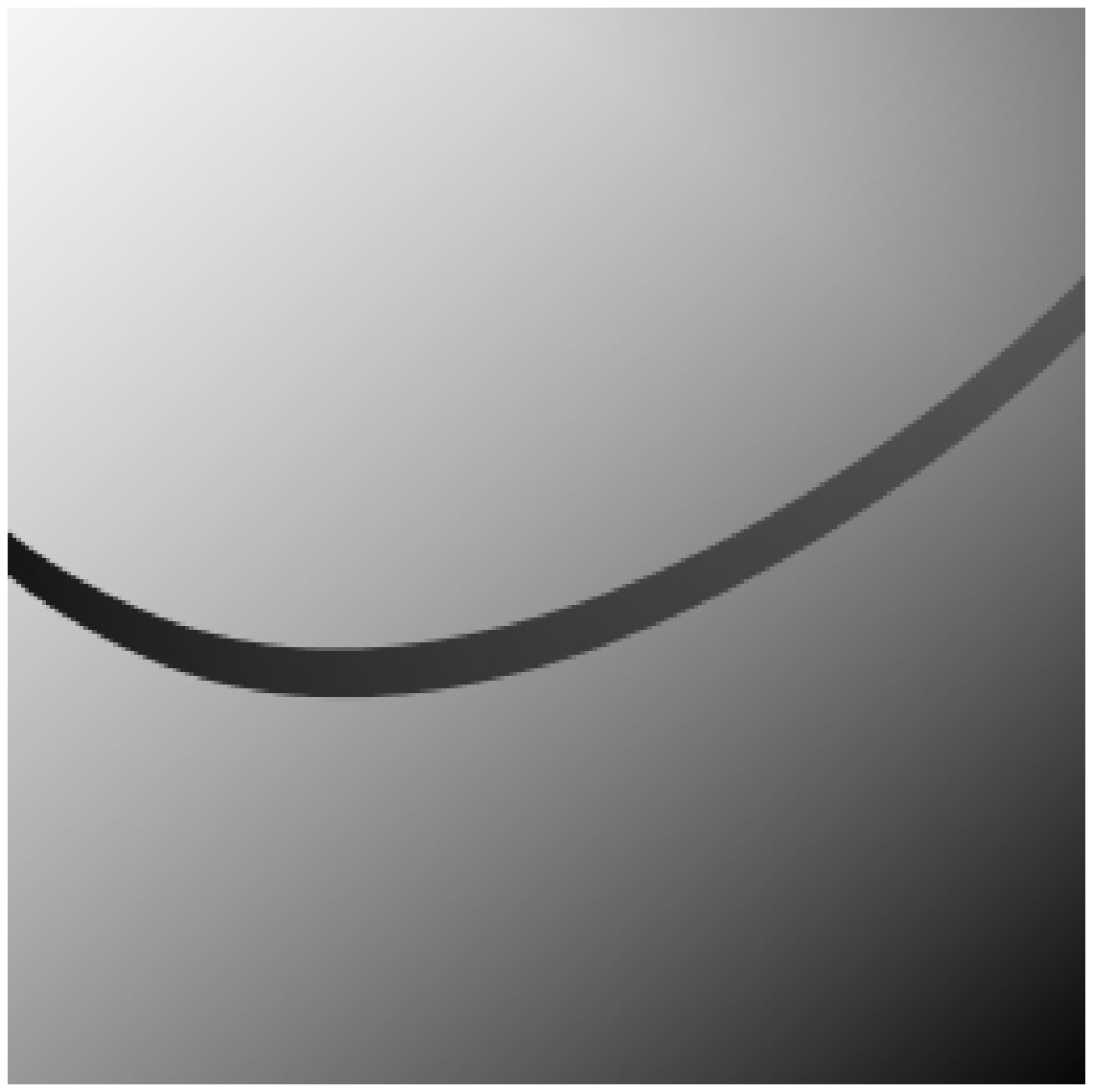}}
\hfil
\subfigure{\includegraphics[width=0.24\linewidth]{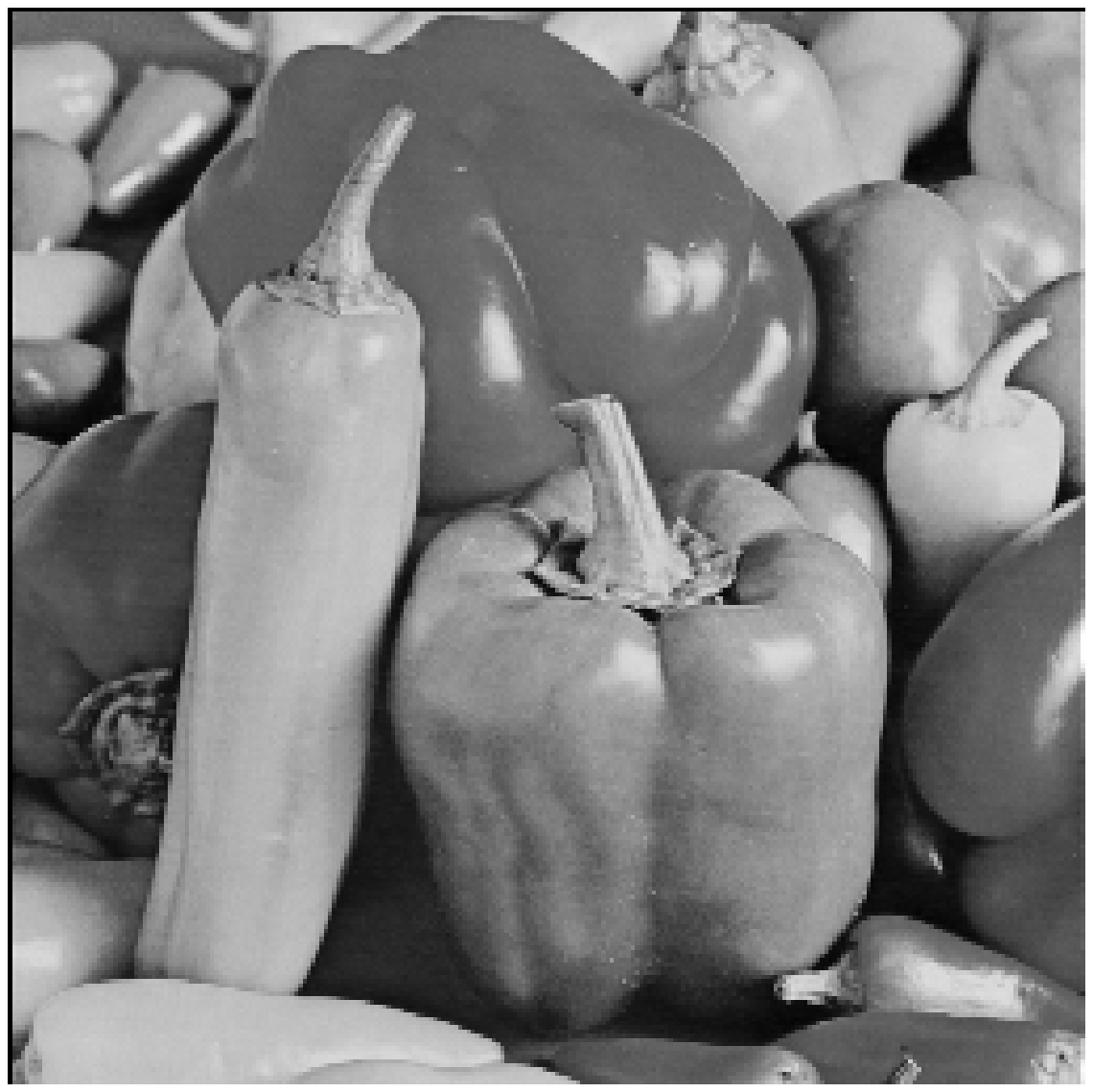}}
\hfil
\subfigure{\includegraphics[width=0.24\linewidth]{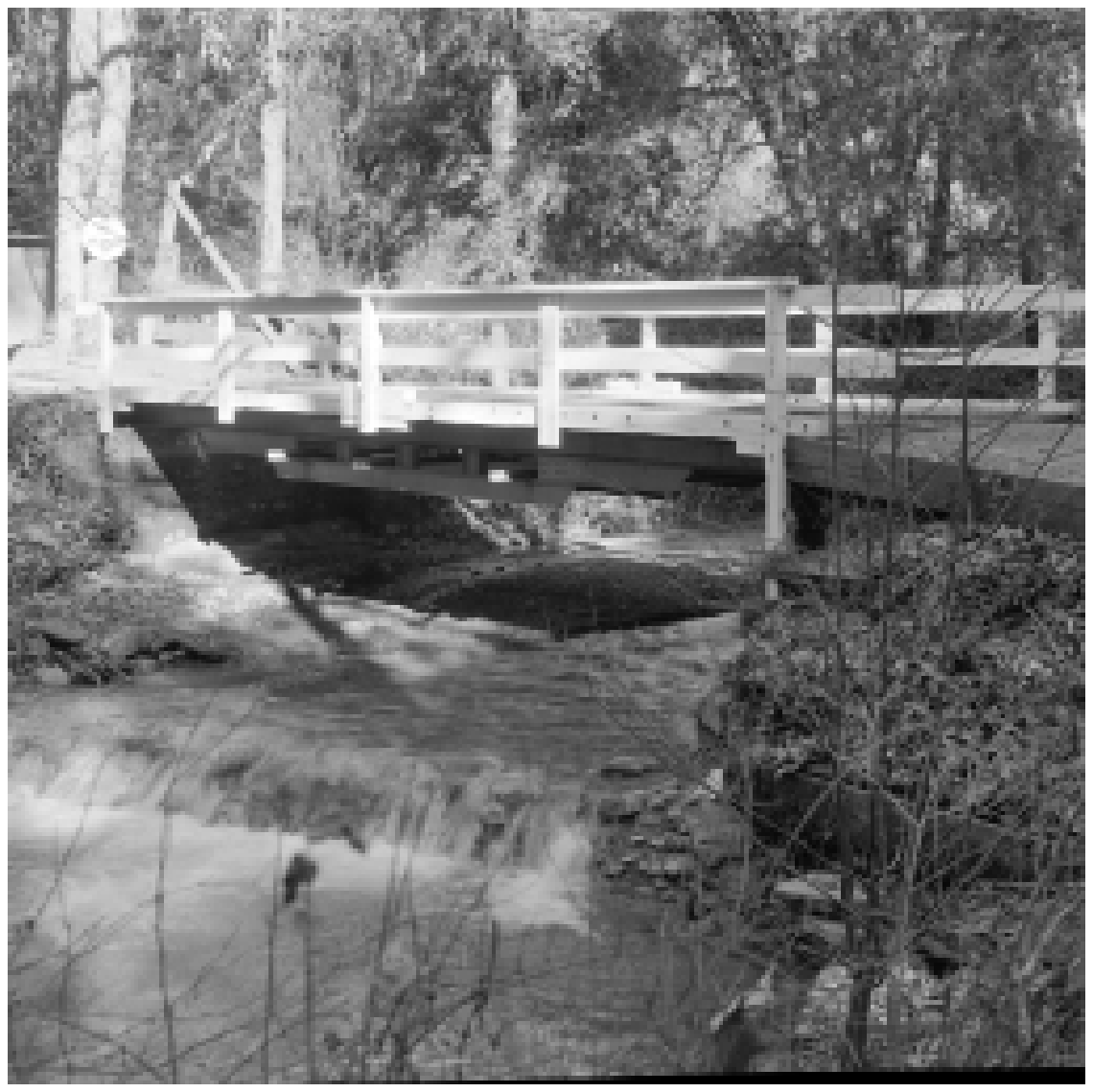}}
\hfil
\subfigure{\includegraphics[width=0.24\linewidth]{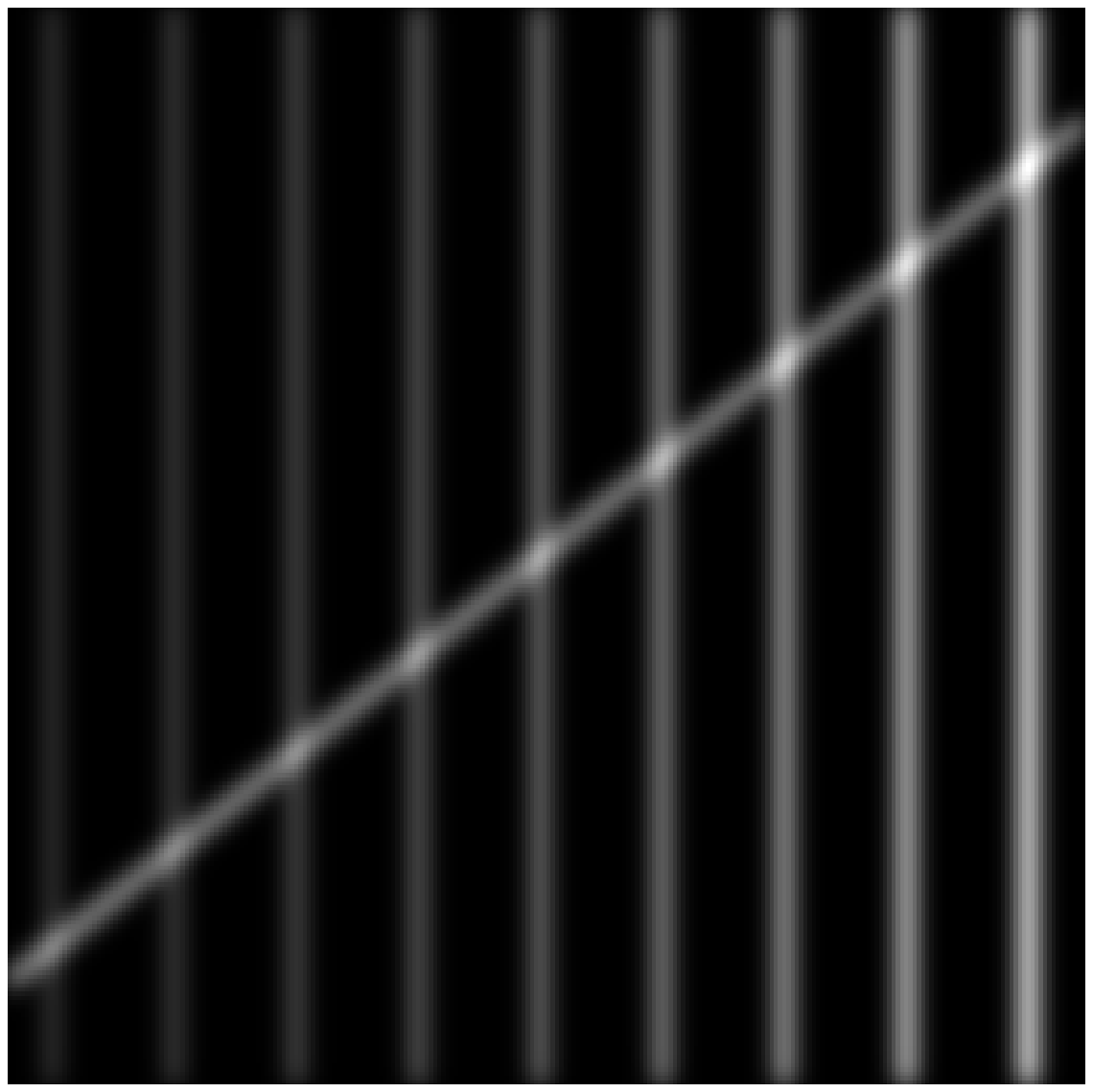}}
}%
\caption{Test images used in this paper. From left to right: Saturn, Flag, Cameraman, House, Swoosh, Peppers, Bridge and Ridges.}
\label{fig:test_images}
\end{figure}

In order to evaluate the SPDA (with and without binning) performance we repeat the denoising experiment performed in \cite{Salmon12Poisson}.
We test the recovery error for various images with different peak values ranging from $0.1$ to $4$.
The tested images appear in Fig.~\ref{fig:test_images}.
The methods we compare to are the NLSPCA \cite{Salmon12Poisson} and the BM3D with
the exact unbiased inverse Anscombe \cite{Makitalo11Optimal} (both with and without binning)
as those are the best-performing methods up to date.
The code for these techniques is available online
and we use the same parameter settings as appears in the code. For the binning we follow \cite{Salmon12Poisson} and use
a $3\times 3$ ones kernel that increases the peak value to be $9$ times higher, and a bilinear interpolation for the upscaling of the low-resolution recovered image.

\begin{table}
\footnotesize
\begin{center}
\begin{tabular}{|c|c|}
\hline
Parameter & Value\\
\hline\hline
Image Size & $256 \times 256$\\
\hline
Patch Size & $20 \times 20$ \\
\hline 
$k$ in First Sparse Coding Step & $k=2$\\
\hline
Cluster Size ($l$) &  $ $50$~\textit{(no binning)}, $6$~\textit{(binning)}$\\
\hline
Dictionary Learning Rounds & $R=5$\\
\hline 
Dictionary Update Inner Iterations & $L_1=2$ and $L=20$\\
\hline
Reclustering & Once \\
\hline
Initial Dictionary (Peak$\le 0.2$) & Trained on Fig.~\ref{fig:triangles} with Peak$=0.2$ \\
\hline 
Initial Dictionary ($0.2<$Peak$\le 4$) & Trained on Fig.~\ref{fig:triangles} with Peak$=2$ \\
\hline 
Initial Dictionary (Peak$> 4$) & Trained on Fig.~\ref{fig:triangles} with Peak$=18$ \\
\hline 
Binning Kernel & $3 \times 3$ Ones Kernel \\
\hline 
\end{tabular}
\end{center}
\caption{Parameter Selection}
\label{tbl:param_sel}
\end{table}

For SPDA we use the following parameter setting:
The size of each patch is set to be $20 \times 20$ pixels.
We start with a sparse coding with a fixed sparsity $k=2$. Then we apply five rounds of sparse coding with the bootstrapping based stopping criterion together with the advanced dictionary learning mechanism that contains a joint update of the dictionary and the representations (with fixed support).
In the first round we apply $2$ inner iterations of the dictionary update and in the rest we use $20$ inner iterations. The reason we use a different number of inner iterations at the first round is that in this round  the supports of the representations are  less reliable as they are selected with a fixed support size. After the first round, the bootstrapping based stopping criterion is being employed and each group is being decoded with a different cardinality leading to a better support selection.
We re-cluster using the outcome of the above process and repeat it again.

We remind the reader that in selecting the clusters size we have a trade-off between the number of groups and size of each cluster. 
As each group selects the same support, having more groups provides us with more information for the dictionary update process. On the other hand, the larger the cluster the more probable it is that we select the ``right'' support.  Of course, we could have used overlapping groups but we have chosen not to do so due to computational reasons.
As a rule of thumb, we have found that having $1000$ groups is enough for the dictionary update. 
As our images are of size $256 \times 256$, this implies a group size $l=50$.
If binning is not used and $l=6$ if it is used. 
Note that the smaller group size in the binning case is compensated by the fact that each patch contains more information as it represents a larger portion in the original image.

The initial dictionary is square ($400 \times 400$) and dependent on the peak value.
For peak$\le0.2$ we use a dictionary trained on the squares image with peak $0.2$,
for $0.2<$peak$\le 4$ we train for peak$=2$ and for peak$>4$ we train for peak$=18$
(this is relevant for SPDA with binning when the original peak is greater than $4/9$). Note that since in Poisson noise the mean is equal to the original signal, a rough estimation for the peak is an easy task.
We note that our algorithm is not sensitive to the initial dictionary selection: if the peak value is inaccurately estimated and the ``wrong'' dictionary is selected, the recovery is not affected significantly. Indeed, we could have trained a different initial dictionary for each peak value. However, we chose not to do so, in order to demonstrate the insensitivity of our scheme to the initialization.
We remark that it is possible to use other reasonable initializations to our update process such as the log of the absolute values of the DCT transform. From our experience, there is \rg{not} much difference in the reconstruction result and the gap in the recovery error is in the range of only $0.2dB$\footnote{A package with the code reproducing all our results can be found at www.cs.technion.ac.il/$\sim$raja.}.
The parameter selection is summarized in Table~\ref{tbl:param_sel}

\begin{figure*}
\centering
{\subfigure[{\em flag} image.]{\includegraphics[width=0.24\linewidth]{flag}}%
\hfil
\subfigure[NLSPCA. PSNR =  20.37dB]{\includegraphics[width=0.24\linewidth]{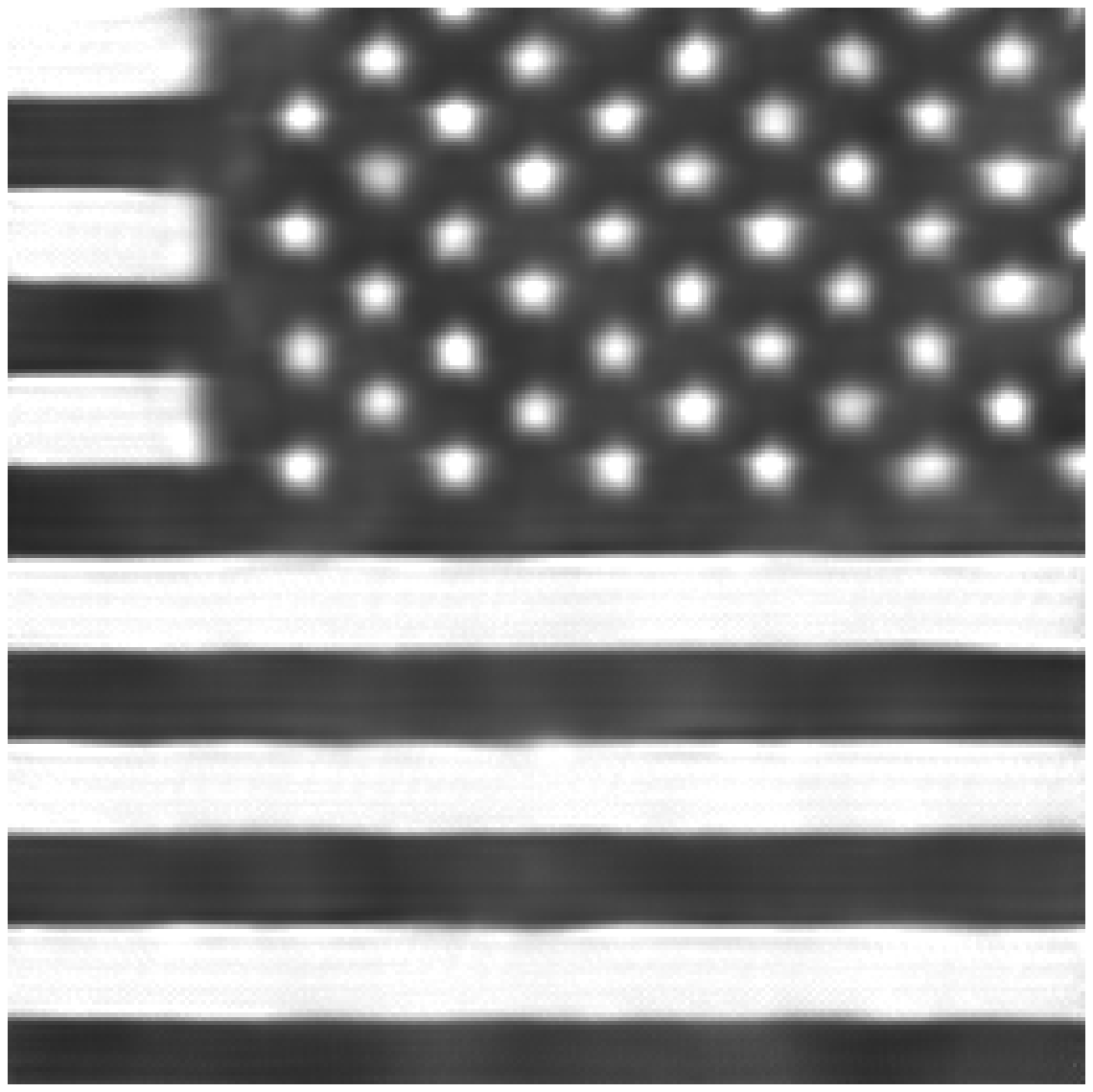}}
\hfil
\subfigure[BM3D. PSNR = 18.51dB.]{\includegraphics[width=0.24\linewidth]{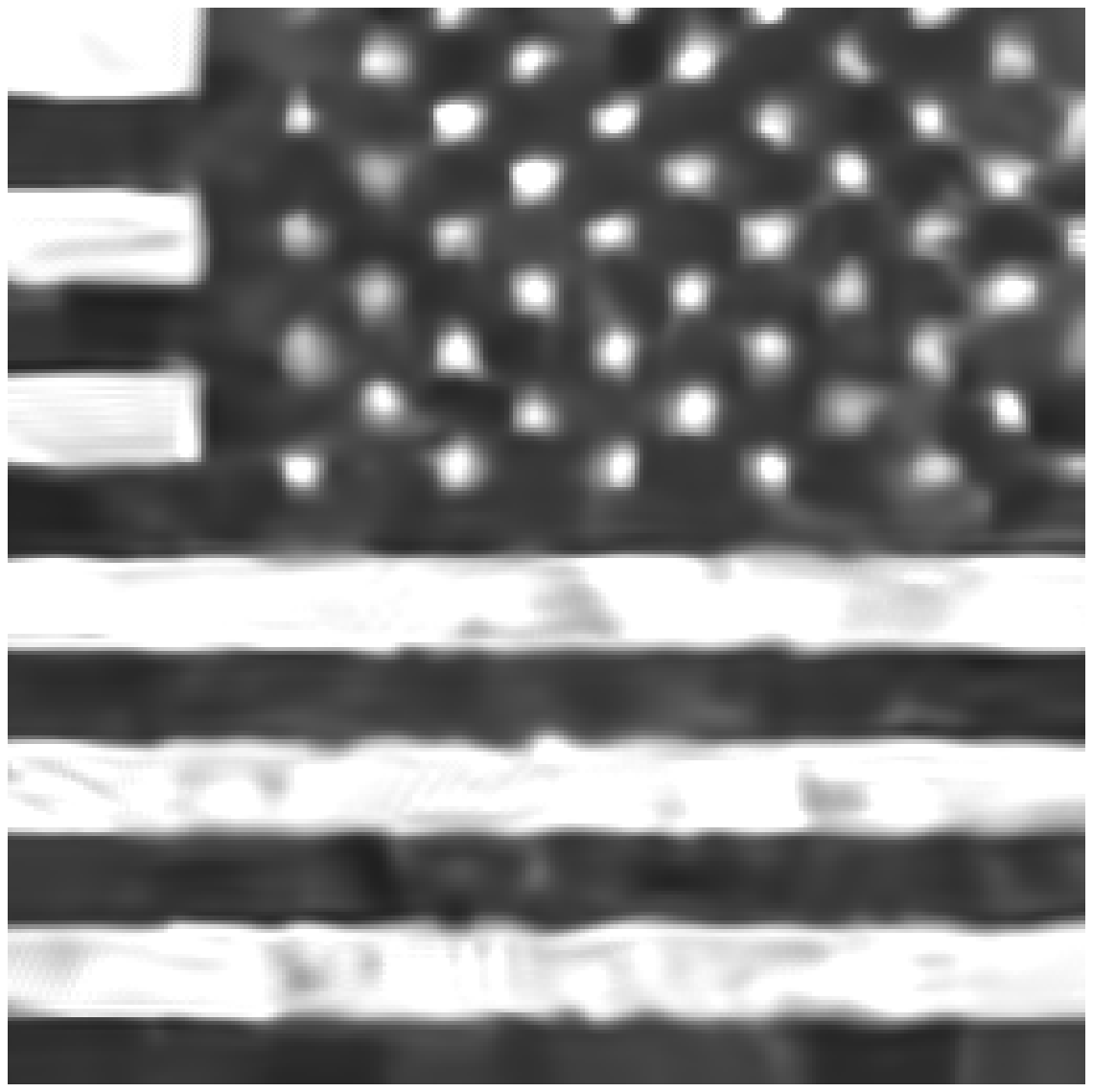}}
\hfil
\subfigure[SPDA. PSNR =  {\bf22.59dB}.]{\includegraphics[width=0.24\linewidth]{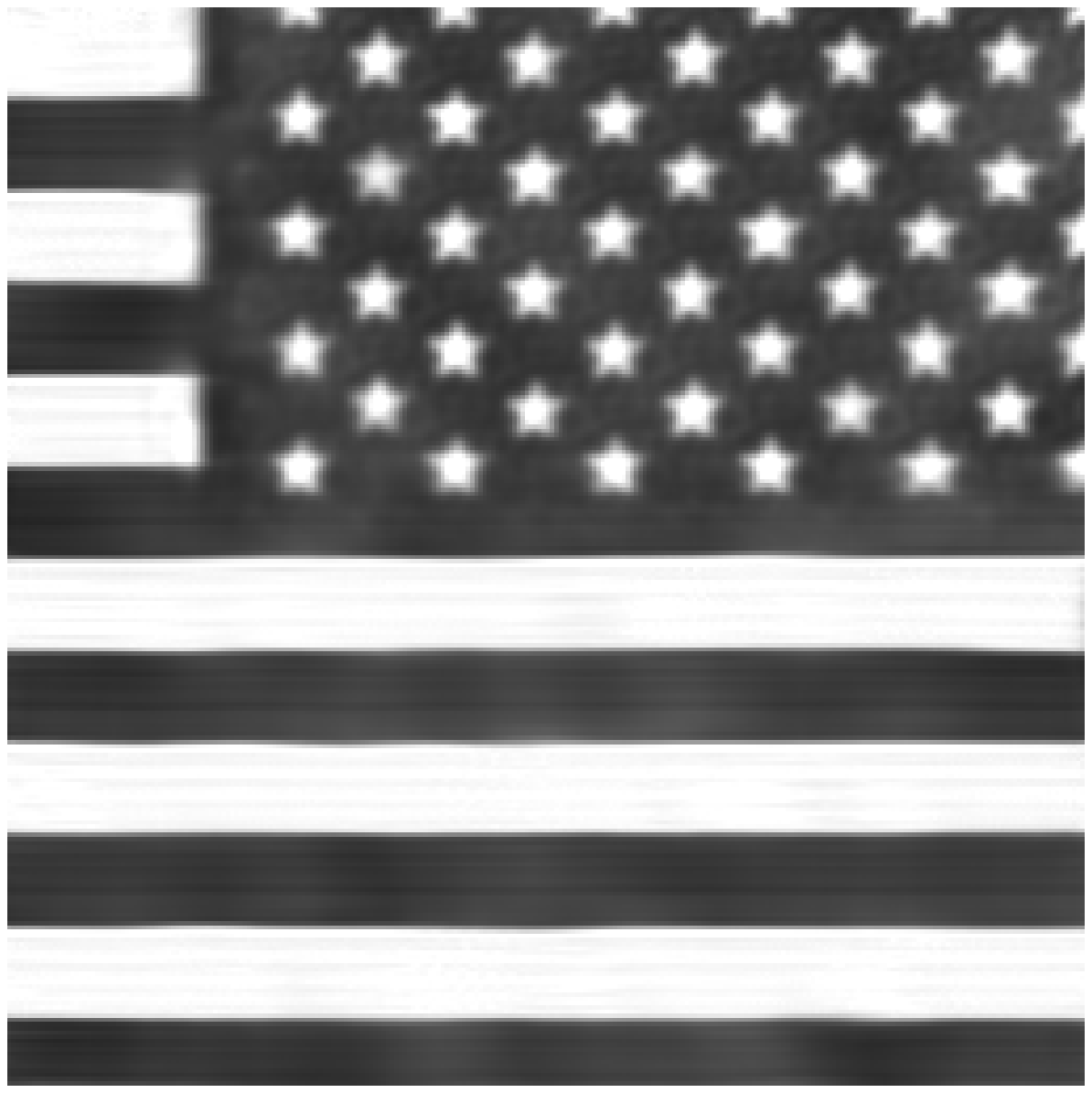}}
\vfil
\subfigure[Noisy image. Peak = 1.]{\includegraphics[width=0.24\linewidth]{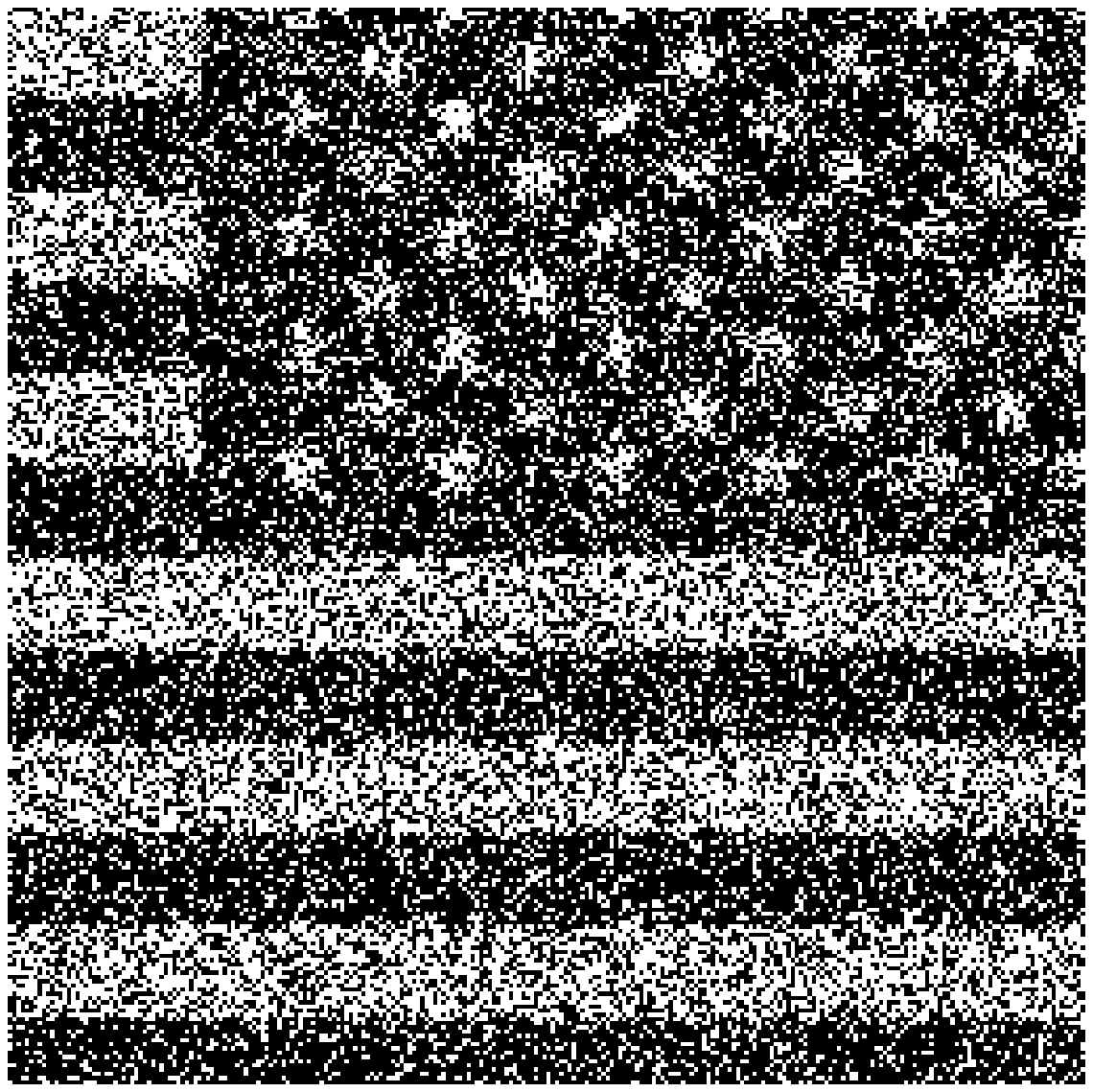}}
\hfil
\subfigure[NLSPCAbin. PSNR =  16.91dB.]{\includegraphics[width=0.24\linewidth]{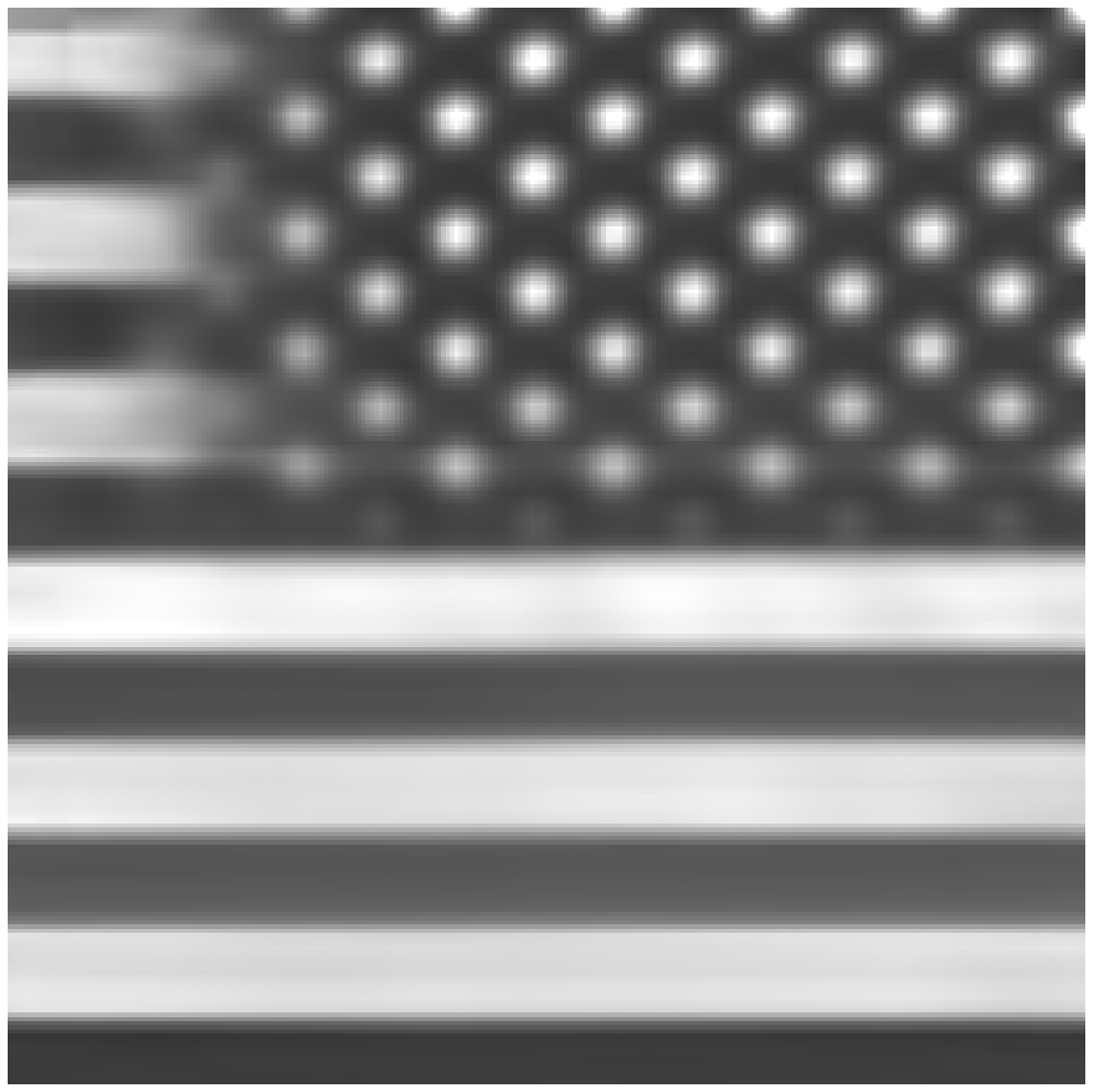}}
\hfil
\subfigure[BM3Dbin. PSNR = 19.41dB.]{\includegraphics[width=0.24\linewidth]{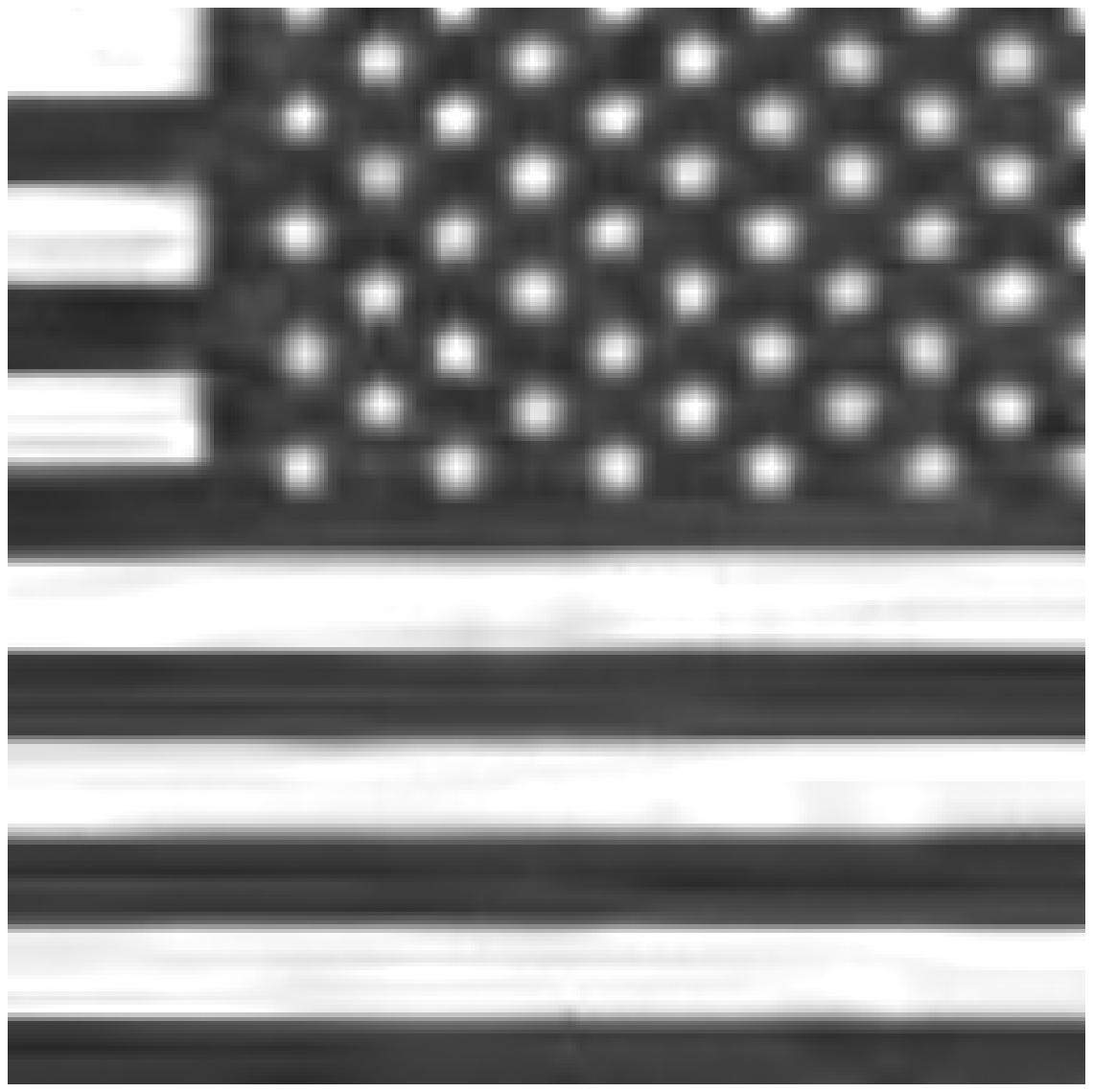}}
\hfil
\subfigure[SPDAbin. PSNR = 19.9dB.]{\includegraphics[width=0.24\linewidth]{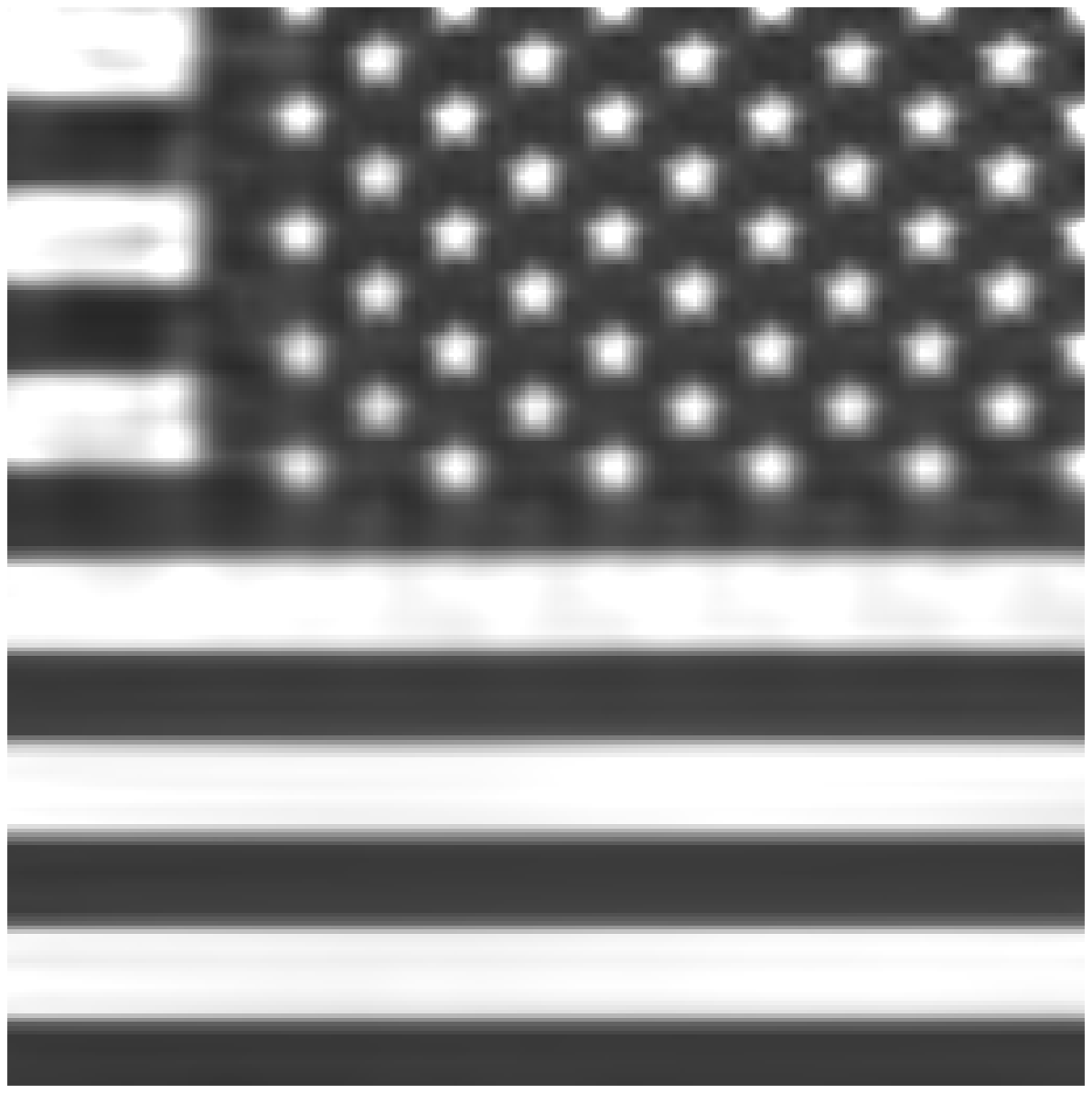}}
}%
\caption{Denoising of {\em flag} with peak = 1. The PSNR is of the presented recovered images.}
\label{fig:flag_recovery}
\end{figure*}

\begin{figure}[htb]
\centering
{\subfigure{\includegraphics[width=0.11\linewidth]{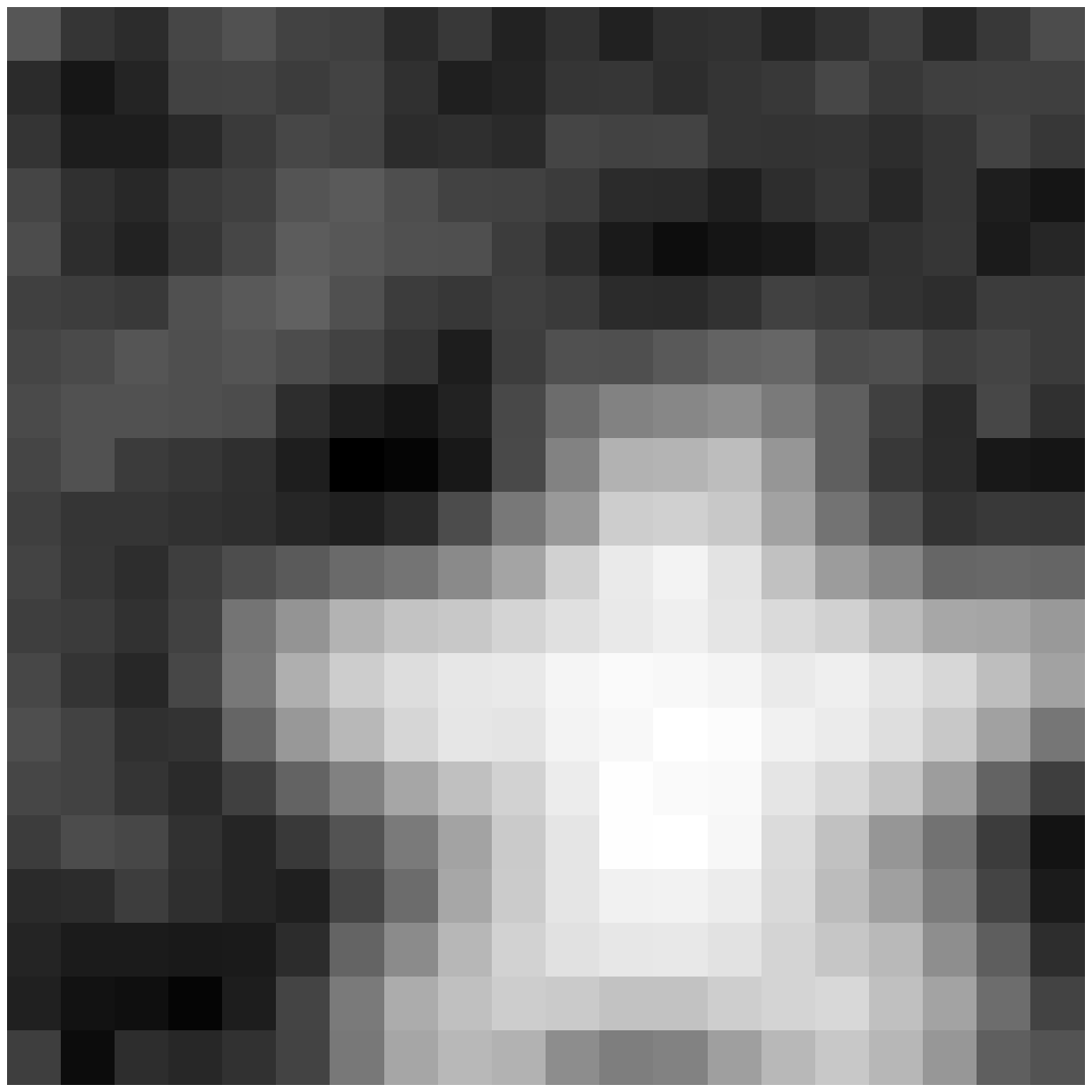}}%
\hfil
\subfigure{\includegraphics[width=0.11\linewidth]{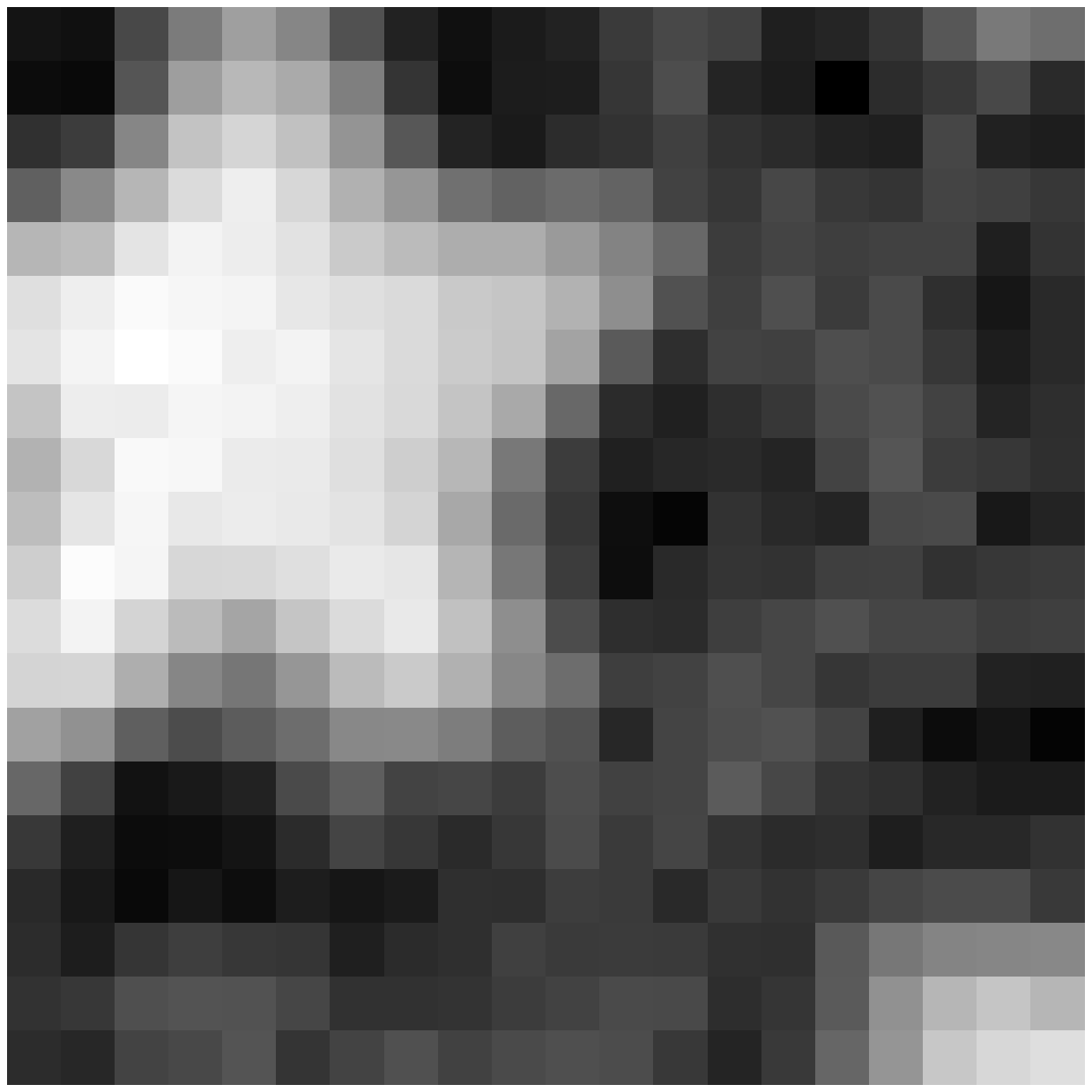}}
\hfil
\subfigure{\includegraphics[width=0.11\linewidth]{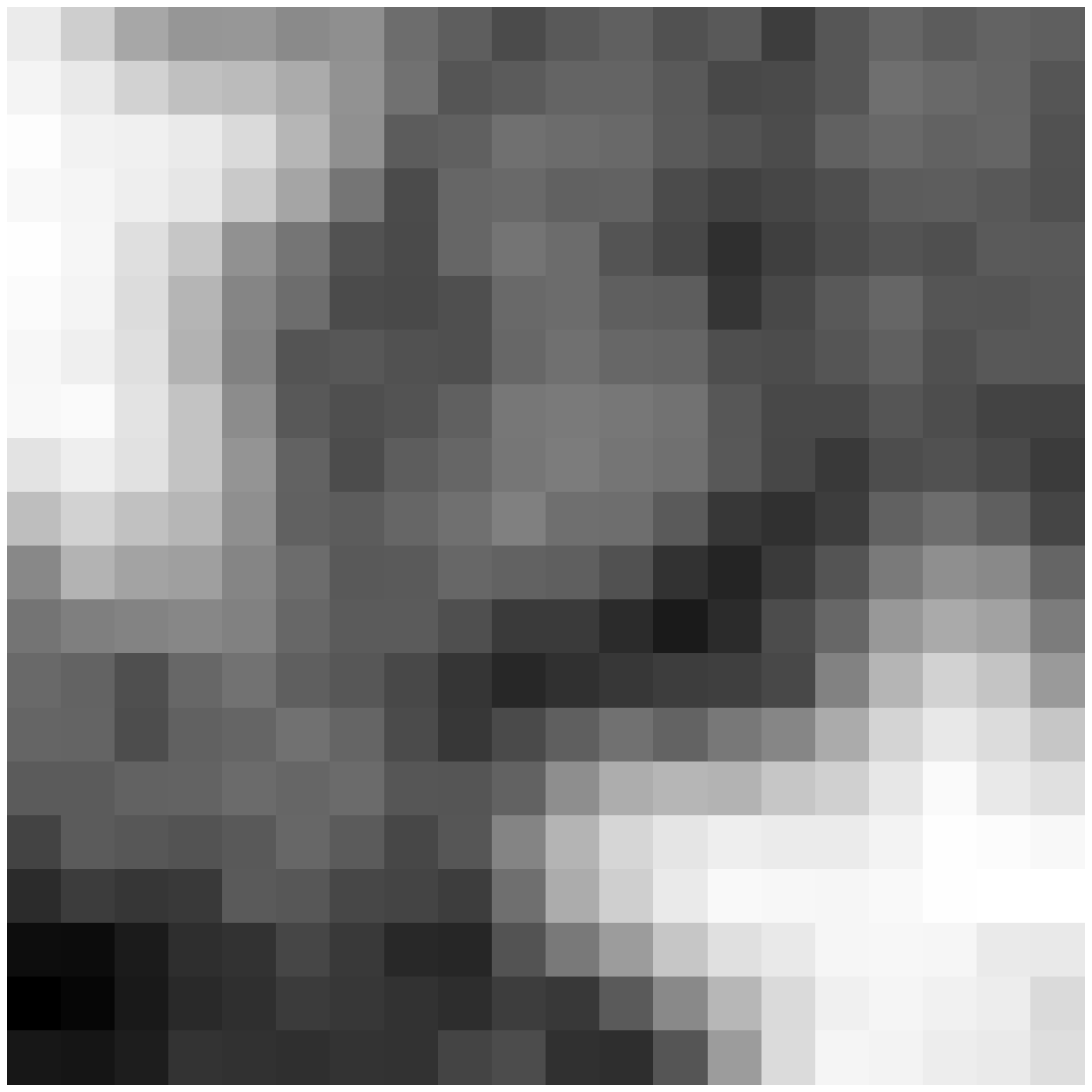}}
\hfil
\subfigure{\includegraphics[width=0.11\linewidth]{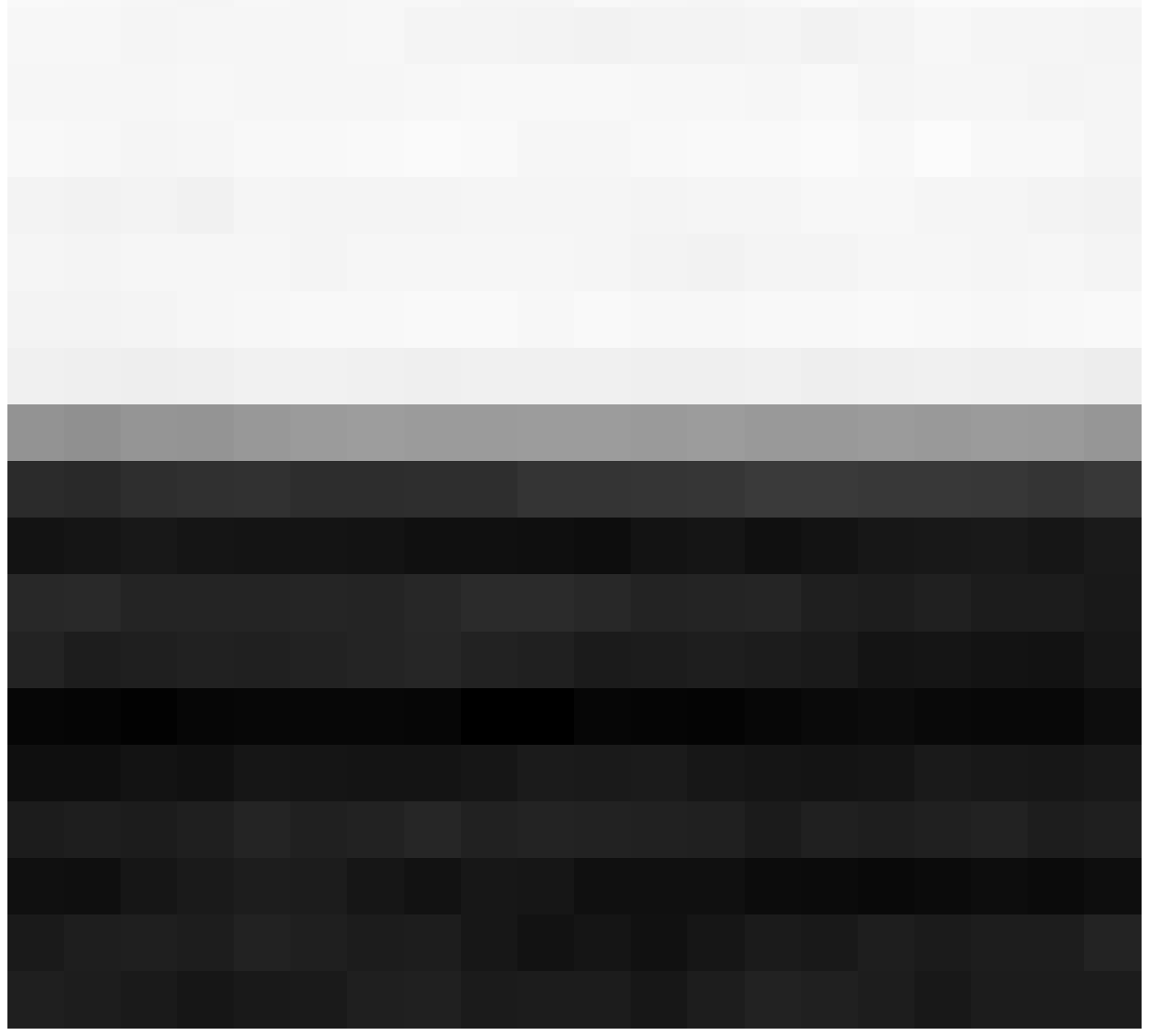}}
\hfil
\subfigure{\includegraphics[width=0.11\linewidth]{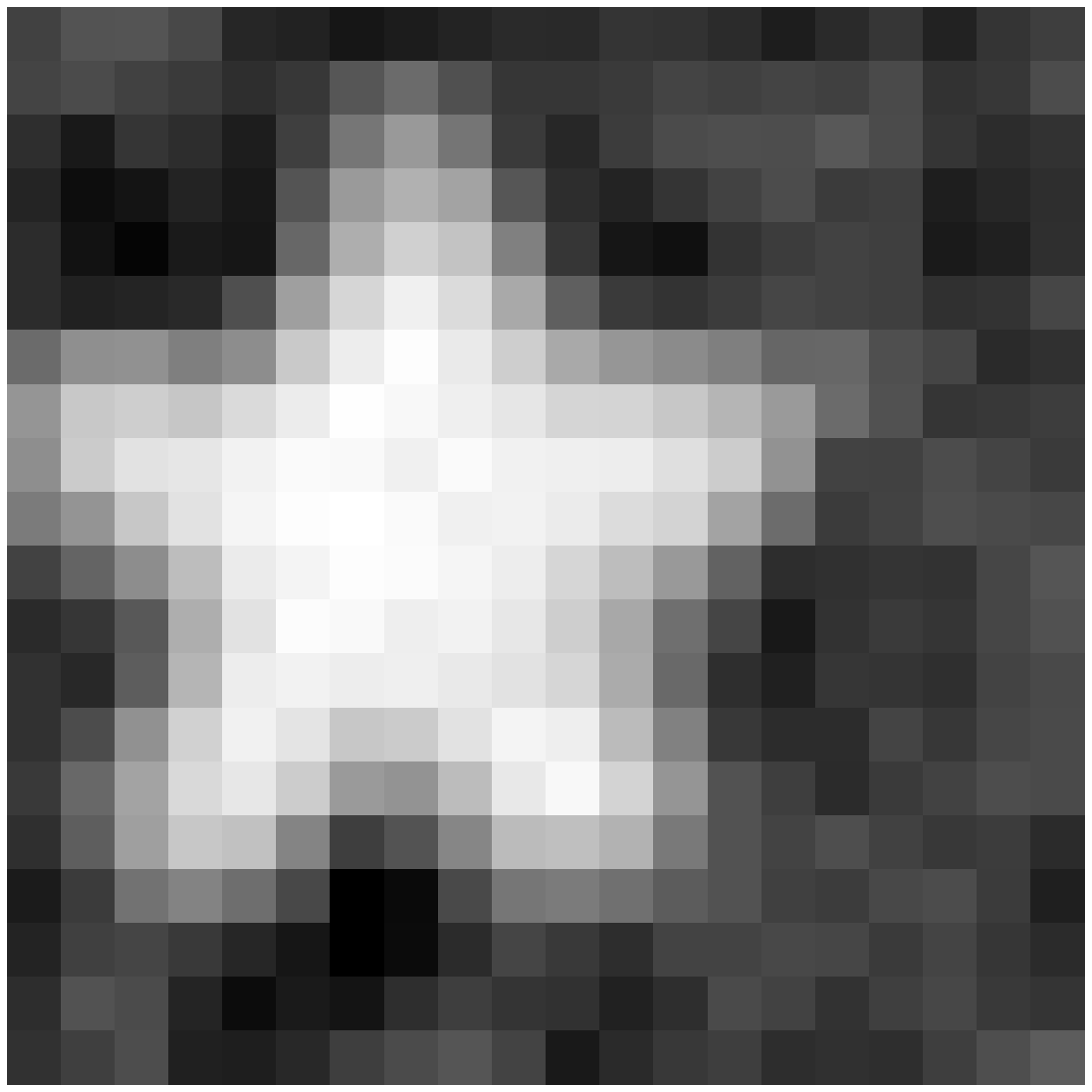}}
\hfil
\subfigure{\includegraphics[width=0.11\linewidth]{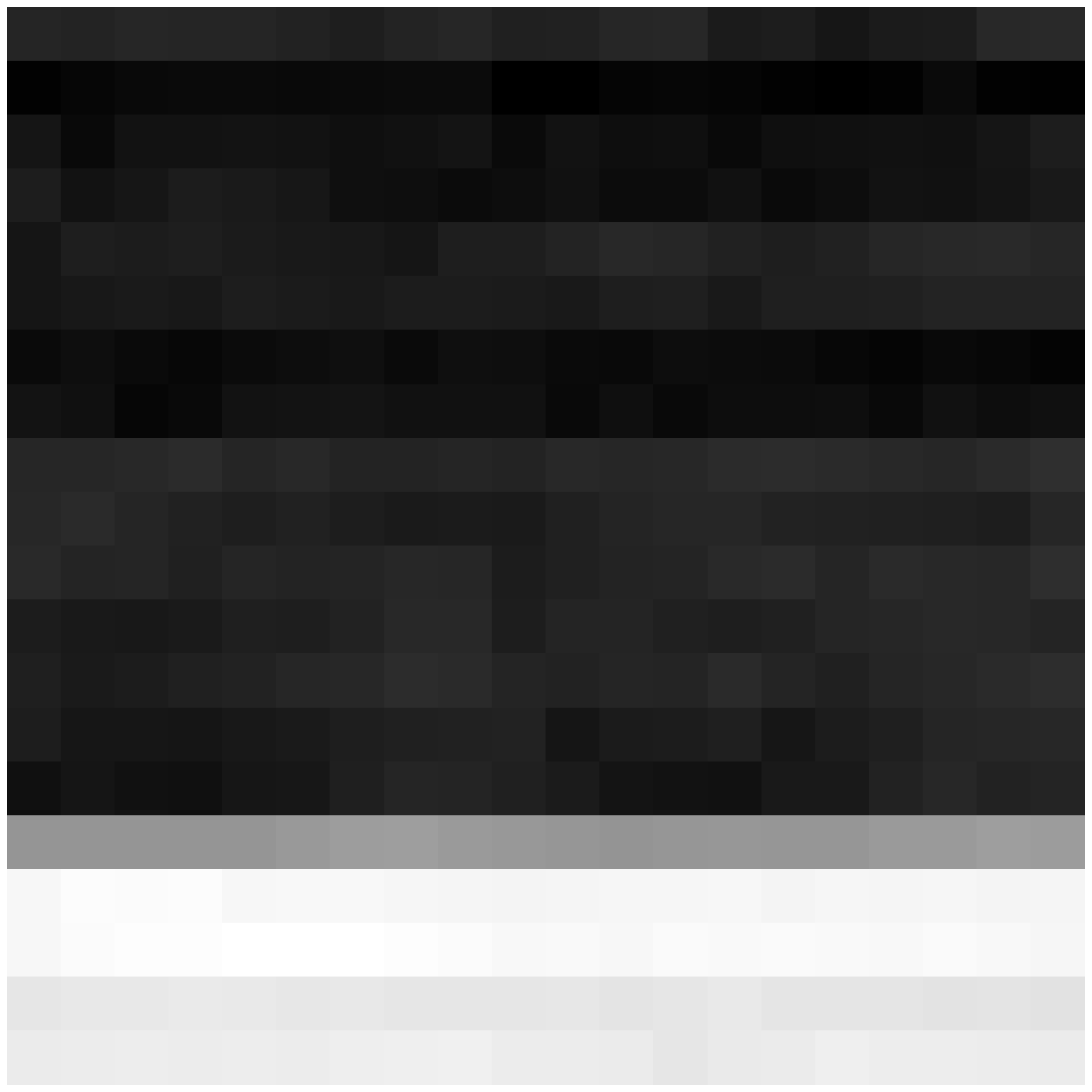}}
\hfil
\subfigure{\includegraphics[width=0.11\linewidth]{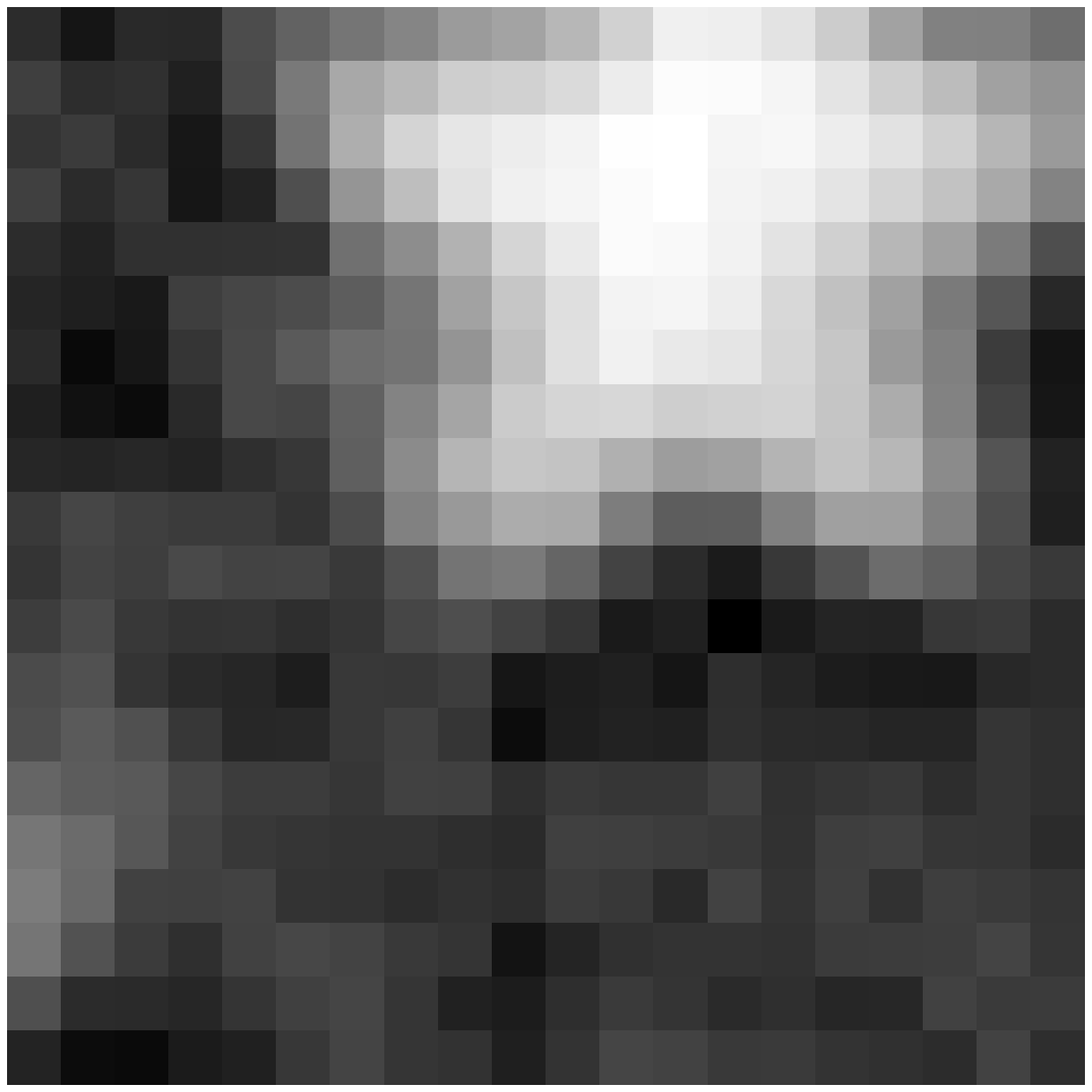}}
\hfil
\subfigure{\includegraphics[width=0.11\linewidth]{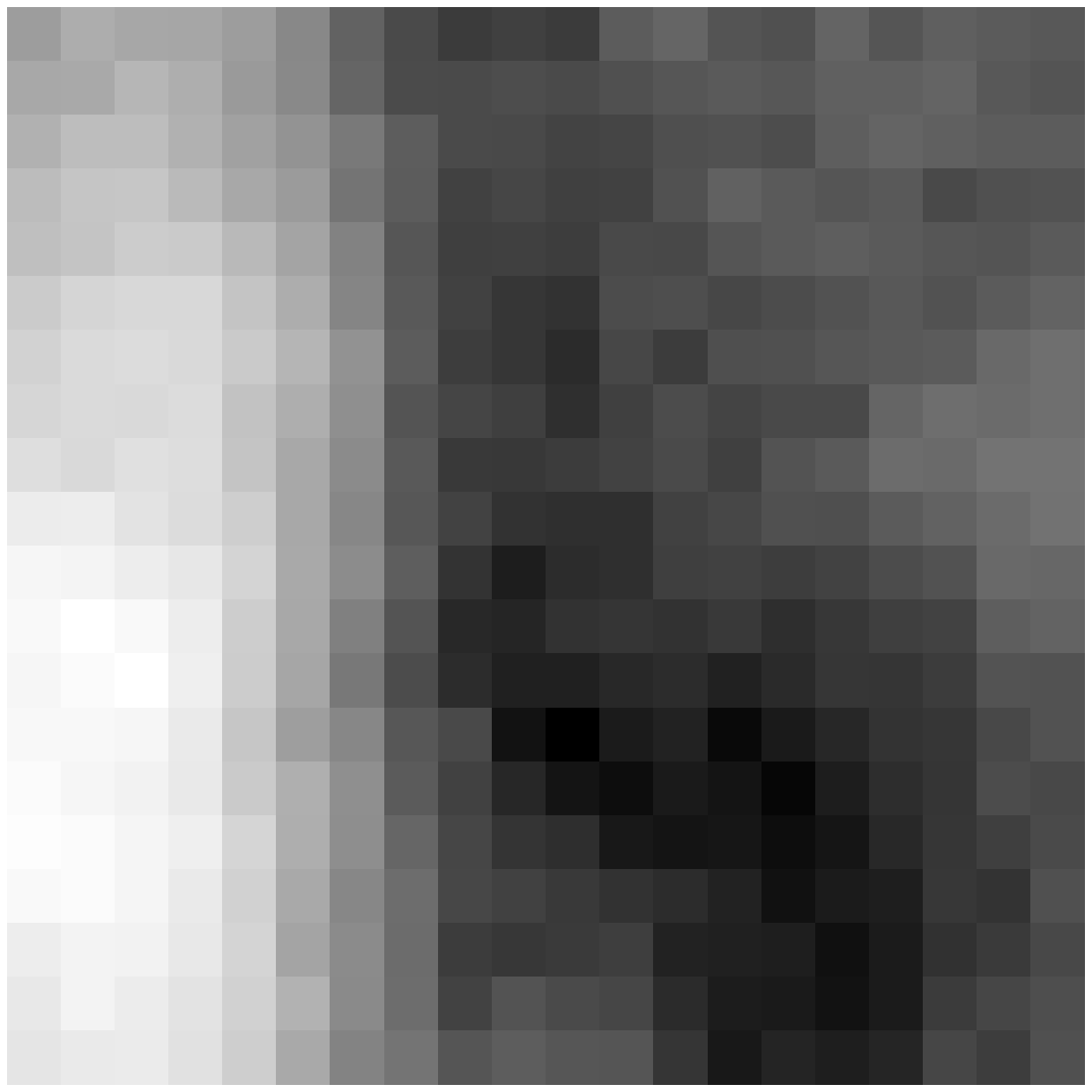}}
}%
\caption{Samples of atoms from the dictionary $\matr{D}$ learned using SPDA for {\em flag} with peak$=1$.}
\label{fig:flag_dictionary}
\end{figure}

\begin{figure*}
\centering
{\subfigure[{\em ridges} image.]{\includegraphics[width=0.24\linewidth]{ridges}}%
\hfil
\subfigure[NLSPCA. PSNR =  19.57dB]{\includegraphics[width=0.24\linewidth]{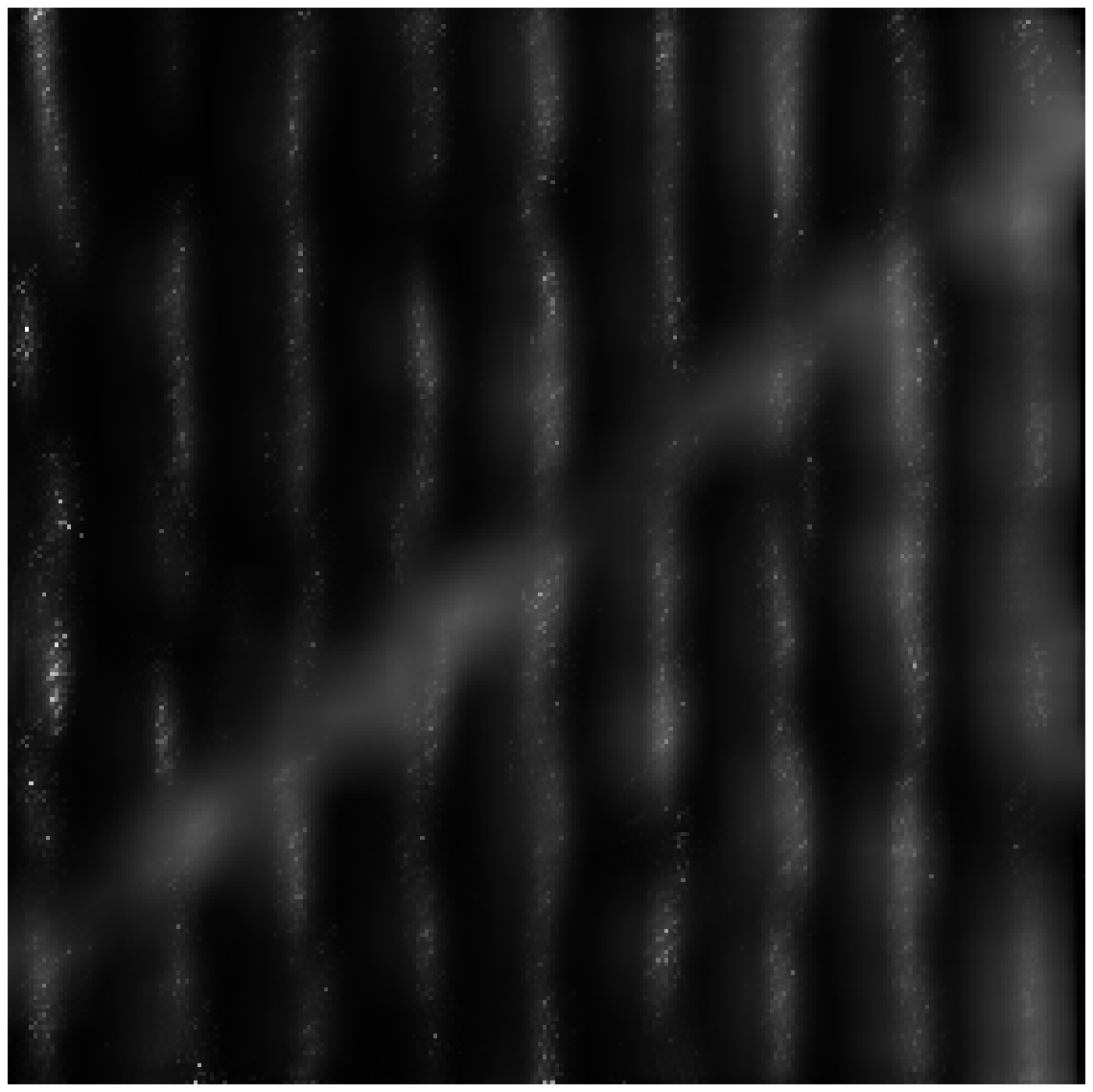}}
\hfil
\subfigure[BM3D. PSNR = 19.66dB.]{\includegraphics[width=0.24\linewidth]{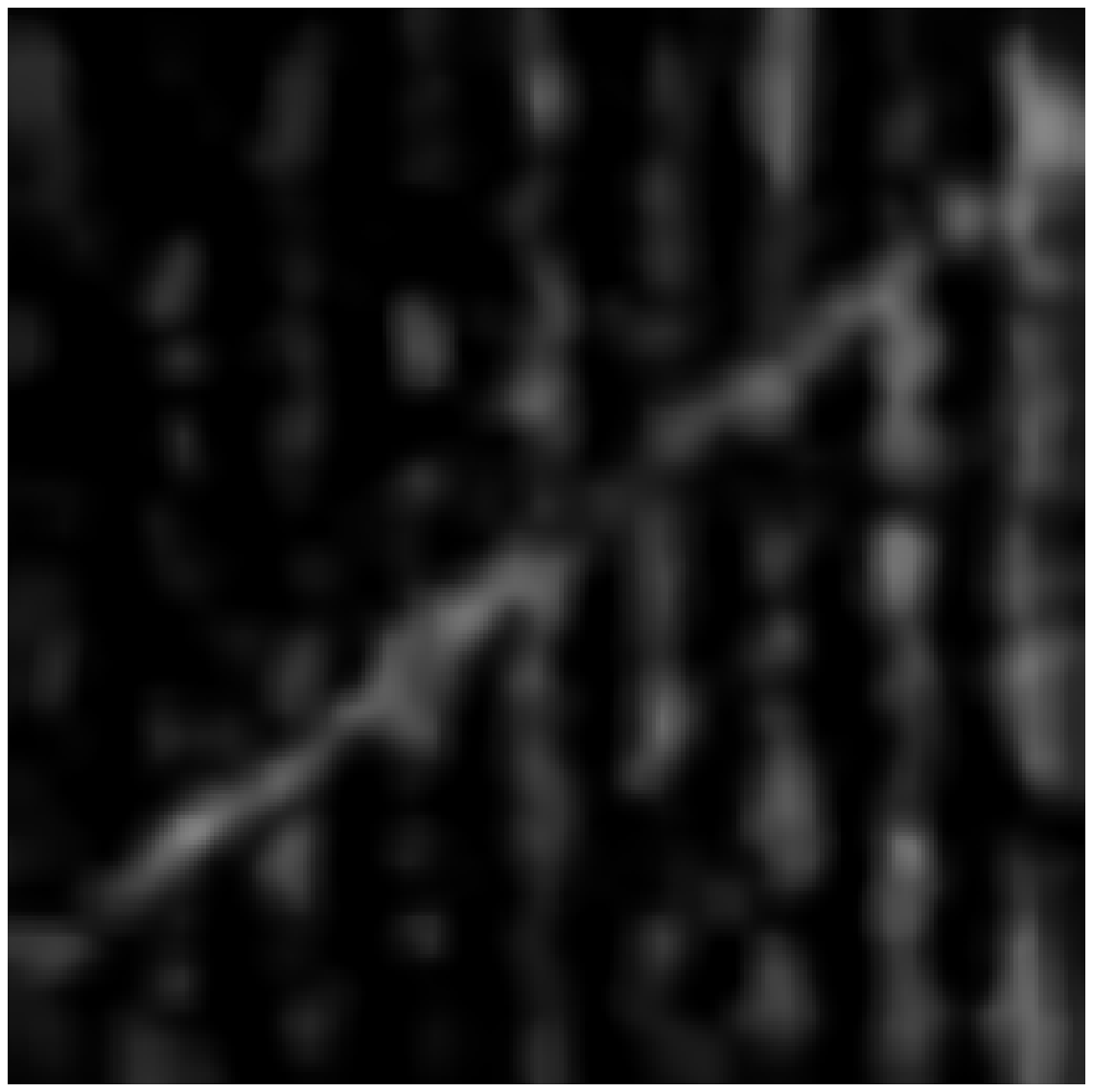}}
\hfil
\subfigure[SPDA. PSNR =  19.82dB.]{\includegraphics[width=0.24\linewidth]{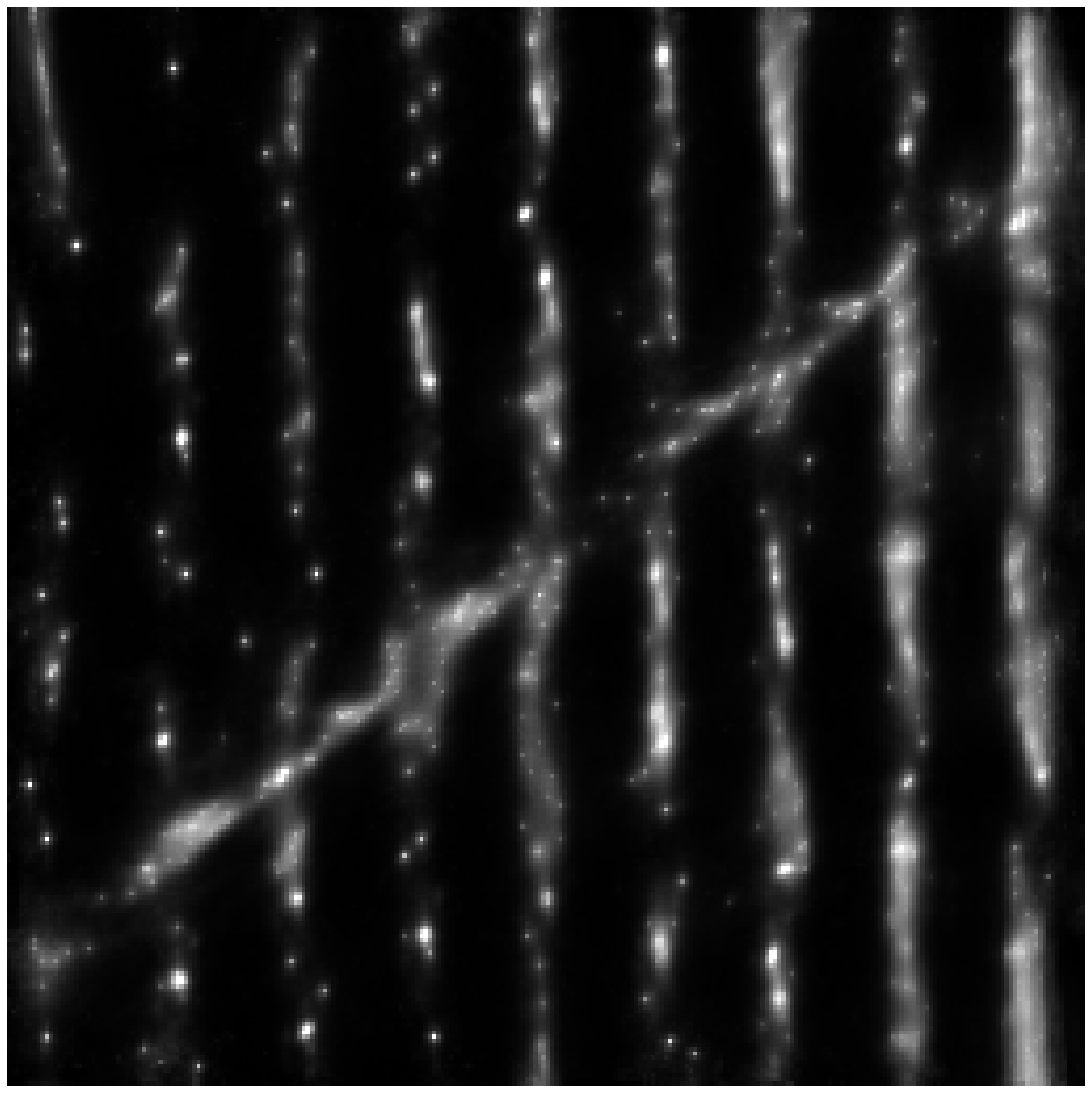}}
\vfil
\subfigure[Noisy image. Peak = 0.1.]{\includegraphics[width=0.24\linewidth]{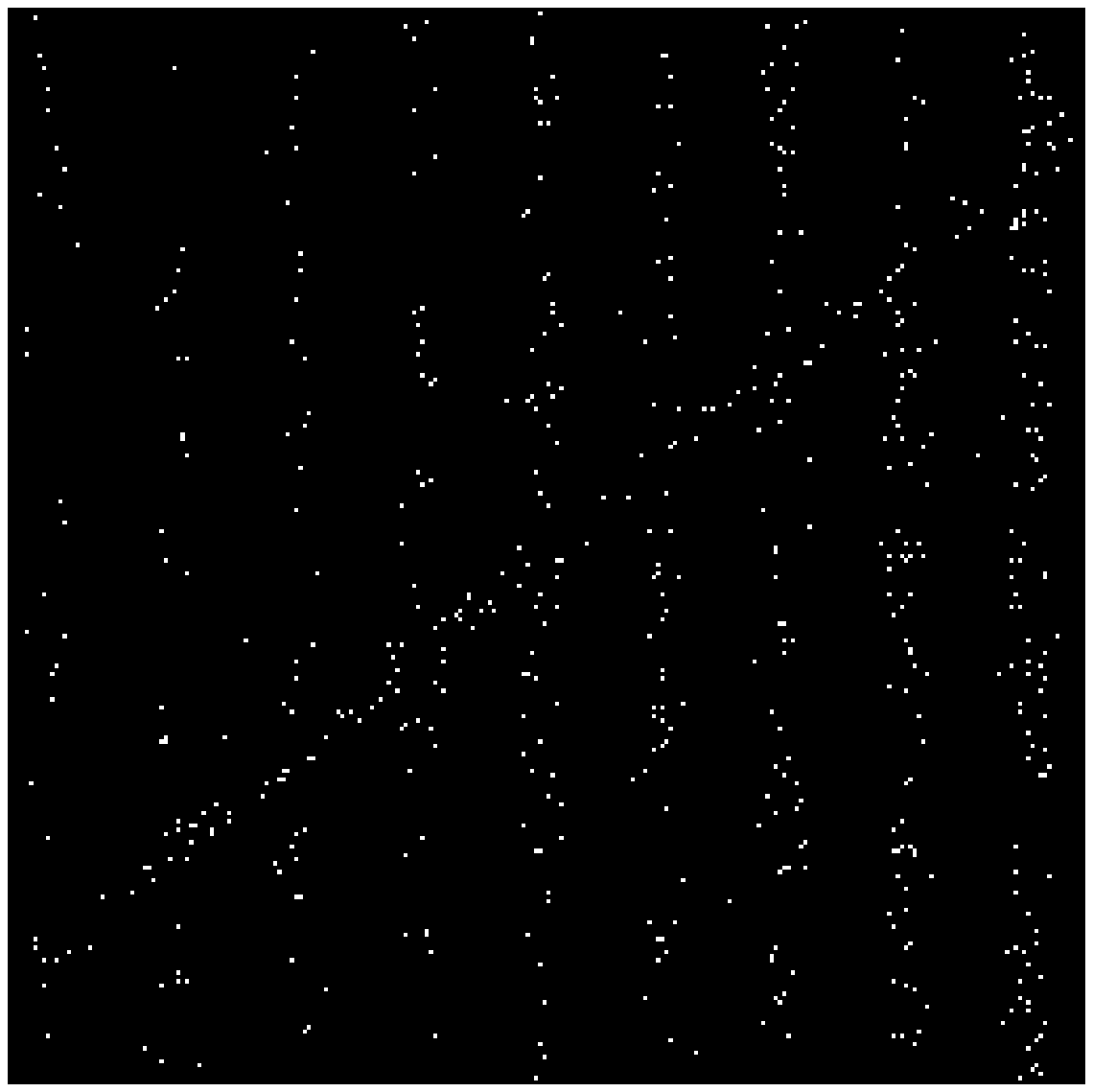}}
\hfil
\subfigure[NLSPCAbin. PSNR = 21.84dB.]{\includegraphics[width=0.24\linewidth]{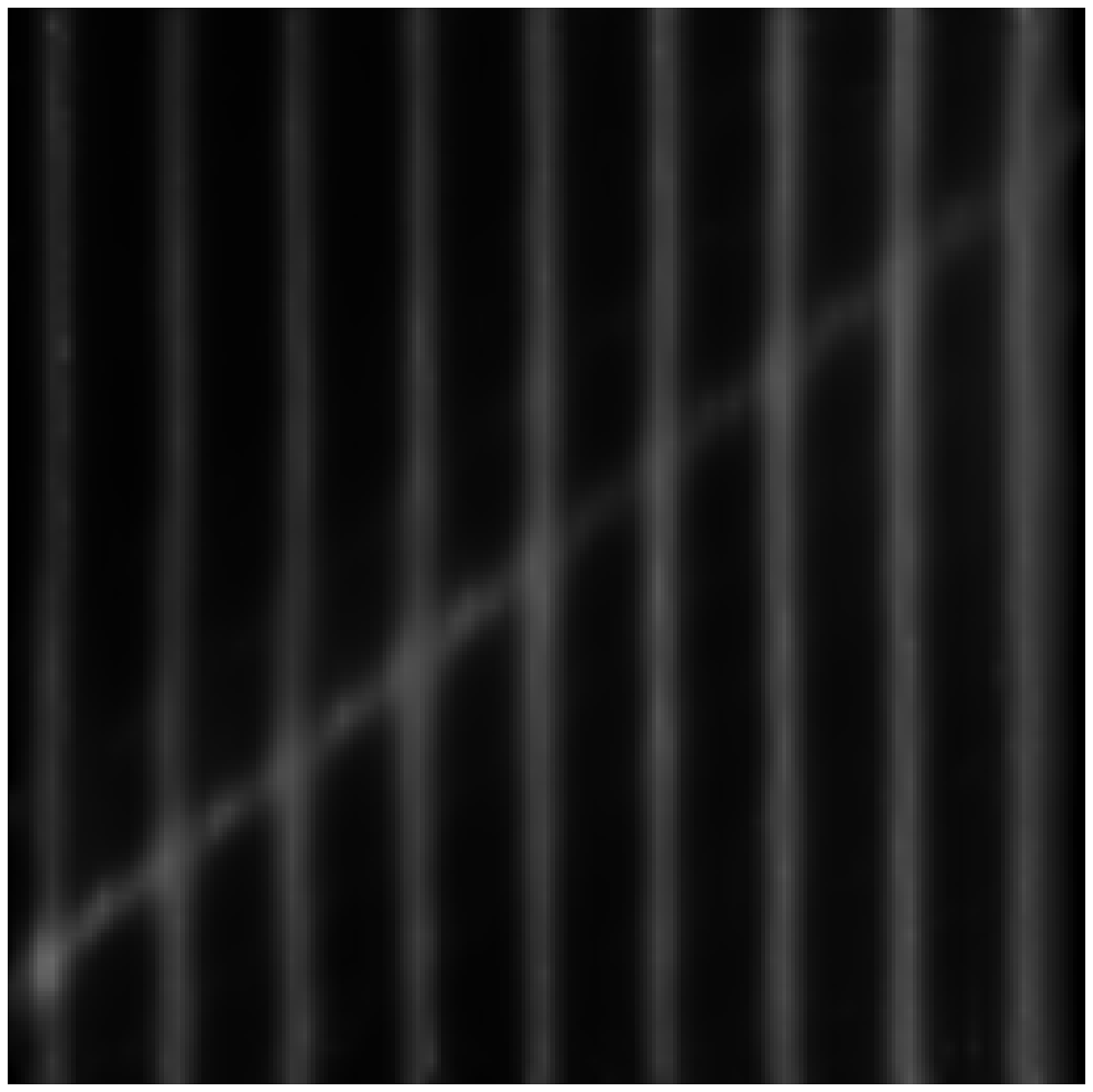}}
\hfil
\subfigure[BM3Dbin. PSNR = 18.98dB.]{\includegraphics[width=0.24\linewidth]{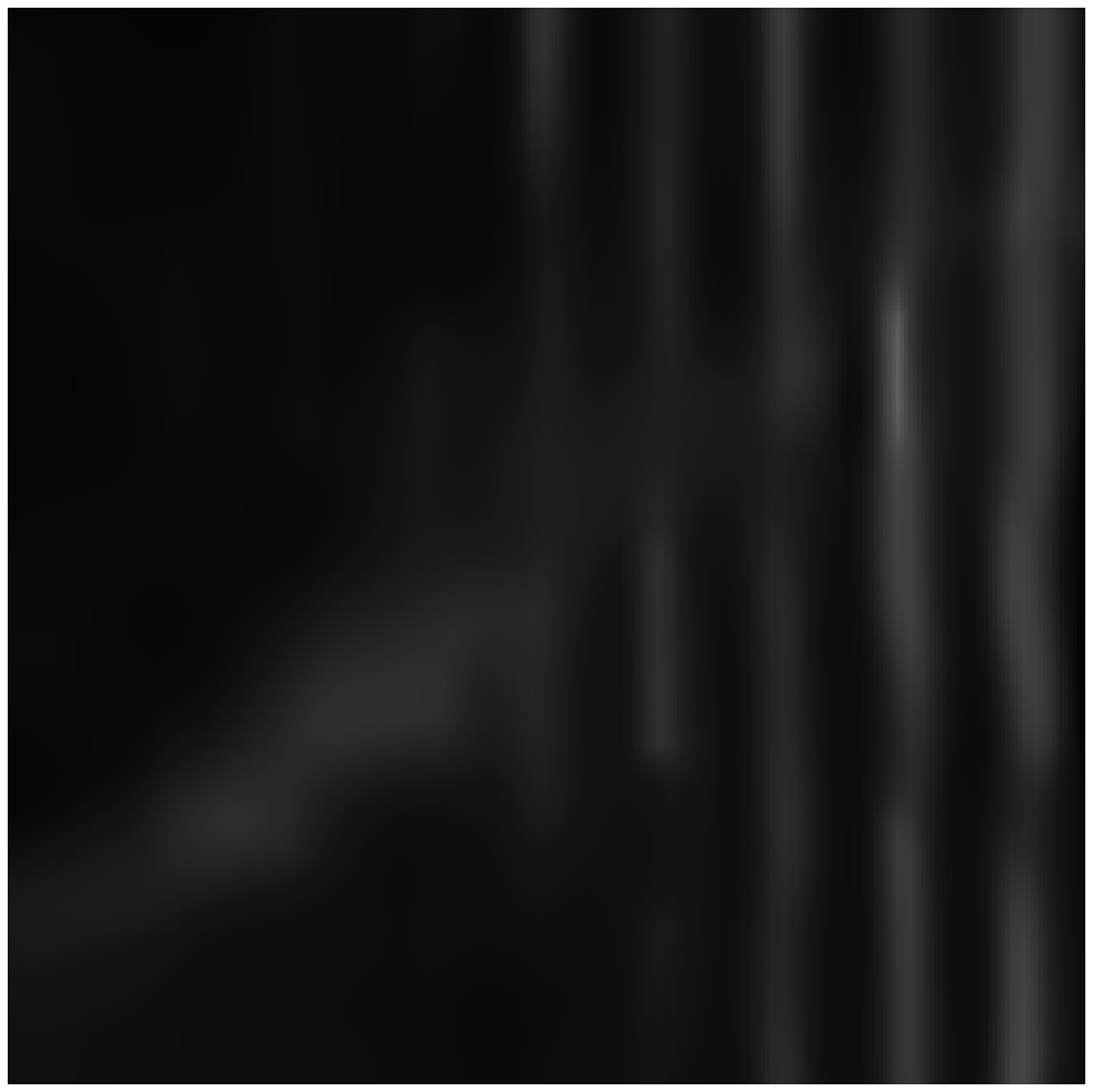}}
\hfil
\subfigure[SPDAbin. PSNR = {\bf 24.43dB}.]{\includegraphics[width=0.24\linewidth]{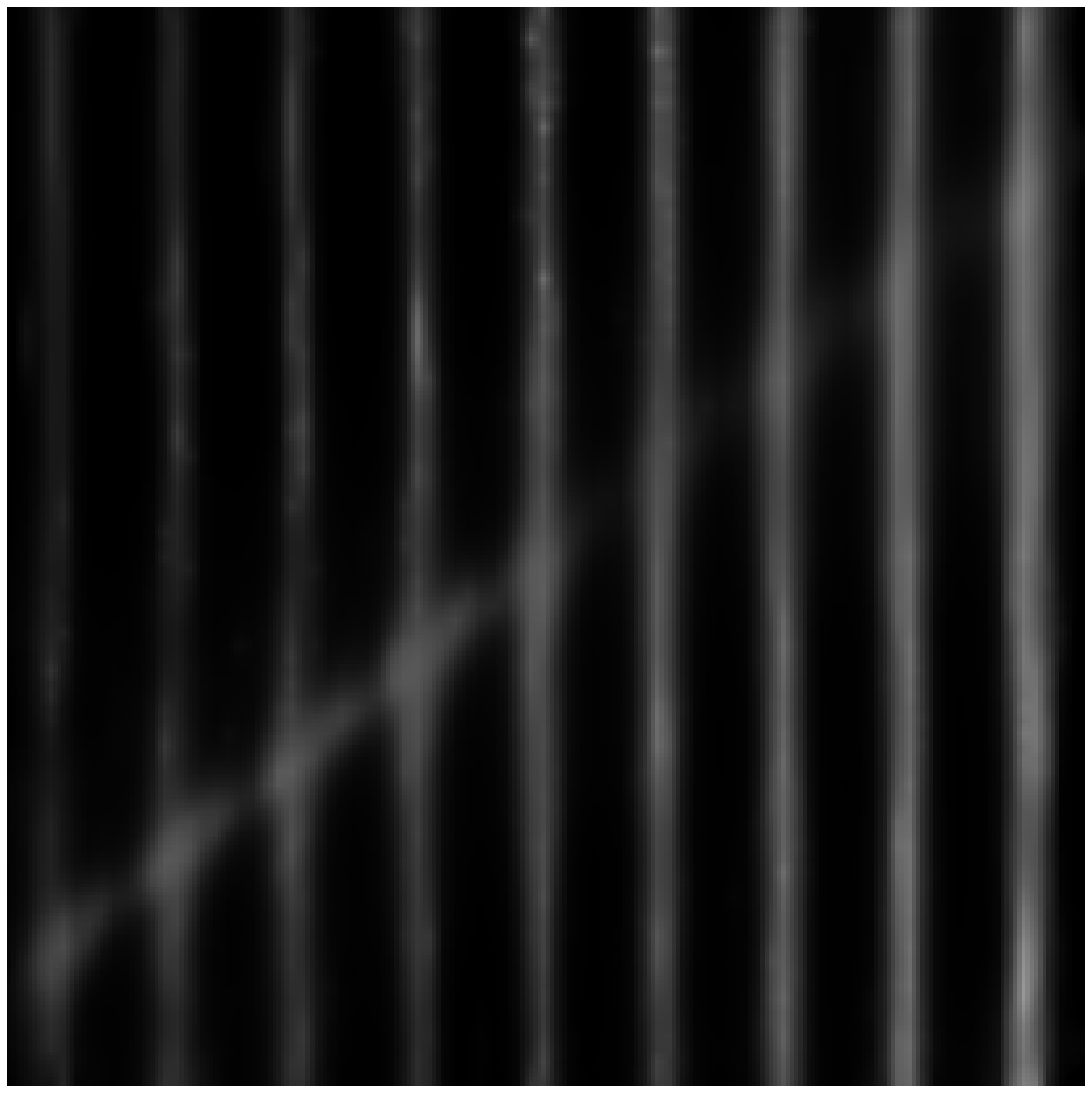}}
}%
\caption{Denoising of {\em ridges} with peak = 0.1. The PSNR is of the presented recovered images.}
\label{fig:ridges_recovery}
\end{figure*}

\begin{figure*}
\centering
{\subfigure[{\em Saturn} image.]{\includegraphics[width=0.24\linewidth]{saturn}}%
\hfil
\subfigure[NLSPCA. PSNR =  22.98dB]{\includegraphics[width=0.24\linewidth]{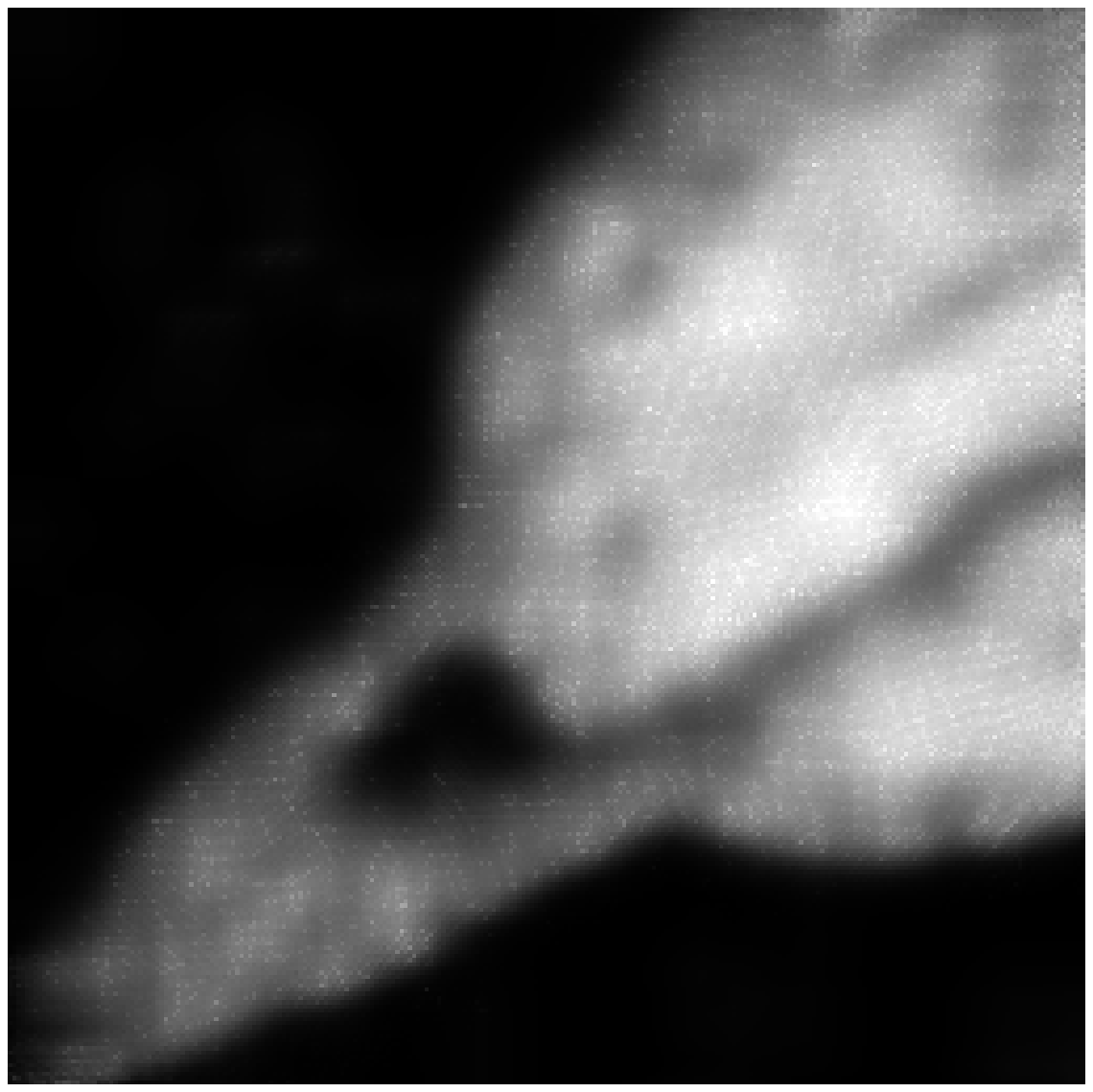}}
\hfil
\subfigure[BM3D. PSNR =  21.71dB.]{\includegraphics[width=0.24\linewidth]{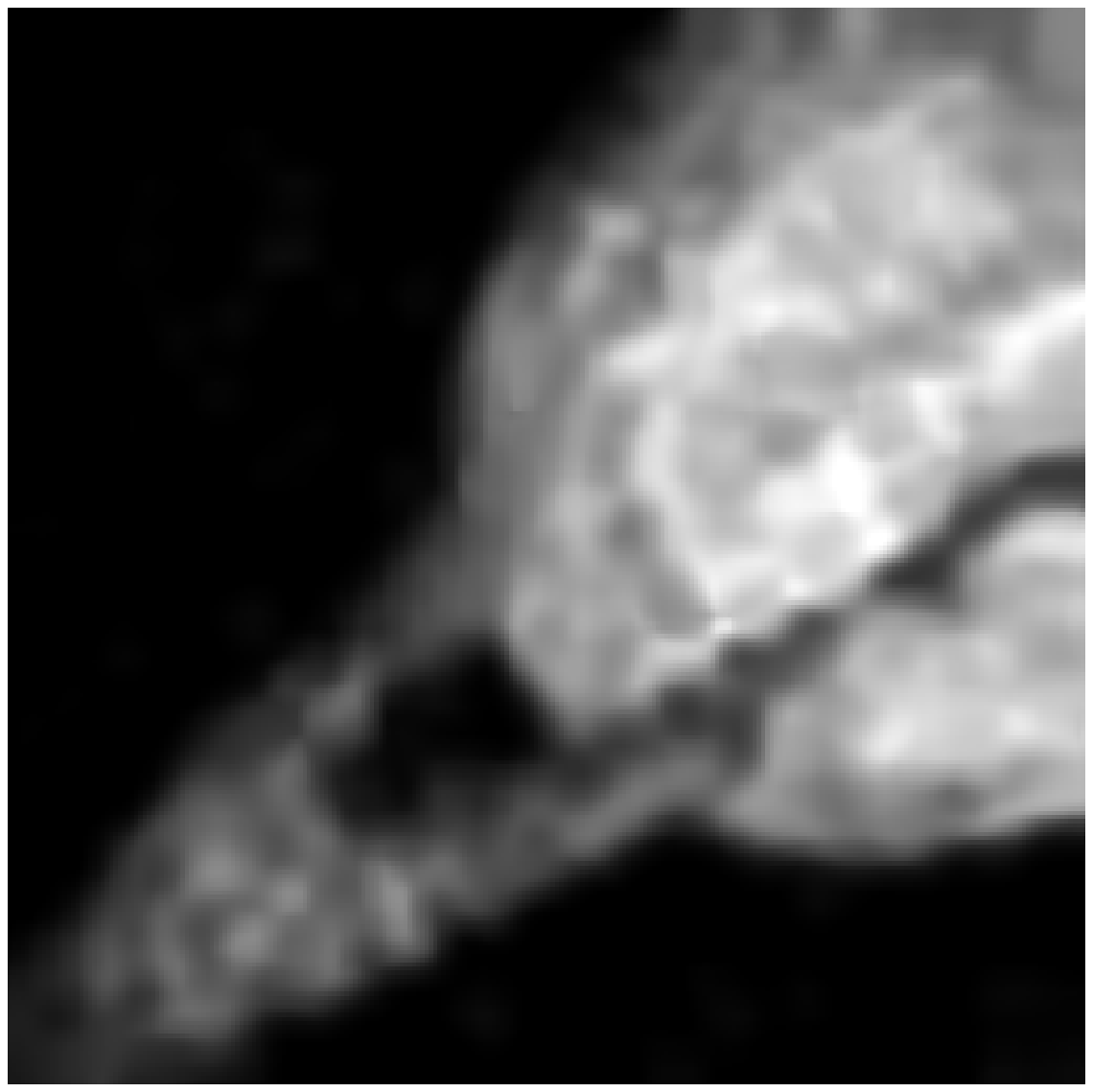}}
\hfil
\subfigure[SPDA. PSNR =  21.47dB.]{\includegraphics[width=0.24\linewidth]{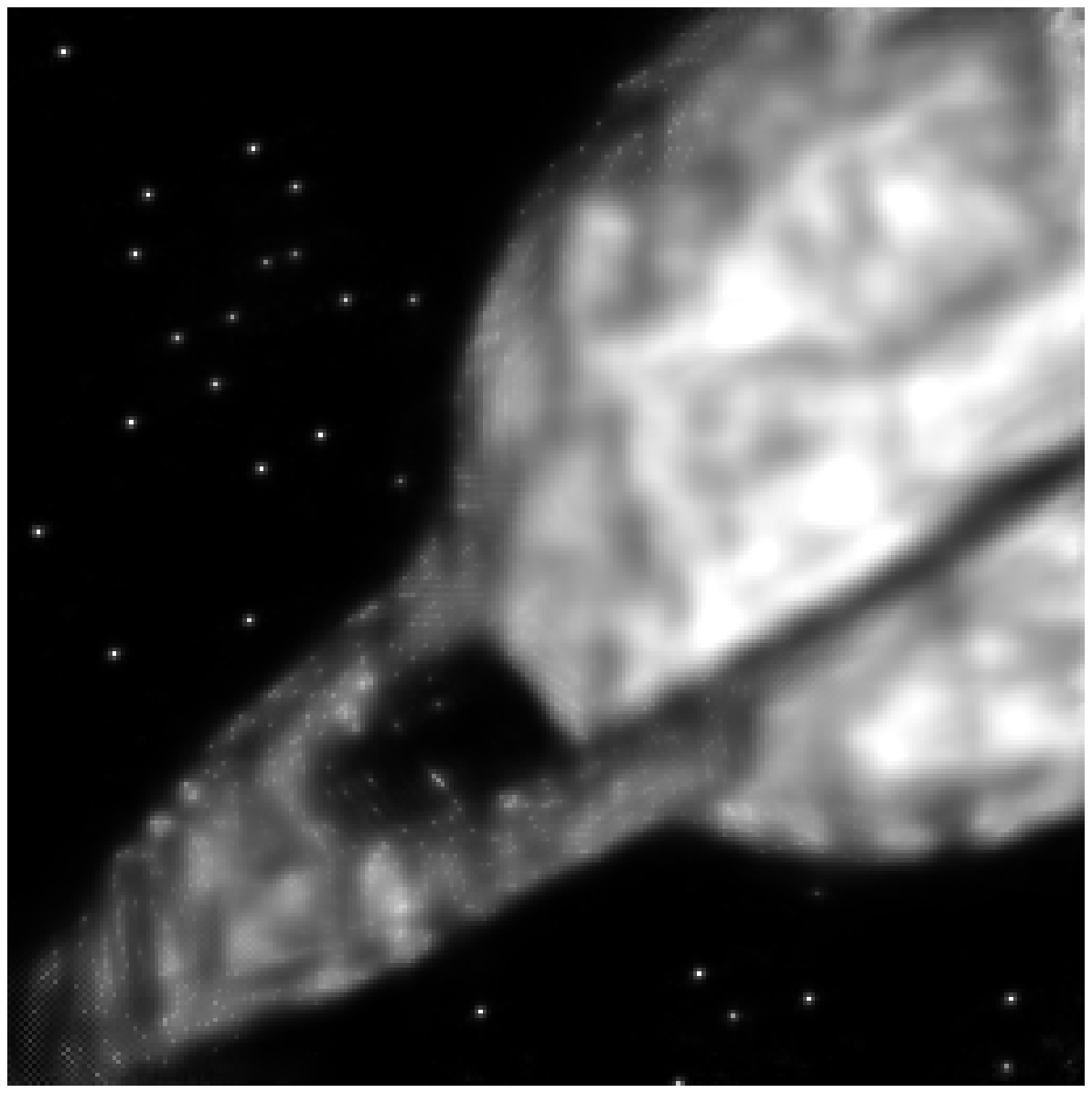}}
\vfil
\subfigure[Noisy image. Peak = 0.2.]{\includegraphics[width=0.24\linewidth]{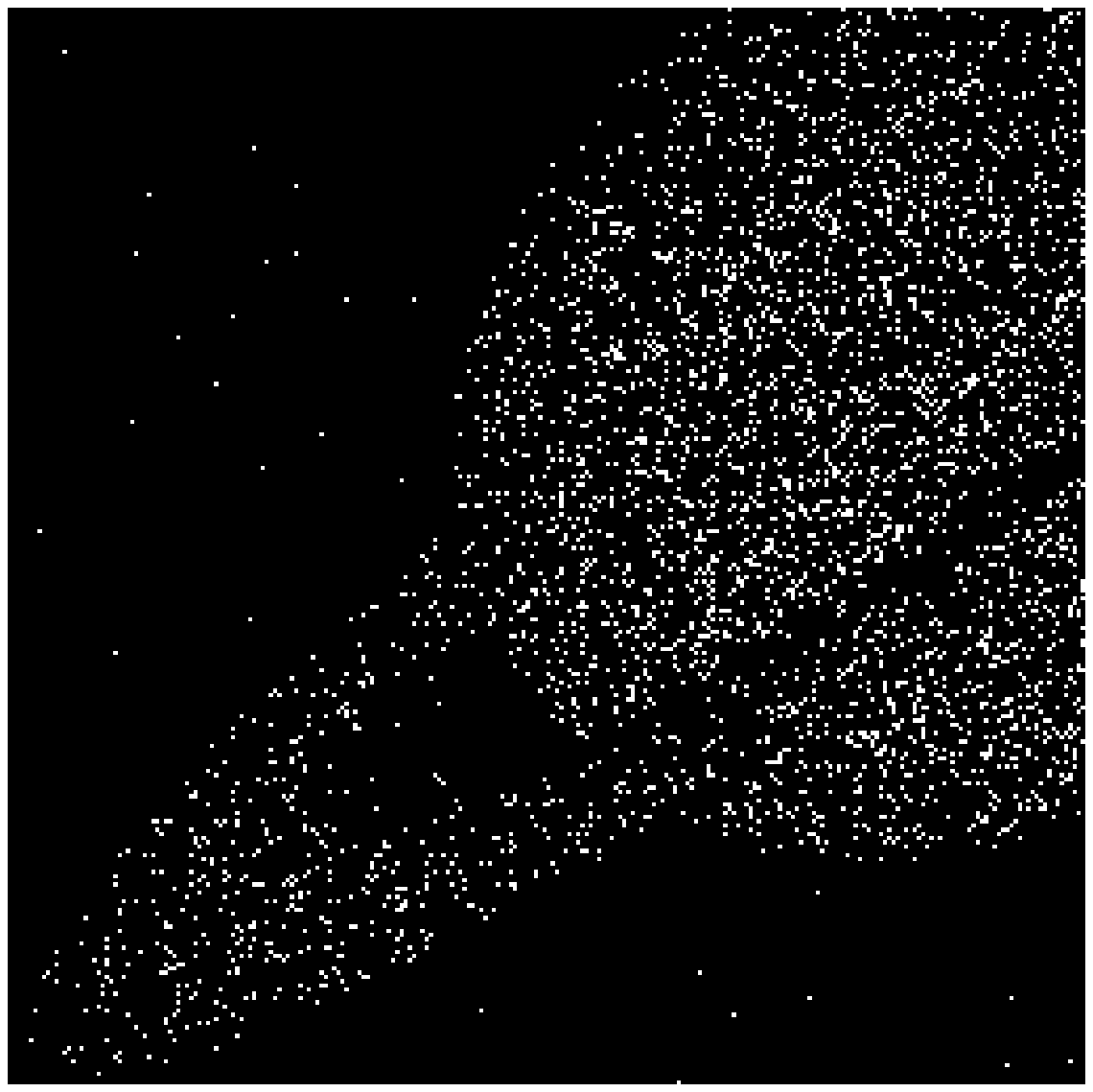}}
\hfil
\subfigure[NLSPCAbin. PSNR =  20.35dB.]{\includegraphics[width=0.24\linewidth]{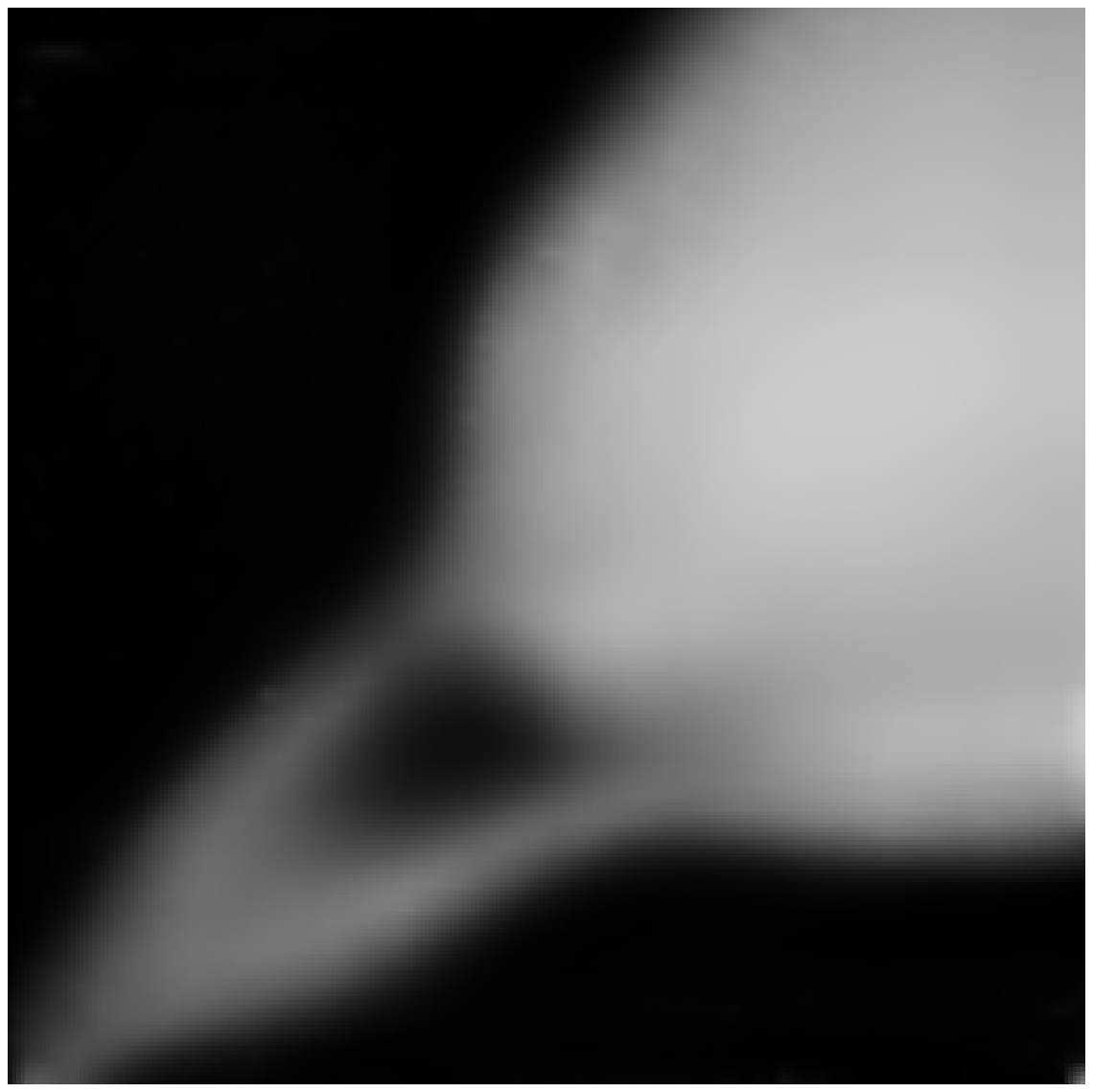}}
\hfil
\subfigure[BM3Dbin. PSNR =  23.16dB.]{\includegraphics[width=0.24\linewidth]{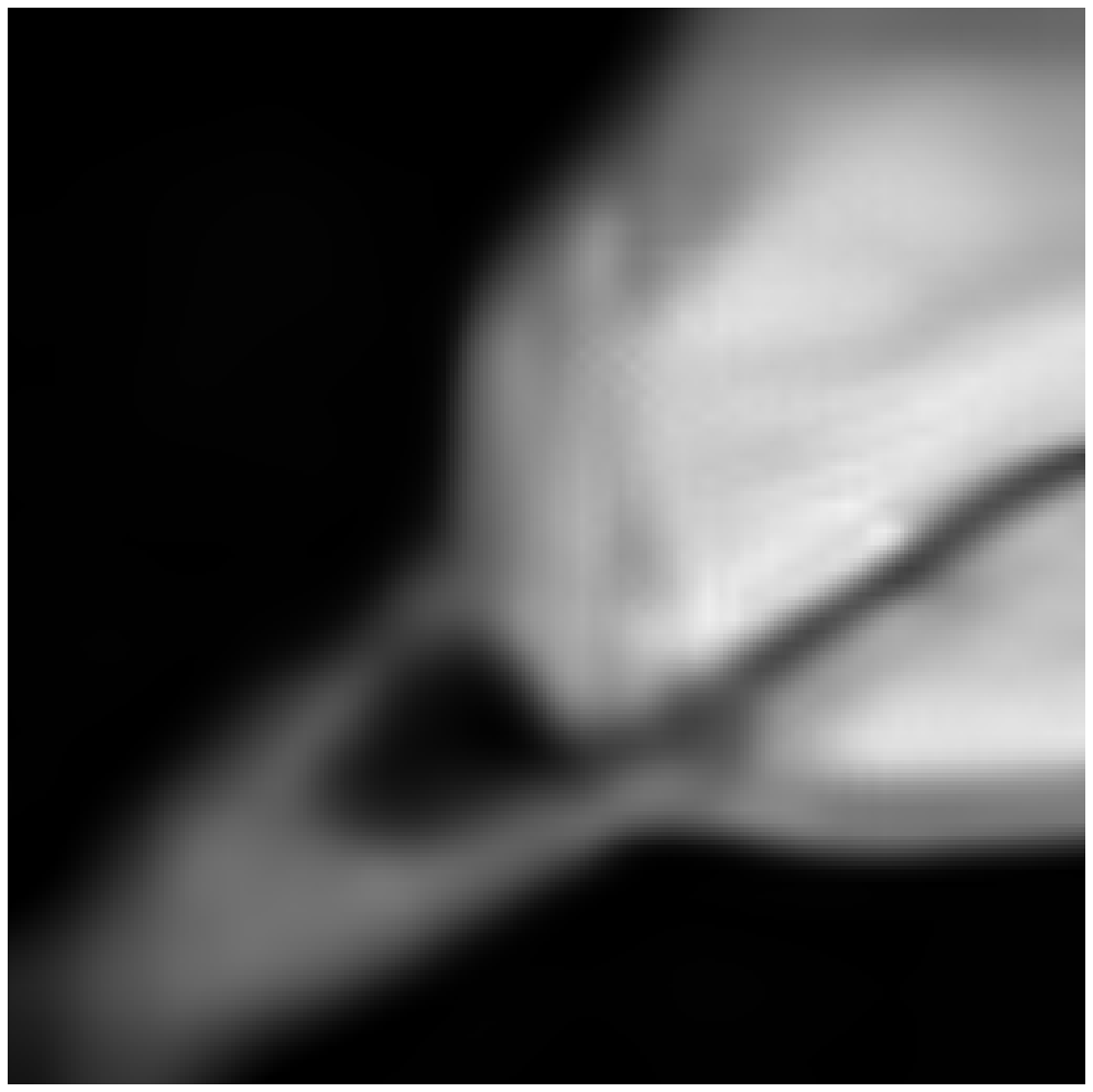}}
\hfil
\subfigure[SPDAbin. PSNR =    {\bf24.35dB}.]{\includegraphics[width=0.24\linewidth]{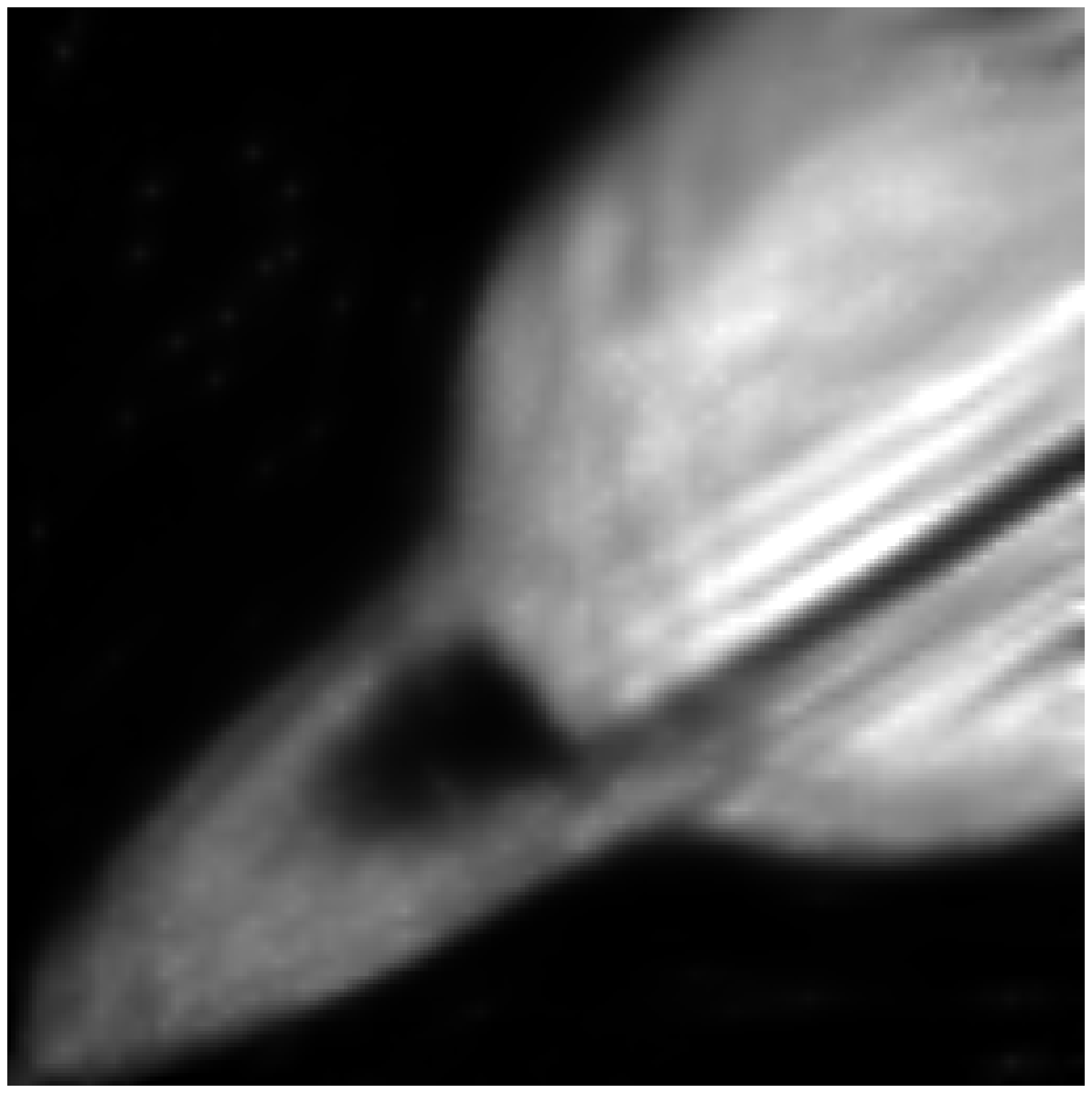}}
}%
\caption{Denoising of {\em Saturn} with peak = 0.2. The PSNR is of the presented recovered images.}
\label{fig:saturn_recovery}
\end{figure*}

\begin{figure*}
\centering
{\subfigure[{\em house} image.]{\includegraphics[width=0.24\linewidth]{house}}%
\hfil
\subfigure[NLSPCA. PSNR =    23.23dB]{\includegraphics[width=0.24\linewidth]{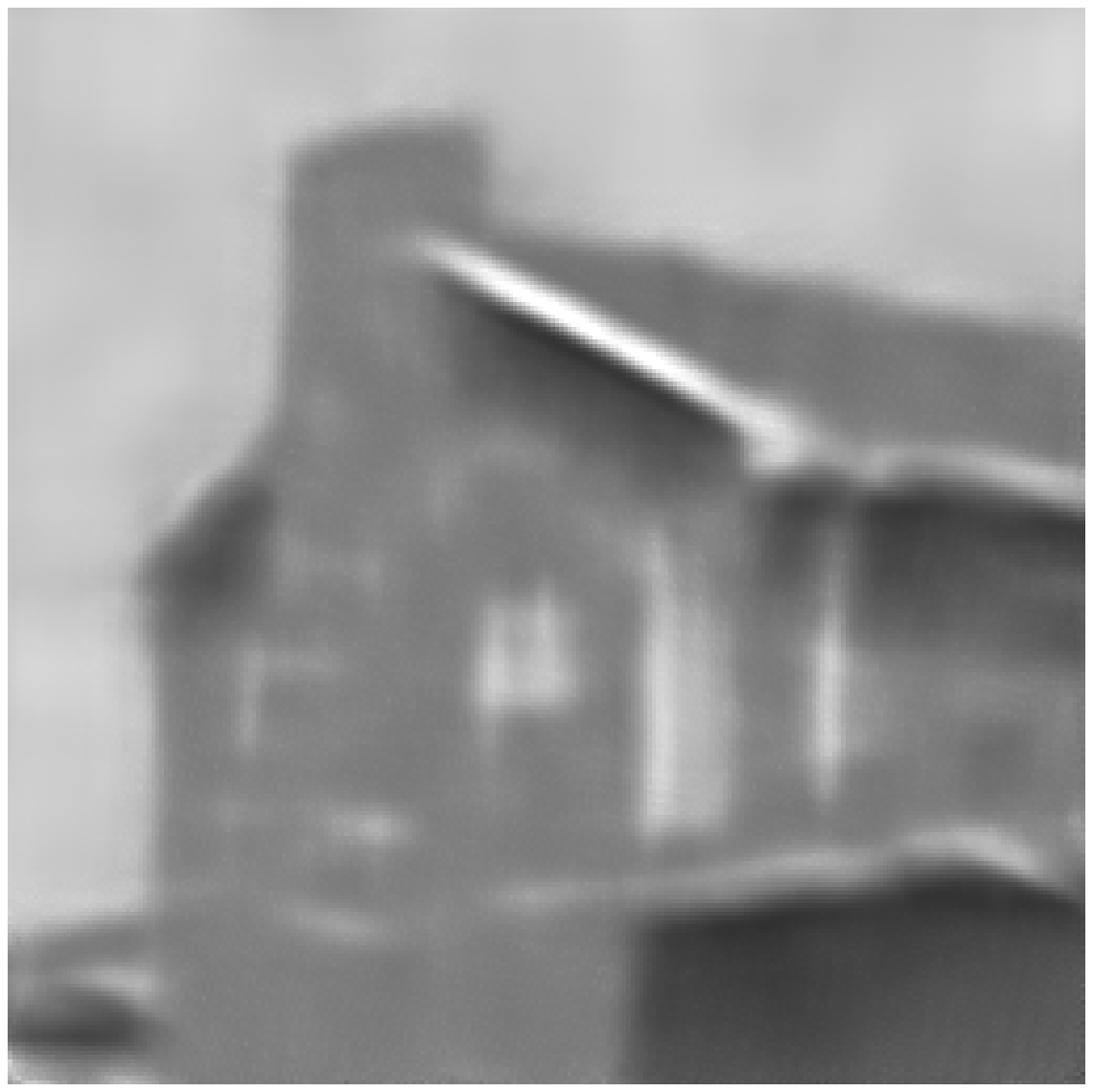}}
\hfil
\subfigure[BM3D. PSNR =  24.06dB.]{\includegraphics[width=0.24\linewidth]{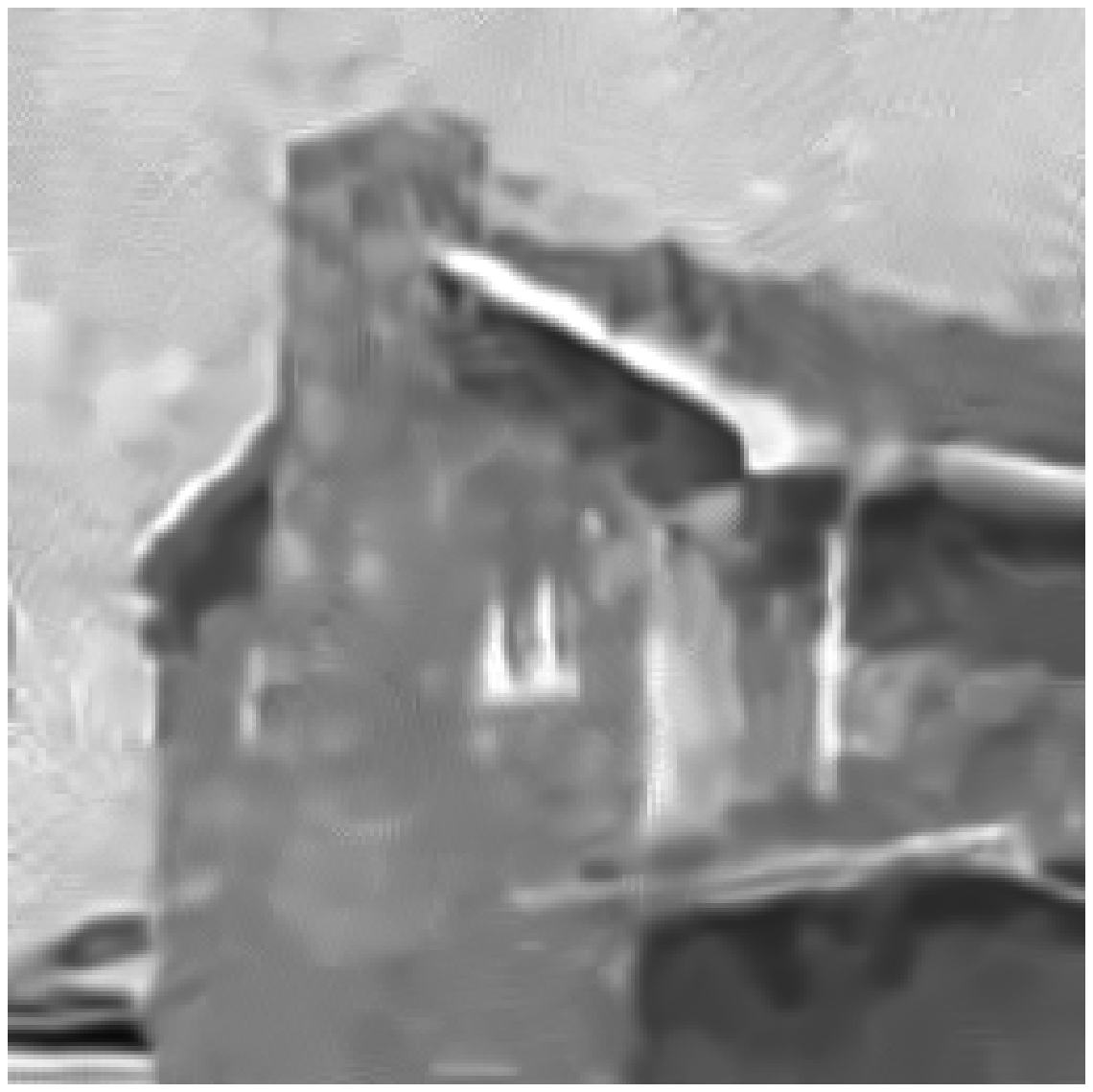}}
\hfil
\subfigure[SPDA. PSNR =    {\bf 24.8dB}.]{\includegraphics[width=0.24\linewidth]{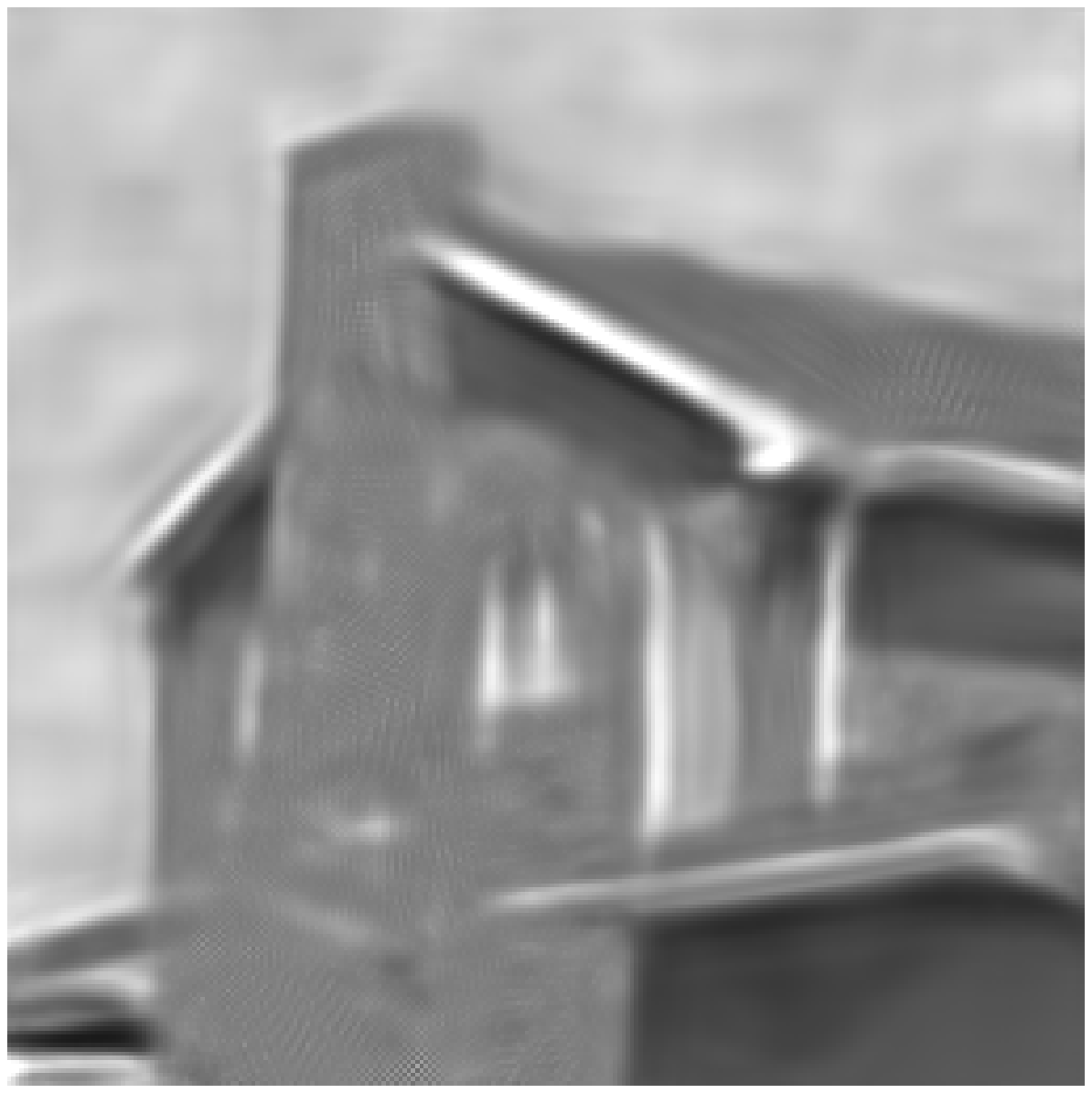}}
\vfil
\subfigure[Noisy image. Peak = 2.]{\includegraphics[width=0.24\linewidth]{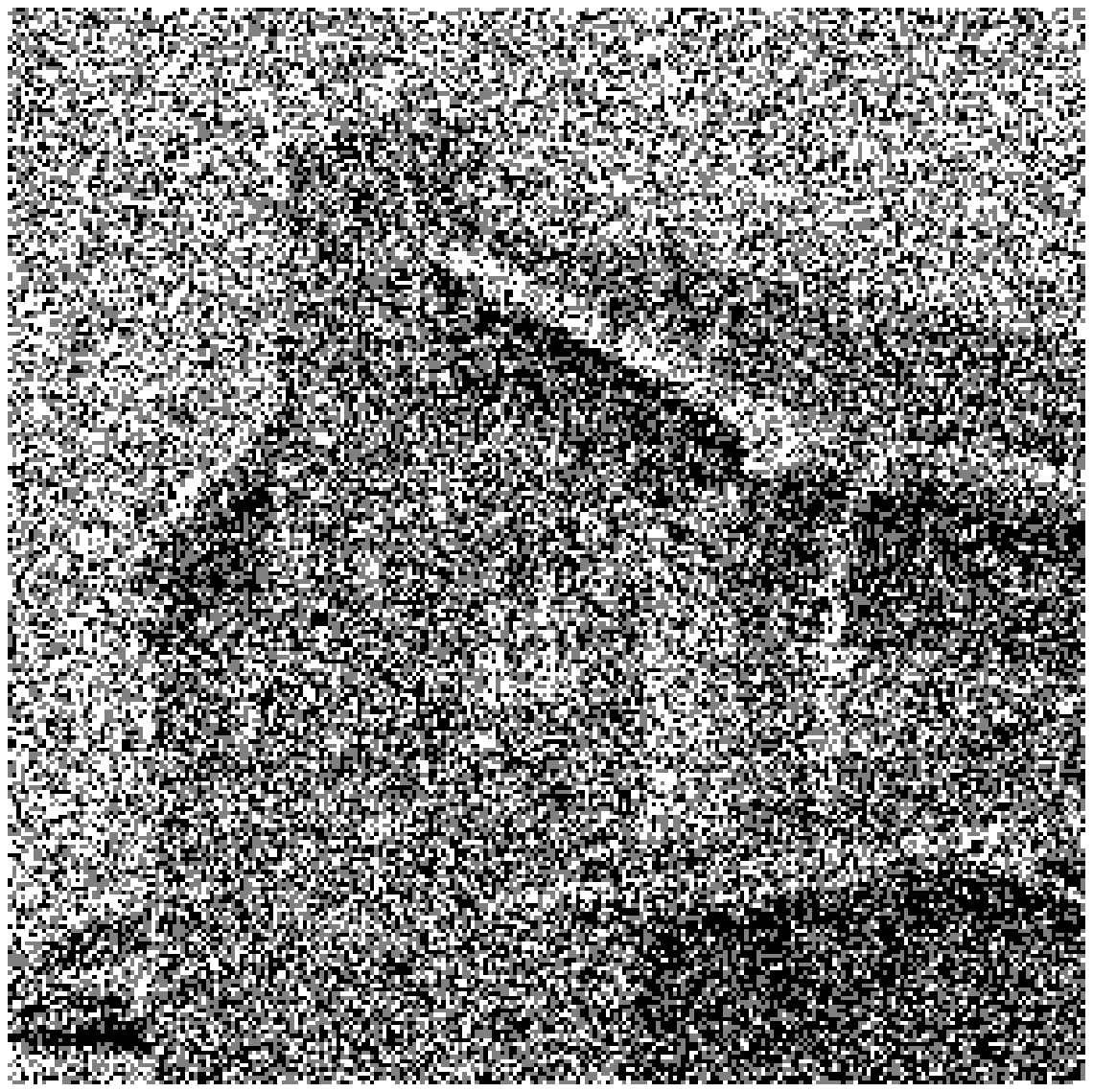}}
\hfil
\subfigure[NLSPCAbin. PSNR = 21.28dB.]{\includegraphics[width=0.24\linewidth]{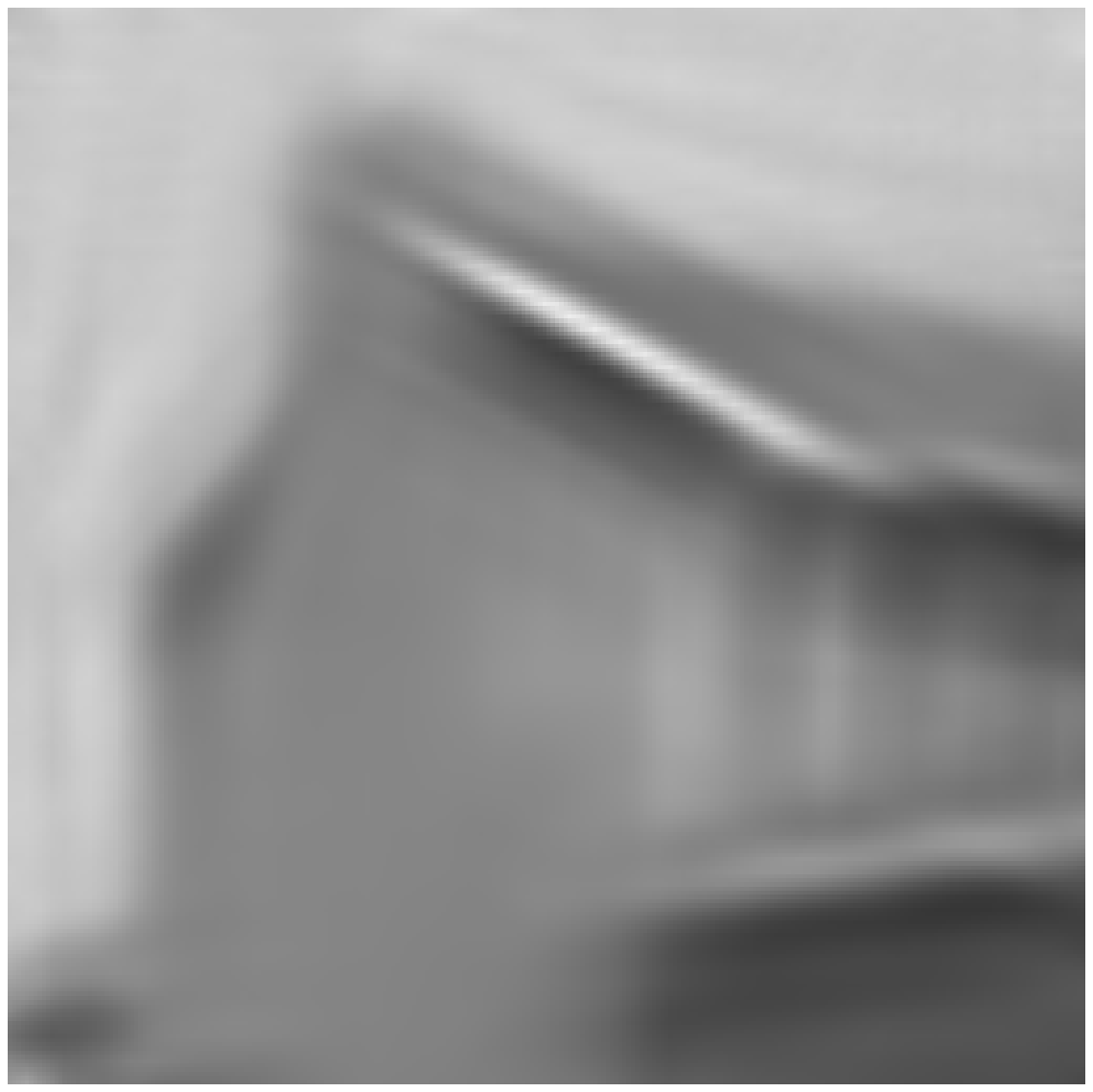}}
\hfil
\subfigure[BM3Dbin. PSNR = 24.23dB.]{\includegraphics[width=0.24\linewidth]{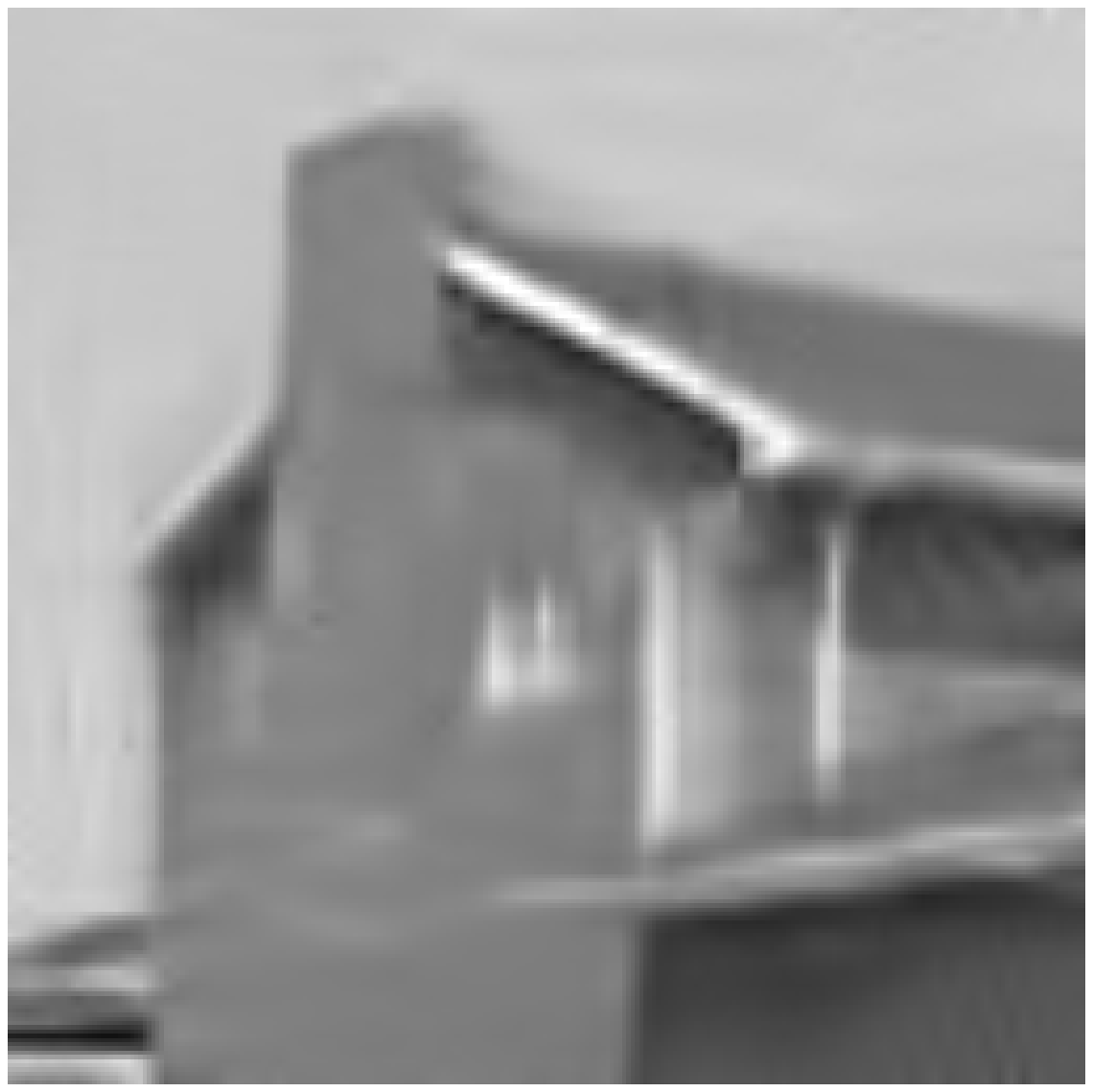}}
\hfil
\subfigure[SPDAbin. PSNR =  21.51dB.]{\includegraphics[width=0.24\linewidth]{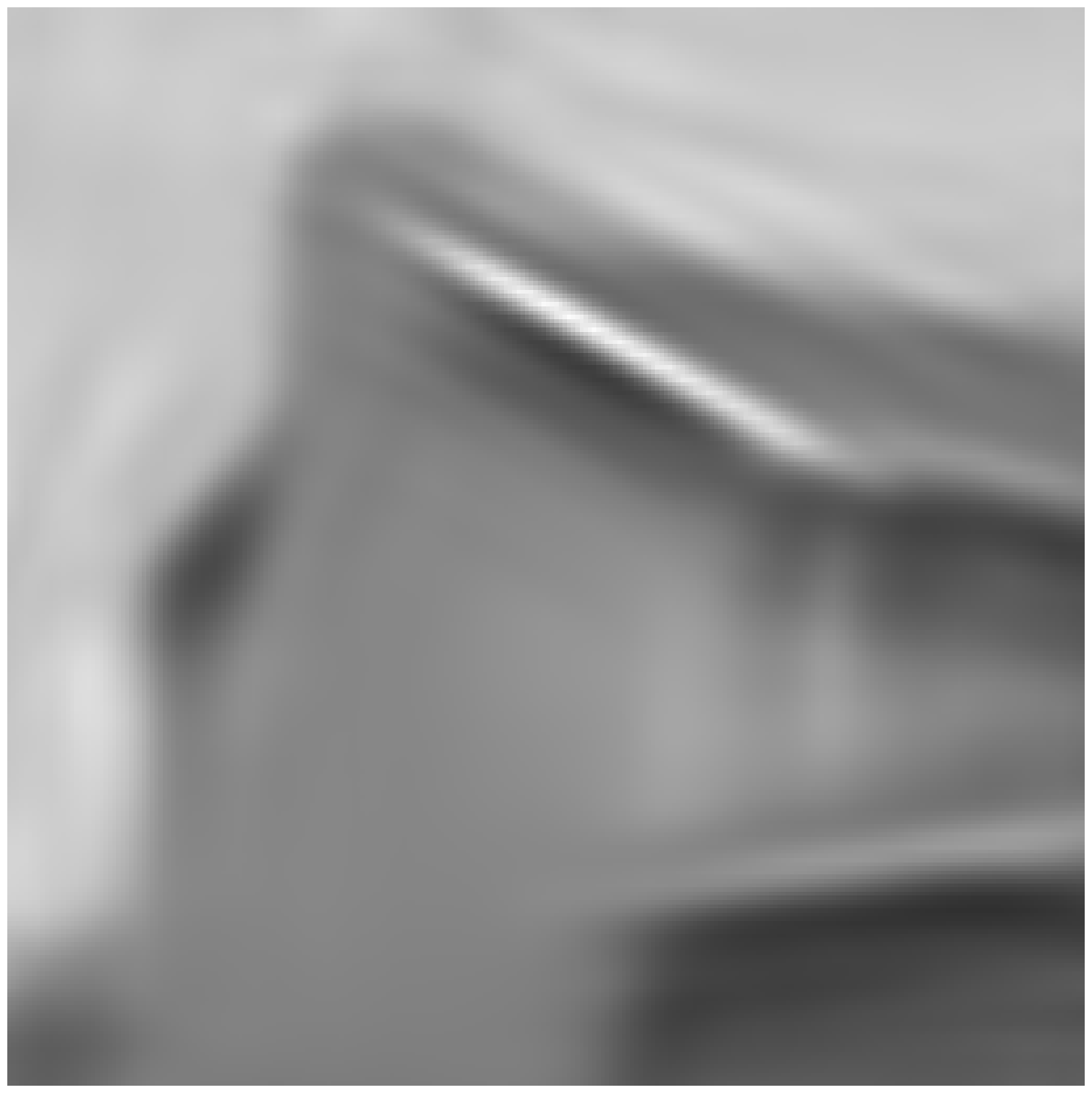}}
}%
\caption{Denoising of {\em house} with peak = 2. The PSNR is of the presented recovered images.}
\label{fig:house_recovery}
\end{figure*}

A comparison between the recovery of the {\em flag} image with peak$=1$ is presented in Fig~\ref{fig:flag_recovery}.
It can be observed that this image is recovered very accurately while in the other methods there are many artifacts.
Samples from the dictionary atoms learned by SPDA are presented in Fig.~\ref{fig:flag_dictionary}.
It can be observed that the dictionary learning process captures the shape of the stars and the lines in the flag.
Figures~\ref{fig:ridges_recovery}, \ref{fig:saturn_recovery} and \ref{fig:house_recovery} present the recovery of {\em ridges}, {\em Saturn} and {\em house} for
peak$=0.2$ and peak$=2$ respectively. It can be seen that for the low peak value the binning methods capture the structure
of the image better and provide lower error. However, when the peak is higher the binning provides degraded performance
compared to the reconstruction using the original image. In all images the SPDA recovers the images' details better.

\begin{table*}
\footnotesize
\begin{center}
\begin{tabular}{|l|c|c|c|c|c|c|c|c|c|c|c|}
\hline
Method & Peak & {Saturn} & {Flag} & {Camera} & {House} & {Swoosh} & {Peppers} & {Bridge} & {Ridges} & & Average\\
\hline\hline
NLSPCA    & & 20.86 & 14.42 & 16.41 & 17.81 & 19.11 & 16.24 & 16.59 & 20.92 && 17.8 \\
NLSPCAbin & & 19.13 & \bf{16.07} & \bf{17.15} & 18.71 & 21.89 & 16.12 & 16.93 & 24.05 && 18.75 \\
BM3D  & 0.1 & 19.42 & 13.05 & 15.66 & 16.28 & 16.93 & 15.61 & 15.68 & 20.06 &&  16.6\\
BM3Dbin   & & 21.19 & 14.23 & 16.91 & 18.62 & 21.90 & 15.92 & 16.91 & 20.40 &&  18.26 \\
SPDA (our work)      & & 17.40 & 13.35 & 14.36 & 14.84 & 15.12 & 14.28 & 14.60 & 19.86 && 15.48  \\
SPDAbin (our work)   & & \bf{22.00} & 15.40 & 16.75 & \bf{18.73} & \bf{21.90} & \bf{16.27} & \bf{16.99} & \bf{25.32} && \bf{19.17}  \\
\hline
NLSPCA    & & 22.90 & 16.48 & 17.79 & 18.91 & 21.10 & 17.45 & 17.46 & 24.22 &&19.54\\
NLSPCAbin & & 20.54 & 16.54 & 17.92 & \bf{19.74} & 24.00 & 16.92 & 17.57 & 25.91 &&  19.89\\
BM3D  & 0.2 & 22.02 & 14.28 & 17.35 & 18.37 & 19.95 & 17.10 & 17.09 & 21.27 &&  18.45\\
BM3Dbin   & & 23.20 & 16.28 & \bf{18.25} & 19.71 & \bf{24.25} & 17.44 & 17.70 & 23.92 &&  20.09\\
SPDA (our work)      && 21.52 & 16.58 & 16.93 & 17.83 & 18.91 & 16.75 & 16.80 & 23.25 && 18.57   \\
SPDAbin (our work)   && \bf{23.99} & \bf{18.26} & 17.95 & 19.62 & 23.53 & \bf{17.59} & \bf{17.82} & \bf{27.22} && \bf{20.75}  \\
\hline
NLSPCA    & & 24.91 & 18.80 & 19.23 & 20.85 & 23.80 & 18.78 & 18.50 & 28.20 && 21.63 \\
NLSPCAbin & & 20.98 & 17.10 & 18.32 & 20.98 & 26.48 & 17.77 & 18.18 & 26.81 && 20.83 \\
BM3D  & 0.5 & 23.86 & 15.87 & 18.83 & 20.27 & 22.92 & 18.49 & 18.24 & 23.37 && 20.29 \\
BM3Dbin   & & 25.70 & 18.40 & \bf{19.64} & \bf{21.71} & 26.33 & \bf{19.01} & \bf{18.67} & 28.23 && 22.21\\
SPDA (our work)      && 25.50 & \bf{19.67} & 18.90 & 20.51 & 24.21 & 18.66 & 18.46 & 27.76 && 21.71 \\
SPDAbin (our work)   && \bf{25.83} & 19.22 & 18.97 & 21.15 & \bf{26.57} & 18.63 & 18.57 & \bf{30.97} && \bf{22.49} \\ 
\hline
NLSPCA    & & 26.89 & 20.26 & 20.32 & 22.09 & 27.42 & 19.62 & 18.94 & 30.57 && 23.26 \\
NLSPCAbin & & 21.18 & 17.07 & 18.50 & 21.26 & 27.62 & 17.80 & 18.20 & 27.58 && 21.15\\
BM3D    & 1 & 25.89 & 18.31 & 20.37 & 22.35 & 26.07 & 19.89 & 19.22 & 26.26 && 22.42\\
BM3Dbin   & & \bf{27.41} & 19.33 & \bf{20.60} & \bf{23.19} & \bf{28.44} & \bf{20.13} & \bf{19.38} & 30.50 && \bf{23.62}\\
SPDA  (our work)     & & 27.02 & \bf{22.54} & 20.23 & 22.73 & 26.28 & 19.99 & 19.20 & 30.93 && 23.61\\
SPDAbin (our work)   & & 27.26 & 19.88 & 19.45 & 21.63 & 28.31 & 18.92 & 18.74 & \bf{32.41} && 23.33  \\
\hline
NLSPCA    & & 28.22 & 20.86 & 20.76 & 23.86 & 29.62 & 20.52 & 19.47 & 31.87 && 24.4\\
NLSPCAbin & & 21.49 & 16.85 & 18.43 & 21.41 & 27.88 & 17.80 & 18.34 & 28.68 && 21.36\\
BM3D    & 2 & 27.42 & 20.81 & \bf{22.13} & 24.18 & 28.09 & \bf{21.97} & \bf{20.31} & 29.82 && 24.59\\
BM3Dbin   & & 28.84 & 20.02 & 21.37 & 24.49 & \bf{29.74} & 21.16 & 20.17 & 32.06 && 24.73\\
SPDA  (our work)     & & \bf{29.38} & \bf{24.92} & 21.54 & \bf{25.09} & 29.27 & 21.23 & 20.15 & 33.40 && \bf{25.62} \\
SPDAbin (our work)   && 28.51 & 19.81 & 19.61 & 21.72 & 28.98 & 19.26 & 18.93 & \bf{33.51} && 23.79  \\ 
\hline
NLSPCA    & & 29.44 & 21.25 & 21.09 & 24.89 & 31.30 & 21.12 & 20.16 & 34.01 && 25.41\\
NLSPCAbin & & 21.20 & 16.50 & 18.45 & 21.44 & 28.01 & 17.82 & 18.34 & 29.09 && 21.36 \\
BM3D    & 4 & 29.40 & 23.04 & \bf{23.94} & 26.04 & 30.72 & \bf{24.07} & 21.50 & 32.39 && 26.89\\
BM3Dbin   & & 30.19 & 20.51 & 21.98 & 25.48 & 31.30 & 22.09 & \bf{20.80} & 33.55 && 25.74\\
SPDA  (our work)     && \bf{31.04} & \bf{26.27} & 21.90 & \bf{26.09} & \bf{33.20} & 22.09 & 20.55 & \bf{36.05} && \bf{27.15} \\
SPDAbin (our work)   && 29.43 & 19.85 & 20.12 & 22.87 & 30.93 & 20.37 & 19.24 & 34.44 && 24.66 \\ 
\hline
\end{tabular}
\end{center}
\caption{Experiments on simulated data (average over five noise realizations).
The images are the same as those in \cite{Salmon12Poisson}.
Best results are marked.}
\label{tbl:exp_psnr}
\end{table*}

The recovery error in terms of PSNR for all images and different peak values appears in Table~\ref{tbl:exp_psnr}.
By looking at the overall performance we can see that our proposed strategy provides better performance on average for $5$ of $6$ tested peak-values. Note that even in the case where it does not behave better, the difference is insignificant ($0.02dB$ for peak$=1$).

For low peak values SPDAbin behaves better and for larger ones SPDA should be preferred. 
The addition of the binning to the algorithms improves their performance significantly in the lower peak values.
As the peak raises, the effectiveness of the binning reduces. Note that the efficiency reduces slower for BM3D. This might be explained by the fact that it relies on the Anscombe that becomes much more effective when the peak increases.

We remark that our algorithm benefits from structured images, and so do the other methods. This advantage, enabling the exploitation of self-similarities in images is very important in real life applications, and it is used extensively in the literature \cite{Buades05NLMEANS,Protter09Generalizing}. In many images we are likely to find the same patterns repeating over and over again, and especially so when it comes to small patches.

Note that for peak intensities equal to $0.1, 0.2$ NLSPCA shows better PSNR results than SPDA, while the situation changes if binning is used. The reason is that in such low peak values the counts are so low that working with the same patch-size used with the higher peaks leads to weaker results. This may lead the dictionary learning process to learn isolated noise points as dictionary elements as happens with SPDA in Fig.~\ref{fig:saturn_recovery}.
 Thus, we believe that with larger patch sizes we could get better results for this peak values. However, using large patch sizes is not feasible computationally. Using binning compensates this computational barrier by working on a low resolution version of the image with the same patch size. In the regular size, NL-PCA overcomes the resolution problem by the fact that it uses large cluster sizes. In our case, we cannot use large cluster sizes as we need as many clusters as possible for the dictionary update step.

\begin{table}
\footnotesize
\begin{center}
\begin{tabular}{|c|c|c|c|}
\hline
SPDA Setup & Peak & With Binning & No Binning\\
\hline\hline
Ans+G &&   16.99   &  17.28 \\ 
I & & 17.21&  15.22\\
II & &  17.26 & 15.23\\
III  & 0.1 & 18.84 & 16.50  \\
IV  & &  18.97  &  15.48  \\
V  & & 19.17 & 15.48 \\
\hline
Ans+G &&   17.91     &  18.41 \\ 
I & & 18.38  &   17.78 \\
II & & 18.38  &  18.42  \\
III  & 0.2 & 20.42 & 18.95 \\
IV & & 20.47  &  18.72\\
V  & & 20.75 & 18.57 \\
\hline
Ans+G &&  19.80      &  20.94 \\ 
I & &  18.90  & 20.31 \\
II & & 18.72  & 20.57  \\
III  & 0.5 & 22.45& 21.55 \\
IV & & 22.23  &  21.62 \\
V  & & 22.49 &  21.71\\
\hline
Ans+G &&    20.73     & 23.19 \\ 
I & &  19.14  &  21.92 \\
II & &   19.00  &  22.35  \\
III  & 1 & 23.17 & 23.45 \\
IV & & 23.02 &  23.43  \\
V & & 23.33 & 23.61 \\
\hline
Ans+G &&   21.69     & 25.08  \\ 
I & & 19.27  &  23.27\\
II & &  19.25 & 23.71  \\
III  & 2 & 23.74& 25.41  \\
IV  & & 23.31 & 25.48  \\
V  & & 23.79 &  25.62\\
\hline
Ans+G &&  22.02     &  26.74  \\ 
I & & 19.27 &   24.09  \\
II & & 19.37 & 24.64   \\
III  & 4 & 24.23 & 26.99  \\
IV & & 24.19  &  27.06\\
V  & & 24.66 & 27.15 \\
\hline
\end{tabular}
\end{center}
\caption{{The effect of the different stages in SPDA: We present the PSNR of the outcome of SPDA in different stages and setups of the algorithms: (I) Only sparse coding with $k =2$; (II) Sparse coding with $k =2$ followed by a sparse coding with the bootstrapping based stopping criterion; (III) If no binning is used and the peaks are $0.1, 0.2$ then SPDA with $120$ simple dictionary learning steps is used. Otherwise SPDA with $20$ simple dictionary learning steps followed by another $20$ steps after reclustering is used; (IV) Five advanced dictionary learning steps with no reclustering; (V) Five advanced dictionary learning steps followed by another five steps after reclustering.
\rg{We compare also to a version of the algorithm adapted to Gaussian noise used with Anscombe (Ans+G).}
The average in the results is over five noise realizations and the eight images in Fig.~\ref{fig:test_images}.}}
\label{tbl:exp_psnr_stages}
\end{table}

In conclusion, SPDA seems to have a good denoising quality for Poisson noisy images with low SNR
achieving state-of-the-art recovery performance.
It is interesting to explore the contribution of each stage of the algorithm to the quality of the recovered image.
Therefore we evaluate the performance of SPDA under five different setups: (I) Applying SPDA with only sparse coding with $k =2$; (II) Sparse coding with $k =2$ followed by a sparse coding with the bootstrapping based stopping criterion; (III) Applying SPDA with simple dictionary learning steps without the joint representation and dictionary update stage. If no binning is performed and peak$\le0.2$ we use $120$ learning iterations. Otherwise, we use $20$ dictionary learning iterations as the first stage, then re-cluster and apply additional $20$ iterations; (IV) Applying SPDA with the five advanced dictionary learning steps with no reclustering; (V) Using the setup from Table~\ref{tbl:exp_psnr}, 
five advanced dictionary learning steps followed by another five steps after reclustering.

For each setup we calculate, for six different peak values, the average PSNR over the eight images in Fig.~\ref{fig:test_images}. The result is presented in Table~\ref{tbl:exp_psnr_stages}. First note that the gap between the recovery result of the simple sparse coding (Setup I) and the one of the advanced dictionary learning with reclustering (Setup V) is $2.63dB$ on average which is very significant.

Looking at the contribution of each stage in the algorithm we observe that the
effect of the bootstrapping based stopping criterion (Setup II) is negligible in the case of binning, while it improves the recovery result by $0.3dB$ in the case of no-binning. We believe that the reason is that the number of atoms used in the recovery determines the resolution of the recovered patches. With binning, a coarser resolution of the image is being processed and therefore the reconstruction result is less sensitive to the number of patches used in recovery, while with no-binning patches with finer resolution are used and therefore the number of atoms to be used for the decoding is more critical to the recovery performance.

Thus, it is noticeable that the main improvement in the recovery is due to the dictionary learning. Using only $5$ dictionary learning steps we get an improvement of $2.43dB$ on average. With re-clustering and an additional $5$ dictionary learning steps we get an improvement of another $0.2dB$.

Though the difference between using SPDA with the advanced dictionary learning steps (Setup V) and the simple dictionary learning steps is not significant on average, it has two advantages: (i) For the peaks $0.1,0.2,0.5$ where the usage of binning is preferable the the improvement is $0.23dB$ on average for SPDAbin and for peaks $1,2,4$ where it is better to use no binning the gap is $0.18dB$ for SPDA. (ii) 
The convergence of the algorithm is faster with the advance learning steps: $10$ advanced learning steps versus $40$ or $120$ simple learning steps. Since the bottleneck of our method is the sparse coding step and each learning step is followed by such a one, using SPDA with the advanced dictionary learning steps (setup V) runs four times faster than with the simple steps (setup III).

\rg{Finally, we compare our algorithm to its ``Gaussian version'' with Anscombe.
This version aims at minimizing the standard $\ell_2$ error objective with the standard sparsity constraints. In the dictionary learning steps we use the MOD algorithm \cite{Engan99Method,Engan07Family} with the advanced update step from \cite{Smith13Improving}.
Other than that we adopt exactly the same setup as in SPDA. We present the average denoising error (over the eight test images in Fig.~\ref{fig:test_images}) in Table~\ref{tbl:exp_psnr_stages} (The algorithm is referred to as Ans+G) for the different peak values.
Note that this algorithm does not perform well when used with binning. The reason is that in the binning scenario there are less patches for updating the dictionary. Therefore we do not benefit from the learning process with binning. Interestingly, this is not the case when we work with the Poisson objective. This shows the advantage of working directly with the Poisson data. 
Without binning the performance of the Gaussian version benefits from the dictionary update as there is a larger number of patches for the learning. However, in this case as well it is better to use SPDA. Note that SPDA outperform its Gaussian version for all peak values except peak$=0.1$, for which it is better to use SPDA with binning anyway. }

\section{Discussion and Conclusion}
\label{sec:conc}

This work proposes a new scheme for Poisson denoising, the sparse Poisson denoising algorithm (SPDA).
It relies on the Poisson statistical model presented in \cite{Salmon12Poisson}
and uses a dictionary learning strategy with a sparse coding algorithm that employs a boot-strapping based  stopping criterion.
The recovery performance of the SPDA are state-of-the-art and in some scenarios outperform the existing algorithms
by more than 1db. However, there is still room for improvement in the current scheme which we leave for a future work:

(i) In our current work we used the same initialization dictionary $\matr{D}$ for all types of images.
However, in many applications the type of images to be observed are known beforehand, e.g., space images
or fluorescence microscopy cell images. Off-line training of a dictionary which is dedicated to a specific task
can improve the current results.

(ii) Setting a suitable number of dictionary learning iterations is important for the quality of the reconstruction. 
An automated tuning technique should be considered for this purpose \cite{Deledalle10Poisson, Giryes11GSURE}.
Setting the same number for all images gives a mid-level results (which in overall are better than other existing results).
Basically we can say that almost in all images we have lost output quality because of the fixed stopping point.
We should mention that in some applications this is not an issue as the tuning can be done manually by the user
based on qualitative results and visual feedback.

\section*{Acknowledgment} R. Giryes thanks the Azrieli Foundation for the Azrieli Fellowship. This research was supported by European Community's FP7- ERC program, grant agreement no. 320649.  In addition, the authors thank the reviewers
of the manuscript for their suggestions which greatly improved the paper.

{\small
\bibliographystyle{IEEEtran}
\bibliography{PoissonDenoising}

\begin{thebibliography}{10}
\providecommand{\url}[1]{#1}
\csname url@samestyle\endcsname
\providecommand{\newblock}{\relax}
\providecommand{\bibinfo}[2]{#2}
\providecommand{\BIBentrySTDinterwordspacing}{\spaceskip=0pt\relax}
\providecommand{\BIBentryALTinterwordstretchfactor}{4}
\providecommand{\BIBentryALTinterwordspacing}{\spaceskip=\fontdimen2\font plus
\BIBentryALTinterwordstretchfactor\fontdimen3\font minus
  \fontdimen4\font\relax}
\providecommand{\BIBforeignlanguage}[2]{{%
\expandafter\ifx\csname l@#1\endcsname\relax
\typeout{** WARNING: IEEEtran.bst: No hyphenation pattern has been}%
\typeout{** loaded for the language `#1'. Using the pattern for}%
\typeout{** the default language instead.}%
\else
\language=\csname l@#1\endcsname
\fi
#2}}
\providecommand{\BIBdecl}{\relax}
\BIBdecl

\bibitem{Salmon12PoissonConf}
J.~Salmon, C.-A. Deledalle, R.~Willett, and Z.~T. Harmany, ``Poisson noise
  reduction with non-local {PCA},'' in \emph{ICASSP, 2012.}, March 2012, pp.
  1109--1112.

\bibitem{Salmon12Poisson}
J.~Salmon, Z.~Harmany, C.-A. Deledalle, and R.~Willett,
  ``\BIBforeignlanguage{English}{Poisson noise reduction with non-local
  {PCA}},'' \emph{\BIBforeignlanguage{English}{Journal of Mathematical Imaging
  and Vision}}, pp. 1--16, 2013.

\bibitem{Danielyan11Deblurring}
A.~Danielyan, V.~Katkovnik, and K.~Egiazarian, ``Deblurring of {P}oissonian
  images using {BM3D} frames,'' \emph{Proc. SPIE}, vol. 8138, pp. 813 812--813
  812--7, 2011.

\bibitem{Rodrigo11Efficient}
E.~Gil-Rodrigo, J.~Portilla, D.~Miraut, and R.~Suarez-Mesa, ``Efficient joint
  poisson-gauss restoration using multi-frame l2-relaxed-l0 analysis-based
  sparsity,'' in \emph{{IEEE International Conference on Image Processing
  (ICIP)}}, Sept. 2011, pp. 1385--1388.

\bibitem{Figueiredo10Restoration}
M.~A.~T. Figueiredo and J.~Bioucas-Dias, ``Restoration of {P}oissonian images
  using alternating direction optimization,'' \emph{IEEE Trans. Image
  Process.}, vol.~19, no.~12, pp. 3133--3145, Dec. 2010.

\bibitem{Zhang12Novel}
X.~Zhang, Y.~Lu, and T.~Chan, ``A novel sparsity reconstruction method from
  {P}oisson data for {3D} bioluminescence tomography,'' \emph{J. Sci. Comput.},
  vol.~50, no.~3, pp. 519--535, Mar. 2012.

\bibitem{Boulanger10Patch}
J.~Boulanger, C.~Kervrann, P.~Bouthemy, P.~Elbau, J.-B. Sibarita, and
  J.~Salamero, ``Patch-based nonlocal functional for denoising fluorescence
  microscopy image sequences,'' \emph{IEEE Trans. on Med. Imag.}, vol.~29,
  no.~2, pp. 442--454, Feb. 2010.

\bibitem{Makitalo11Optimal}
M.~Makitalo and A.~Foi, ``Optimal inversion of the {Anscombe} transformation in
  low-count {Poisson} image denoising,'' \emph{IEEE Trans. on Image Proces.},
  vol.~20, no.~1, pp. 99--109, Jan. 2011.

\bibitem{Zhang08Wavelets}
B.~Zhang, J.~Fadili, and J.~Starck, ``Wavelets, ridgelets, and curvelets for
  poisson noise removal,'' \emph{IEEE Trans. on Image Processing}, vol.~17,
  no.~7, pp. 1093--1108, July 2008.

\bibitem{anscombe48transformation}
F.~J. Anscombe, ``The transformation of {P}oisson, {B}inomial and
  negative-{B}inomial data,'' \emph{Biometrika}, vol.~35, no. 3-4, pp.
  246--254, 1948.

\bibitem{Fisz55Limiting}
M.~Fisz, ``The limiting distribution of a function of two independent random
  variables and its statistical application,'' \emph{Colloquium Mathematicum},
  vol.~3, pp. 138--146, 1955.

\bibitem{Dabov07BM3D}
K.~Dabov, A.~Foi, V.~Katkovnik, and K.~Egiazarian, ``Image denoising by sparse
  3-d transform-domain collaborative filtering,'' \emph{IEEE Trans. on Image
  Processing}, vol.~16, no.~8, pp. 2080--2095, 2007.

\bibitem{Mairal09Non}
J.~Mairal, F.~Bach, J.~Ponce, G.~Sapiro, and A.~Zisserman, ``Non-local sparse
  models for image restoration,'' in \emph{ICCV, 2009}, 2009, pp. 2272--2279.

\bibitem{Yu12Solving}
G.~Yu, G.~Sapiro, and S.~Mallat, ``Solving inverse problems with piecewise
  linear estimators: From {G}aussian mixture models to structured sparsity,''
  \emph{IEEE Trans. on Image Processing}, vol.~21, no.~5, pp. 2481 --2499, may
  2012.

\bibitem{Harmany12SPIRAL}
Z.~Harmany, R.~Marcia, and R.~Willett, ``This is {SPIRAL-TAP}: Sparse poisson
  intensity reconstruction algorithms -- theory and practice,'' \emph{IEEE
  Trans. on Image Processing}, vol.~21, no.~3, pp. 1084 --1096, march 2012.

\bibitem{Giryes12Sparsity}
R.~Giryes and M.~Elad, ``Sparsity based poisson denoising,'' in \emph{IEEE 27th
  Convention of Electrical Electronics Engineers in Israel (IEEEI), 2012},
  2012, pp. 1--5.

\bibitem{Dupe13greedy}
F.-X. Dupe and S.~Anthoine, ``A greedy approach to sparse {P}oisson
  denoising,'' in \emph{IEEE International Workshop on Machine Learning for
  Signal Processing (MLSP), 2013}, Sept 2013, pp. 1--6.

\bibitem{Dinh14Composite}
A.~K. Quoc Tran-Dinh and V.~Cevher, ``Composite self-concordant minimization,''
  \emph{CoRR}, vol. abs/1308.2867, 2014.

\bibitem{Giryes14Inpainting}
R.~Giryes and M.~Elad, ``Sparsity based poisson inpainting,'' in \emph{to
  appear in IEEE International Conference on Image Processing ({ICIP})}, 2014.

\bibitem{Smith13Improving}
L.~Smith and M.~Elad, ``Improving dictionary learning: Multiple dictionary
  updates and coefficient reuse,'' \emph{IEEE Signal Processing Letters},
  vol.~20, no.~1, pp. 79--82, Jan. 2013.

\bibitem{Elad10Sparse}
M.~Elad, \emph{Sparse and Redundant Representations: From Theory to
  Applications in Signal and Image Processing}, 1st~ed.\hskip 1em plus 0.5em
  minus 0.4em\relax Springer Publishing Company, Incorporated, 2010.

\bibitem{Lingenfelter09Sparsity}
D.~J. Lingenfelter, J.~A. Fessler, and Z.~He, ``Sparsity regularization for
  image reconstruction with poisson data,'' \emph{Proc. SPIE}, vol. 7246, pp.
  72\,460F--72\,460F--10, 2009.

\bibitem{Mester11Improving}
H.~Burger and S.~Harmeling, ``Improving denoising algorithms via a multi-scale
  meta-procedure,'' in \emph{Pattern Recognition}, ser. Lecture Notes in
  Computer Science, R.~Mester and M.~Felsberg, Eds.\hskip 1em plus 0.5em minus
  0.4em\relax Springer Berlin Heidelberg, 2011, vol. 6835, pp. 206--215.

\bibitem{Deledalle10Poisson}
C.-A. Deledalle, F.~Tupin, and L.~Denis, ``Poisson {NL} means: Unsupervised non
  local means for poisson noise,'' in \emph{ICIP, 2010}, sept. 2010, pp. 801
  --804.

\bibitem{Buades05NLMEANS}
A.~Buades, B.~Coll, and J.~Morel, ``A review of image denoising algorithms,
  with a new one,'' \emph{Multiscale Model. Simul.}, vol.~4, no.~2, pp.
  490--530, 2005.

\bibitem{Protter09Generalizing}
M.~Protter, M.~Elad, H.~Takeda, and P.~Milanfar, ``Generalizing the
  nonlocal-means to super-resolution reconstruction,'' \emph{IEEE Trans. Image
  Process.}, vol.~18, no.~1, pp. 36--51, Jan 2009.

\bibitem{Engan99Method}
K.~Engan, S.~Aase, and J.~Hakon~Husoy, ``Method of optimal directions for frame
  design,'' in \emph{IEEE International Conference on Acoustics, Speech and
  Signal Processing (ICASSP)}, vol.~5, 1999, pp. 2443--2446.

\bibitem{Engan07Family}
K.~Engan, K.~Skretting, and J.~H. Husøy, ``Family of iterative ls-based
  dictionary learning algorithms, ils-dla, for sparse signal representation,''
  \emph{Digital Signal Processing}, vol.~17, no.~1, pp. 32 -- 49, 2007.

\bibitem{Giryes11GSURE}
R.~Giryes, M.~Elad, and Y.~C. Eldar, ``The projected {GSURE} for automatic
  parameter tuning in iterative shrinkage methods,'' \emph{Applied and
  Computational Harmonic Analysis}, vol.~30, no.~3, pp. 407 -- 422, 2011.

\end{thebibliography}
}

\end{document}